%% file: manuscript.tex
\documentclass[a4paper,11pt]{article}

\input{math_commands.tex}

\usepackage{amsmath,amsfonts,amssymb,amsthm}

\usepackage[margin=1in]{geometry}

\usepackage{parskip}

\usepackage{hyperref}
\usepackage{xcolor}
\hypersetup{
    colorlinks,
    linkcolor={red!50!black},
    citecolor={blue!50!black},
    urlcolor={blue!80!black}
}

\usepackage{url}

\usepackage{graphicx}
\usepackage{bmpsize}
\usepackage{grffile}

\usepackage{authblk}

\usepackage{bbm}
\usepackage{booktabs}
\usepackage{caption}
\usepackage{subcaption} 

\title{Robustifying State-space Models for Long Sequences via Approximate Diagonalization}

\author{
	Annan Yu,$^{1}$\thanks{Corresponding author: \url{ay262@cornell.edu}.} \hspace{+0.3cm} 
 Arnur Nigmetov,$^2$ \hspace{+0.3cm}  
 Dmitriy Morozov,$^2$ \hspace{+0.3cm} \vspace{0.2cm}  
 Michael W. Mahoney,$^{2,3,4}$ \hspace{+0.5cm} 
 N. Benjamin Erichson$^{2,3}$  \vspace{+0.3cm} \\ 
	
	$^1$ Center for Applied Mathematics, Cornell University \\
	$^2$ Lawrence Berkeley National Laboratory \\
	$^3$ International Computer Science Institute \\
	$^4$ Department of Statistics, University of California at Berkeley
}

\date{}

\begin{document}

\maketitle

\begin{abstract}

State-space models (SSMs) have recently emerged as a framework for learning long-range sequence tasks. 
An example is the structured state-space sequence (S4) layer, which uses the diagonal-plus-low-rank structure of the HiPPO initialization framework. 
However, the complicated structure of the S4 layer poses challenges; and, in an effort to address these challenges, models such as S4D and S5 have considered a purely diagonal structure. 
This choice simplifies the implementation, improves computational efficiency, and allows channel communication.
However, diagonalizing the HiPPO framework is itself an ill-posed problem.
In this paper, we propose a general solution for this and related ill-posed diagonalization problems in machine learning.
We introduce a generic, backward-stable ``perturb-then-diagonalize'' (PTD) methodology, which is based on the pseudospectral theory of non-normal operators, and which may be interpreted as the approximate diagonalization of the non-normal matrices defining SSMs.
Based on this, we introduce the S4-PTD and S5-PTD models.
Through theoretical analysis of the transfer functions of different initialization schemes, we demonstrate that the S4-PTD/S5-PTD initialization strongly converges to the HiPPO framework, while the S4D/S5 initialization only achieves weak convergences. 
As a result, our new models show resilience to Fourier-mode noise-perturbed inputs, a crucial property not achieved by the S4D/S5 models. 
In addition to improved robustness, our S5-PTD model averages 87.6\% accuracy on the Long-Range Arena benchmark, demonstrating that the PTD methodology helps to improve the accuracy of deep learning models.
\end{abstract}

\input{body.tex}

\subsubsection*{Acknowledgments}

This work was supported by the U.S. Department of Energy, Office of Science,
Office of Advanced Scientific Computing Research,
Scientific Discovery through Advanced Computing (SciDAC) program,
under Contract Number DE-AC02-05CH11231 at Lawrence Berkeley National Laboratory.
It used the Lawrencium computational cluster provided by the IT Division at the Lawrence Berkeley National Laboratory
and resources of the National Energy Research Scientific Computing Center
(NERSC, using award ASCR-ERCAP0023337),
a U.S. Department of Energy Office of Science User Facility located at Lawrence Berkeley National Laboratory,
both operated under Contract No. DE-AC02-05CH11231.
NBE would also like to acknowledge NSF, under Grant No. 2319621, for providing partial support of this work. Our conclusions do not necessarily reflect the position or the policy of our sponsors, and no official endorsement should be inferred.

\bibliography{references}
\bibliographystyle{plain}

\clearpage
\input{supplement}

\end{document}

%% file: math_commands.tex
\usepackage{amsmath,amsfonts,bm}

\usepackage{amsmath,amssymb,amsthm,enumerate,enumitem,hyperref,algorithm,algorithmic,cancel,xcolor,array,tabularx,cleveref,overpic,ctable,cases,diagbox,pifont}

\usepackage[labelfont=bf,tableposition=top]{caption}
\usepackage{mathtools}
\usepackage{esint}
\usepackage{overpic}

\theoremstyle{definition}

\newtheorem{thm}{Theorem}

\newtheorem{lem}{Lemma}

\usepackage{listings}
\usepackage[mathscr]{euscript}

\newcommand{\N}{\mathbb{N}}

\newcommand{\R}{\mathbb{R}}
\newcommand{\C}{\mathbb{C}}

\newcommand{\norm}[1]{\left\lVert#1\right\rVert}

\newcommand{\abs}[1]{\left|#1\right|}

\newcommand{\comment}[1]{}

\def\eqref#1{equation~\ref{#1}}

\def\1{\bm{1}}

\DeclareMathAlphabet{\mathsfit}{\encodingdefault}{\sfdefault}{m}{sl}
\SetMathAlphabet{\mathsfit}{bold}{\encodingdefault}{\sfdefault}{bx}{n}



%% file: body.tex
\section{Introduction}

Sequential data are pervasive across a wide range of fields, including natural language processing, speech recognition, robotics and autonomous systems, as well as scientific machine learning and financial time-series analysis, among others. 
Given that many of these applications produce exceedingly long sequences, sequential models need to capture long-range temporal dependencies in order to yield accurate predictions.
To this end, many specialized deep learning methods have been developed to deal with long sequences, including recurrent neural networks (RNNs)~\cite{arjovsky2016unitary,chang2019antisymmetricrnn,erichson2021lipschitz,rusch2021unicornn,erichson2022gated,orvieto2023resurrecting}, convolutional neural networks (CNNs)~\cite{bai2018empirical,romero2021ckconv}, continuous-time models (CTMs)~\cite{gu2021combining,yildiz2021continuous}, and transformers~\cite{katharopoulos2020transformers,choromanski2020rethinking,kitaev2020reformer,zhou2022fedformer,nie2023a}.

Over the past few years, the new class of state-space models (SSMs) gained vast popularity for sequential modeling due to their outstanding performance on the Long-Range Arena (LRA) dataset~\cite{tay2020long}. 
An SSM is built upon a continuous-time linear time-invariant (LTI) dynamical system $\Sigma = (\mathbf{A},\mathbf{B},\mathbf{C},\mathbf{D})$, which is a system of linear ODEs given by
\begin{equation}
	\label{eq.LTIDS}
	\begin{aligned}
		\mathbf{x}'(t) &= \mathbf{A} \mathbf{x}(t) + \mathbf{B} \mathbf{u}(t), \\
		\mathbf{y}(t) &= \mathbf{C} \mathbf{x}(t) + \mathbf{D} \mathbf{u}(t),
	\end{aligned}
\end{equation}
where $\mathbf{A} \in \C^{n \times n}$, $\mathbf{B} \in \C^{n \times m}$, $\mathbf{C} \in \C^{p \times n}$, $\mathbf{D} \in \C^{p \times m}$ are the state, input, output and feedthrough matrices; and $\mathbf{u}(t) \in \C^m, \mathbf{x}(t) \in \C^n, \mathbf{y}(t) \in \C^p$ are the inputs, states, and outputs of the system, respectively. The system can be discretized at time steps $j\Delta t$, where $\Delta t > 0$ and $j = 1, \ldots, L$, to be fed with sequential inputs of length $L$. To store and process the information of the long sequential inputs online, the SSMs are often initialized by a pre-designed LTI system. 
One of the most popular schemes is called ``HiPPO initialization''~\cite{voelker2019legendre,gu2020hippo}, in which the Legendre coefficients of the input history at time $t$, i.e., $\mathbf{u} \cdot \mathbbm{1}_{[0,t]}$, are stored and updated in the state vector $\mathbf{x}(t)$.
This initialization is specifically designed to model long-range dependencies in sequential data. 
The recently proposed S4 model~\cite{gu2022efficiently} leverages the HiPPO initialization and accelerates training and inference by decomposing $\mathbf{A}$ into the sum of a diagonal matrix and a low-rank one. 
The diagonal-plus-low-rank (DPLR) structure yields a barycentric representation~\cite{antoulas1986scalar} of the transfer function of~\cref{eq.LTIDS} that maps inputs to outputs in the frequency domain, enabling fast computation in the frequency domain~\cite{aumann2023practical}.

While the DPLR structure achieves an asymptotic speed-up of the model, considering $\mathbf{A}$ to be a diagonal matrix results in a simpler structure. Compared to a DPLR matrix $\mathbf{A}$, a diagonal SSM is not only faster to compute and easier to implement, but it also allows integrating channel communication via parallel scans~\cite{smith2023simplified}, thereby improving its performance on long-range tasks. Unfortunately, the problem of diagonalizing the HiPPO framework is exponentially ill-conditioned, as $n$ increases. Hence, while~\cite{gu2022efficiently} shows analytic forms of the eigenvalues and eigenvectors of HiPPO matrices, they suffer from an exponentially large variance and cannot be used in practice. 
So far, the most popular way of obtaining a diagonal SSM is to simply discard the low-rank part from the DPLR structure, leveraging a stable diagonalization algorithm for a normal matrix. 
Discarding the low-rank component changes the underlying diagonalization problem, however; and it abandons the theoretical insights about HiPPO.  Still, the resulting model almost matches S4's performance, in practice. Such diagonal models are called S4D~\cite{gu2022parameterization} when the systems are single-input/single-output (i.e., $m = p = 1$) and S5~\cite{smith2023simplified} when the systems are multiple-input/multiple-output (i.e., $m = p > 1$), which enables channel communication.

The issue of ill-posed diagonalization problems is not merely specific to SSMs. For example, it is known that non-normal matrices make RNNs more expressive~\cite{kerg2019non,orhan2019improved}. 
More generally, non-normality plays an important role in the training of certain neural networks~\cite{sengupta2018robust,kumar2022non}. While the ill-posedness of the diagonalization problem essentially prevents accurate computation of eigenvalues and eigenvectors (i.e., we cannot have a small forward error) --- in fact, the true spectral information becomes meaningless in this case --- using a backward stable eigensolver, one can recover the non-normal matrix accurately (i.e., we can have a small backward error) from the wrong eigenvalues and eigenvectors.

In this paper, we propose a generic ``perturb-then-diagonalize" (PTD) methodology as a backward stable eigensolver. 
PTD is based on the idea that a small random perturbation remedies the problem of the blowing up of eigenvector condition number~\cite{davies2008approximate,davies2009perturbations,banks2021gaussian}, regularizing the ill-posed problem into a close but well-posed one.
It is based on the pseudospectral theory of non-normal operators~\cite{trefethen2005spectra} and may be interpreted as the approximate diagonalization of the non-normal matrices.
In the context of SSMs, our PTD method can be used to diagonalize the highly non-normal HiPPO framework. Based on this, we introduce the S4-PTD and S5-PTD models. Our method is flexible, and it can be used to diagonalize many SSM initialization schemes that may be invented in the future.

\textbf{Contribution.} \ Here are our main contributions:
\begin{enumerate}[leftmargin=*]
\item We propose a ``perturb-then-diagonalize" (PTD) methodology that solves ill-posed diagonalization problems in machine learning when only the backward error is important.

\item We provide a fine-grained analysis that compares the S4 and the S4D initialization. In particular, we quantify the change of the transfer function when discarding the low-rank part of HiPPO, which is done in the diagonal S4D/S5 initialization. We show that while the outputs of the S4D/S5 system on a \textit{fixed} smooth input converge to those of the S4 system at a linear rate as $n \rightarrow \infty$ (see~\cref{sec:converge}), the convergence is not uniform across all input functions (see~\cref{sec:noconverge}). 

\item Based on our theoretical analysis, we observe, using both a synthetic example (see~\cref{sec:failuremodes}) and the sequential CIFAR task (see~\cref{sec:robustness}), that the S4D/S5 models are very sensitive to certain Fourier-mode input perturbations, which impairs the robustness of the models. 

\item Based on diagonalizing a perturbed HiPPO matrix, we propose the S4-PTD and S5-PTD models.
Our models are robust to Fourier-mode input perturbations. We theoretically estimate the effect of the perturbation (see~\cref{sec:S4-PTD}). We propose computing the perturbation matrix by solving an optimization problem with a soft constraint. Moreover, our method is not restricted to the HiPPO matrix but can be applied to any initializations.

\item We provide an ablation study for the size of the perturbation in our models. We also evaluate our S4-PTD and S5-PTD models on LRA tasks, which reveals that the S4-PTD model outperforms the S4D model, while the S5-PTD model is comparable with the S5 model (see~\cref{sec:LRA}).
\end{enumerate}

\section{Preliminaries and notation}
\label{sec:prelim}

Given an LTI system in~\cref{eq.LTIDS}, we say it is asymptotically stable if the eigenvalues $\lambda_j$ of $\mathbf{A}$ are all contained in the left half-plane, i.e., if $\text{Re}(\lambda_j) < 0$ for all $1 \leq j \leq n$. 
The \emph{transfer function} of the LTI system is defined by
\begin{equation}
\label{eq.transfer}
G(s) = \mathbf{C} (s\mathbf{I} - \mathbf{A})^{-1} \mathbf{B} + \mathbf{D}, \qquad s \in \C \setminus \Lambda(\mathbf{A}),
\end{equation}
where $\mathbf{I} \in \R^{n \times n}$ is the identity matrix and $\Lambda(\mathbf{A})$ is the spectrum of $\mathbf{A}$. 
The transfer function $G$ is a rational function with $n$ poles (counting multiplicities) at the eigenvalues of $\mathbf{A}$. 
Assume $\mathbf{x}(0) = \boldsymbol{0}$. 
Then the transfer function maps the inputs to the outputs of the LTI system in the Laplace domain by multiplication, i.e.,
$(\mathcal{L}\mathbf{y})({s}) = G({s}) (\mathcal{L}\mathbf{u})({s})$ for all $s \in \C$,
where $\mathcal{L}$ is the Laplace transform operator (see~\cite{zhou1998essentials}). 
Assume the LTI system in~\cref{eq.LTIDS} is asymptotically stable and the input $\mathbf{u}(t)$ is bounded and integrable (with respect to the Lebesgue measure) as $t$ ranges over $\R$.
Then the Laplace transform reduces to the Fourier transform:
\begin{equation}\label{eq.transferfourier}
\hat{\mathbf{y}}(s) = G(is) \hat{\mathbf{u}}(s), \qquad s \in \R,
\end{equation}
where $\hat{\mathbf{y}}$ and $\hat{\mathbf{u}}$ are the Fourier transforms of $\mathbf{y}$ and $\mathbf{u}$, respectively, and $i$ is the imaginary unit. 
Let $\mathbf{V} \in \C^{n \times n}$ be an invertible matrix. 
We can conjugate the system $(\mathbf{A},\mathbf{B},\mathbf{C},\mathbf{D})$ by $\mathbf{V}$, which yields $(\mathbf{V}^{-1} \mathbf{A} \mathbf{V}, \mathbf{V}^{-1} \mathbf{B},\mathbf{C} \mathbf{V},\mathbf{D})$. 
Since the transfer function is conjugation-invariant, the two systems map the same inputs $\mathbf{u}(\cdot)$ to the same outputs $\mathbf{y}(\cdot)$, while the states $\mathbf{x}(\cdot)$ are transformed by $\mathbf{V}$. If $\mathbf{A}$ is a normal matrix, i.e., $\mathbf{A} \mathbf{A}^* = \mathbf{A}^* \mathbf{A}$, then $\mathbf{V}$ is unitary, in which case transforming the states by $\mathbf{V}$ is a well-conditioned problem and can be done without loss of information. Issues arise, however, when $\mathbf{A}$ is non-normal and $\mathbf{V}$ is ill-conditioned.

The state-space models use LTI systems to process time series inputs. 
Different initializations can be tailored to tasks with different natures, such as the range of dependency~\cite{gu2022train}.
A particularly successful initialization scheme used in the S4 model is the so-called HiPPO initialization. 
While there exist several variants of HiPPO, the most popular HiPPO-LegS matrices are defined by
\begin{equation}
\label{eq.HiPPO}
(A_H)_{jk} = - \begin{cases}
	\mathbbm{1}_{\{j > k\}} \sqrt{2j-1}\sqrt{2k-1}&, \qquad \text{if } j \neq k,\\
	j&, \qquad \text{if } j = k,
\end{cases}
\qquad
(B_H)_{j\ell} = \sqrt{(2j-1)/2},
\end{equation}
for all $1 \leq j, k \leq n$ and $1 \leq \ell \leq m$, where $\mathbbm{1}_{\{j > k\}}$ is the indicator that equals $1$ if $j > k$ and $0$ otherwise. 
Such a system guarantees that the Legendre coefficients of the input history $\mathbf{u} \cdot \mathbbm{1}_{[0,t]}$ (with respect to a scaled measure) are stored in the states $\mathbf{x}(t)$ over time~\cite{gu2020hippo}. 
Since computing with the dense matrix $\mathbf{A}_H$ is practically inefficient, one conjugates the HiPPO system with a matrix $\mathbf{V}_H$ to simplify the structure of $\mathbf{A}_H$. The matrix $\mathbf{A}_H$ in~\cref{eq.HiPPO} has an ill-conditioned eigenvector matrix~\cite{gu2022efficiently}; consequently, instead of solving the ill-posed problem that diagonalizes $\mathbf{A}_H$, one exploits a diagonal-plus-low-rank (DPLR) structure: 
\begin{equation}
\label{eq.DPLR}
\mathbf{A}_{H} = \mathbf{A}_{H}^\perp - \frac{1}{2}\mathbf{B}_{H}\mathbf{B}_{H}^\top, \qquad (A_H^\perp)_{jk} = -\frac{1}{2}\begin{cases}
	(-1)^{\mathbbm{1}_{\{j < k\}}}\sqrt{2j-1}\sqrt{2k-1} & , \qquad j \neq k, \\
	1 & , \qquad j = k,
\end{cases}
\end{equation}
where $\mathbf{A}_{H}^\perp$ is a skew-symmetric matrix that can be unitarily diagonalized into $\mathbf{A}_{H}^\perp = \mathbf{V}_{H} \boldsymbol{\Lambda}_{H} \mathbf{V}_{H}^{-1}$. 
The S4 model leverages the HiPPO matrices by initializing 
\begin{equation}
\label{eq.initS4}
\mathbf{A}_{\text{DPLR}} = \boldsymbol{\Lambda}_{H} - \frac{1}{2}\mathbf{V}_{H}^{-1} \mathbf{B}_H \mathbf{B}_H^\top \mathbf{V}_{H}, \qquad \mathbf{B}_{\text{DPLR}} = \mathbf{V}_{H}^{-1} \mathbf{B}_H
\end{equation}
and $\mathbf{C}_{\text{DPLR}}$ and $\mathbf{D}_{\text{DPLR}}$ randomly. 
Such a system $\Sigma_{\text{DPLR}} = (\mathbf{A}_{\text{DPLR}},\mathbf{B}_{\text{DPLR}},\mathbf{C}_{\text{DPLR}},\mathbf{D}_{\text{DPLR}})$ is conjugate via $\mathbf{V}_H$ to $(\mathbf{A}_H, \mathbf{B}_H, \mathbf{C}_{\text{DPLR}}\mathbf{V}_H^{-1}, \mathbf{D}_{\text{DPLR}})$.
Hence, they share the transfer function and the same mapping from the inputs $\mathbf{u}(\cdot)$ to the outputs $\mathbf{y}(\cdot)$. 
The S4D model further simplifies the structure by discarding the rank-$1$ part from $\mathbf{A}_H$ and therefore initializes 
\begin{equation}
\label{eq.initS4D}
\mathbf{A}_{\text{Diag}} = \boldsymbol{\Lambda}_{H}, \qquad \mathbf{B}_{\text{Diag}} = \frac{1}{2} \mathbf{V}_{H}^{-1} \mathbf{B}_H,
\end{equation}
and $\mathbf{A}_{\text{Diag}}$ is henceforth restricted to be diagonal. 
While both the S4 and S4D models restrict that $m = p = 1$, i.e., the LTI systems are single-input/single-output (SISO), the S5 model, which also initializes $\mathbf{A}_{\text{Diag}} = \boldsymbol{\Lambda}_H$ and requires it to be diagonal throughout training, leverages multiple-input/multiple-output (MIMO) systems by allowing $m = p > 1$. We provide more background information on LTI systems and state-space models in sequential modeling in Appendix~\ref{sec:morebackground}.

Throughout this paper, we use $\norm{\cdot}$ to denote a vector or matrix $2$-norm. Given an invertible square matrix $\mathbf{V}$, we use $\kappa(\mathbf{V}) = \|\mathbf{V}\| \|\mathbf{V}^{-1}\|$ to denote its condition number. Given a number $1 \leq p \leq \infty$ and a measurable function $f: \R \rightarrow \C$, we use $\norm{f}_{L^p}$ for the standard $L^p$-norm of $f$ with respect to the Lebesgue measure on $\R$ and $L^p(\R) = \{f: \R \rightarrow \C \mid \norm{f}_{L^p} < \infty\}$.

\section{Theory of the diagonal initialization of state-space models}
\label{sec:theorydiag}

The S4 model proposes to initialize the SSM to store the Legendre coefficients of the input signal in the states $\mathbf{x}$~\cite{gu2020hippo}. This initialization, however, has an ill-conditioned spectrum, preventing a stable diagonalization of the SSM. On the other hand, the S4D model uses a different initialization scheme that has a stable spectrum, allowing for stable diagonalization; however, such initialization lacks an interpretation of the states $\mathbf{x}$. In this section, we conduct a fine-grained analysis of the two initializations, which shows that: 
\begin{enumerate}
    \item for any fixed input signal $\mathbf{u}(\cdot)$ with sufficient smoothness, the outputs of the two systems $\Sigma_{\text{DPLR}}$ and $\Sigma_{\text{Diag}}$ converge to each other with a linear rate (of which the previous analysis is devoid) as $n \rightarrow \infty$ (see~\cref{sec:converge}); and 
    \item by viewing $\Sigma_{\text{DPLR}}$ and $\Sigma_{\text{Diag}}$ as linear operators that map input signals to the outputs, the operators do not converge in the operator norm topology as $n \rightarrow \infty$ (see~\cref{sec:noconverge}). 
\end{enumerate}
While the first observation partially justifies the success of the S4D model, the second one allows us to observe that the diagonal initialization is unstable under certain Fourier-mode input perturbations (see~\cref{sec:failuremodes}). 
In this section, we assume $m = p = 1$, which is consistent with the S4 and S4D models. Still, our theory can be related to the S5 model, as shown in~\cite{smith2023simplified}.

\subsection{A simplified formula for the transfer function deviation}

\begin{figure}
\centering
\begin{overpic}[angle=270, width = 0.8\textwidth]{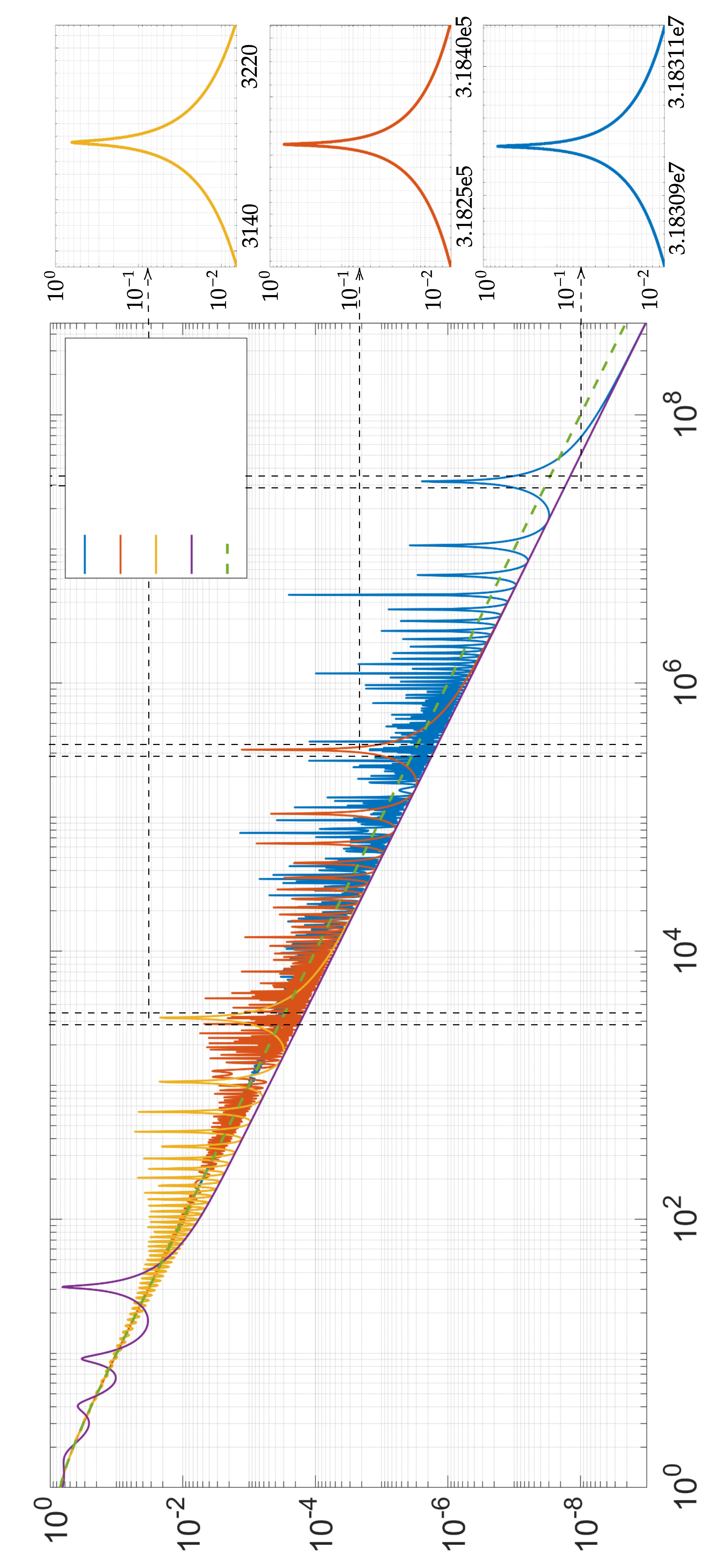}
	\put(38,-1) {\small{frequency}}
	\put(-2.5,17.5) {\rotatebox{90}{\small{magnitude}}}
	\put(66,39.9) {\scalebox{.55}{$G_{\text{Diag}} (n \!=\! 10)$}}
	\put(66,37.6) {\scalebox{.55}{$G_{\text{Diag}} (n \!=\! 100)$}}
	\put(66,35.3) {\scalebox{.55}{$G_{\text{Diag}} (n \!=\! 1000)$}}
	\put(66,33.0) {\scalebox{.55}{$G_{\text{Diag}} (n \!=\! 10000)$}}
	\put(66,30.7) {\scalebox{.55}{$G_{\text{DPLR}}$}}
\end{overpic} \vspace{0.0cm}
\caption{The magnitude of transfer function of the S4 model, $\abs{G_{\text{DPLR}}(si)}$, and that of the S4D model, $\abs{G_{\text{Diag}}(si)}$ with $\mathbf{C}_{\text{DPLR}} = \mathbf{C}_{\text{Diag}} = \mathbf{e}_1^\top \mathbf{V}_H$ and the SSM size $n$ set to different values. Note that $G_{\text{DPLR}}$ stays the same regardless of $n$. By zooming into the last spike of $\abs{G_{\text{Diag}}(si)}$, we see that the peak remains $\Theta(1)$ as $n \rightarrow \infty$ (see the right panels). The figure shows that $G_{\text{Diag}}$ is oscillatory while $G_{\text{DPLR}}$ is smooth; moreover, $\abs{G_{\text{Diag}}(si)}$ does not converge to $\abs{G_{\text{DPLR}}(si)}$ uniformly.}
\label{fig:transfer}
\end{figure}
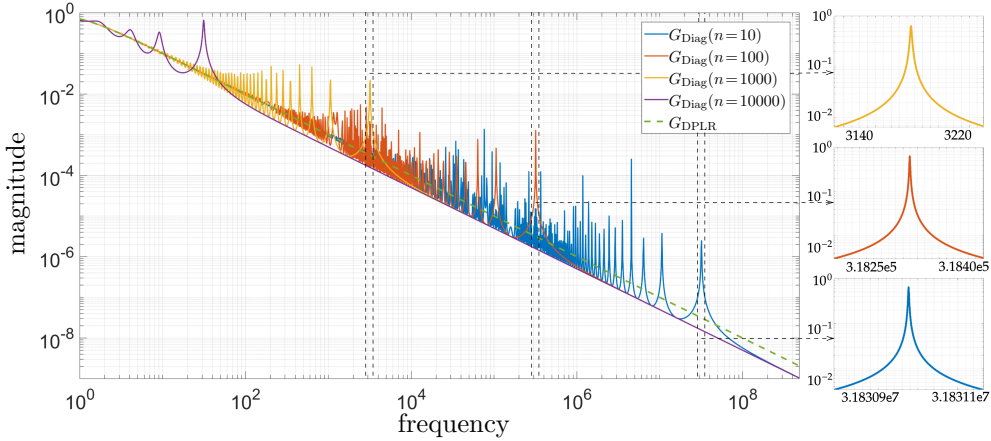

Fix an integer $1 \leq \ell \leq n$. We assume that $\mathbf{C}_{\text{DPLR}} = \mathbf{C}_{\text{Diag}} = \mathbf{e}_\ell^\top \mathbf{V}_H$, where $\mathbf{e}_\ell^\top$ is the $\ell$th standard basis, and $\mathbf{D}_{\text{DPLR}} = \mathbf{D}_{\text{Diag}}$. For a general $\mathbf{C}_{\text{DPLR}} = \mathbf{C}_{\text{Diag}}$, we can decompose it onto the orthonormal basis $\{\mathbf{e}_\ell^\top \mathbf{V}_H \mid 1 \leq \ell \leq n\}$ and study each component separately using the theory developed in this section. Let $G_{\text{DPLR}}$ and $G_{\text{Diag}}$ be the transfer functions of $\Sigma_{\text{DPLR}}$ and $\Sigma_{\text{Diag}}$, respectively, i.e.,
\begin{equation}\label{eq.twotransfer}
G_{\text{DPLR}\!}(s) \!=\! \mathbf{C}_{\text{DPLR}\!} (s\mathbf{I} \!-\! \mathbf{A}_{\text{DPLR}})^{\!-1} \mathbf{B}_{\text{DPLR}\!} + \mathbf{D}_{\text{DPLR}}, \,\, G_{\text{Diag}\!}(s) \!=\! \mathbf{C}_{\text{Diag}\!} (s\mathbf{I} \!-\! \mathbf{A}_{\text{Diag}})^{\!-1} \mathbf{B}_{\text{Diag}\!} + \mathbf{D}_{\text{Diag}}.
\end{equation}
Our first result (\Cref{lem.difftransfer}) concerns $G_{\text{DPLR}} - G_{\text{Diag}}$. Recall that by~\cref{eq.transferfourier}, the difference between the two LTI systems is controlled by $G_{\text{DPLR}} - G_{\text{Diag}}$. In particular, $|G_{\text{DPLR}}(si) - G_{\text{Diag}}(si)|$ measures the difference between the outputs of the two systems given a frequency-$s$ input.
We use Lemma~\ref{lem.difftransfer} to study the (non-)convergence of the two systems (see~\Cref{thm.weakstarconverge} and~\ref{thm.noconverge}); moreover, it reveals the ways that the S4D model can become unstable (see~\Cref{fig:transfer} and~\cref{sec:failuremodes}).

\begin{lem}\label{lem.difftransfer}
Let $G_{\text{DPLR}}$ and $G_{\text{Diag}}$ be defined by~\cref{eq.twotransfer}. For any $s \in \C$ with $\text{Re}(s) = 0$, we have
\begin{equation}\label{eq.difftransfer}
	G_{\text{DPLR}}(s) - G_{\text{Diag}}(s) = \frac{-s\frac{(-1)^{n-1} \prod_{j=1}^{n-1} (j-s)}{\prod_{j=1}^n (j+s)} \sqrt{2\ell-1} \frac{\prod_{j=0}^{\ell-2}(s-j)}{\prod_{j=1}^\ell (s+j)}}{\sqrt{2} \left(1 + s\frac{(-1)^{n-1} \prod_{j=1}^{n-1}(j-s)}{\prod_{j=1}^n (j+s)}\right)}.
\end{equation}
\end{lem}

The proof of~\Cref{lem.difftransfer} is technical and therefore deferred to~\Cref{sec:proofdifftransfer}. The idea is to expand the term $(s\mathbf{I} - \mathbf{A}_{H}^\perp + \mathbf{B}_{H}\mathbf{B}_{H}^\top/2)^{-1}$ in the expression of $ G_{\text{Diag}}$ using the Woodbury matrix identity~\cite{woodbury1950inverting}, which leads to a primary term $(s\mathbf{I} - \mathbf{A}_{H}^\perp)^{-1}$ that gets canceled with that in $G_{\text{DPLR}}$ and residual terms that are expanded by Cramer's rule. The importance of~\cref{eq.difftransfer} is that it reduces the complicated matrix inversions to elementary operations, enabling further analysis.

In~\Cref{fig:transfer}, we plot the magnitude of transfer functions $\abs{G_{\text{DPLR}}}$ and $\abs{G_{\text{Diag}}}$ when $\ell = 1$ and $n = 10, 10^2, 10^3$, and $10^4$, respectively. Note that when $\ell = 1$, $G_{\text{DPLR}}$ is independent of $n$ (see~\Cref{sec:proofdifftransfer}). 
We see that as $n$ increases, $G_{\text{Diag}}$ approaches $G_{\text{DPLR}}$ in the low-frequency domain, i.e., when $\abs{s}$ is small. 
The spikes in the plot of $\abs{G_{\text{Diag}}}$ are caused by the winding of $\big(s(-1)^{n-1} \prod_{j=1}^{n-1}(j-s)\big)\big/\big(\prod_{j=1}^n (j+s)\big)$ around the origin in~\cref{eq.difftransfer}. 
Moreover, for every $n \geq 1$, zooming into the last spike located at $\abs{s} = \Theta(n^2)$ reveals that it has a constant magnitude (see the subplots on the right in~\Cref{fig:transfer}). 
Hence, the convergence of $G_{\text{Diag}}$ to $G_{\text{DPLR}}$ is non-uniform (see~\cref{sec:noconverge}). 
Moreover, the frequency response is unstable at input frequencies $s$ near these spikes, suggesting that the S4D model is not robust to certain input perturbations (see~\cref{sec:failuremodes}).

\subsection{Input-wise convergence of the diagonal initialization}
\label{sec:converge}

Here, we apply~\Cref{lem.difftransfer} to show that for a fixed input signal $\mathbf{u}(\cdot)$, the outputs of $\Sigma_{\text{DPLR}}$ and $\Sigma_{\text{Diag}}$ converge to each other as $n \rightarrow \infty$. 
If we think of $\Sigma_{\text{DPLR}}$ and $\Sigma_{\text{Diag}}$ as linear operators on the Hilbert space $L^1(\R) \cap L^2(\R)$ that map the input $\mathbf{u}$ to the output $\mathbf{y}$, then in the language of functional analysis, we have $\Sigma_{\text{DPLR}} - \Sigma_{\text{Diag}}$ converges to zero in the weak$^*$ topology. Moreover, while the previous result~\cite{gu2022parameterization} does not have a rate of convergence, we show that the convergence is linear. In fact, the rate is sharp (see Appendix~\ref{sec:experimentconverge}). This partially explains why the S4D model matches the performance of the S4 model in practice.

\begin{thm}\label{thm.weakstarconverge}
Let $\mathbf{u}(\cdot) \in L^2(\R)$ be an input function with $\norm{\mathbf{u}}_{L^2} = 1$. Let $\mathbf{y}_{\text{DPLR}}(\cdot)$ and $\mathbf{y}_{\text{Diag}}(\cdot)$ be the outputs of $\Sigma_{\text{DPLR}}$ and $\Sigma_{\text{Diag}}$ given the input $\mathbf{u}(\cdot)$ and the initial states $\mathbf{x}(0) = \boldsymbol{0}$, respectively. For some $q > 1/2$, suppose $\abs{\hat{\mathbf{u}}(s)} = \mathcal{O}(\abs{s}^{-q})$ as $\abs{s} \rightarrow \infty$. Then, we have $\norm{\mathbf{y}_{\text{DPLR}} - \mathbf{y}_{\text{Diag}}}_{L^2} = \mathcal{O}\big(n^{-1}\big)\sqrt{\ell}$ as $n \rightarrow \infty$, where the constant in the $\mathcal{O}$-notation only depends on $q$ and the constant in $\hat{\mathbf{x}}(s) = \mathcal{O}(\abs{s}^{-q})$. The constant does not depend on $q$ if we restrict $q \in [q',\infty)$ for a fixed $q' > 1/2$.
\end{thm}

The proof is deferred to Appendix~\ref{sec:proofconverge}. Since the Fourier transform interchanges smoothness and decay, what~\Cref{thm.weakstarconverge} says is that under a mild assumption that $\mathbf{u}(\cdot)$ is sufficiently smooth, the output of the diagonal system converges linearly to that of the DPLR system as $n \rightarrow \infty$.

\subsection{System-wise divergence of the diagonal initialization}
\label{sec:noconverge}

We showed in section~\ref{sec:converge}  that the outputs of the diagonal system and the DPLR system converge for a fixed smooth input $\mathbf{u}(\cdot)$. 
Two questions naturally arise from this result: 
\begin{enumerate}
    \item Can we remove the smoothness assumption on the input?
    \item Is the convergence uniform across all input signals?
\end{enumerate}
Unfortunately, the answers to both questions are negative: 
the outputs of the two systems diverge given the unit impulse (i.e., the Dirac delta function) as the input (see~\Cref{sec:experimentconverge}); moreover, the two systems are proven to diverge from each other in the operator norm topology.

\begin{thm}\label{thm.noconverge}
The function $G_{\text{DPLR}}(s) - G_{\text{Diag}}(s)$ does not converge to zero uniformly on the imaginary axis as $n \rightarrow \infty$. In particular, for every $n \geq 1$, there exists an input signal $\mathbf{u}_n(\cdot) \in L^1(\R) \cap L^2(\R)$ such that if we let $\mathbf{y}_{n,\text{DPLR}}$ and $\mathbf{y}_{n,\text{Diag}}$ be the outputs of $\Sigma_{\text{DPLR}}$ and $\Sigma_{\text{Diag}}$ of degree $n$, respectively, then we have $\|\mathbf{y}_{n,\text{DPLR}} - \mathbf{y}_{n,\text{Diag}}\|_{L^2}$ does not converge to $0$ as $n \rightarrow \infty$.
\end{thm}

The construction of $\mathbf{u}_n(\cdot)$ can be made explicit, as we show in the next subsection. We defer the proof to~\Cref{sec:proofnoconverge}. The strategy is to carefully analyze the location and magnitude of the last spike in the plot of $\abs{G_{\text{Diag}}}$ (see~\Cref{fig:transfer}) and show that it does not vanish as $n \rightarrow \infty$. In summary, while a sufficiently large S4D model mimics its S4 alternative on a fixed smooth input, when we predetermine a size $n$, they inevitably disagree, by a large amount, on some inputs. Moreover, the smoothness condition is important: for a non-smooth input, the outputs of the two systems can diverge even if we allow $n$ to be arbitrarily large (see Appendix~\ref{sec:experimentconverge}).

\subsection{Implication of the theory: non-robustness of the diagonal initialization}
\label{sec:failuremodes}

The analysis of $G_{\text{DPLR}}$ and $G_{\text{Diag}}$ in this section suggests the following caveat: while the S4 and the S4D models tend to pertain similar behaviors as $n$ gets large, the diagonal initialization scheme used by the S4D model is less robust to perturbations in the frequency domain (see~\Cref{fig:transfer}). 
In particular, by~\cref{eq.transferfourier}, for a fixed state size $n$, input signals with frequency modes dense at the spikes of the plot of $\abs{G_{\text{Diag}}}$ are harder to process for the SSM. 
In turn, the S4D model is unstable near these modes. This does not happen with the S4 model. Our observation suggests that instead of replacing the ill-posed diagonalization problem with a well-conditioned but distinct one (i.e., the S4D initialization), which creates a large backward error $|G_{\text{DPLR}}(s) - G_{\text{Diag}}(s)|$, one should solve the ill-posed problem using a backward stable algorithm, even if the forward error (i.e., the miscalculation of eigenvalues and eigenvectors) will be large (see~\cref{sec:S4-PTD}).

We demonstrate on a synthetic example that the S4D model, regardless of its size $n$, is not robust under input perturbation of certain frequency modes (which depend on $n$).
Our training set contains sinusoidal signals parameterized by a frequency $s$ and an amplitude $A$: 
\[
\mathbf{u}_j(t) = A_j \sin(s_jt),
\]
where $A_j \in [0,1]$ and $s_j \in S_{\text{interp}} := [0,40] \cup [60,100]$ for an interpolation problem or $s_j \in S_{\text{extrap}} := [0,80]$ for an extrapolation problem.
We sample each input function $\mathbf{u}_j(\cdot)$ uniformly on $t \in [0,10^4]$ and train an S4 model and an S4D model, respectively. 
Our goal is to learn $s$ and $A$ from the sequential input. 

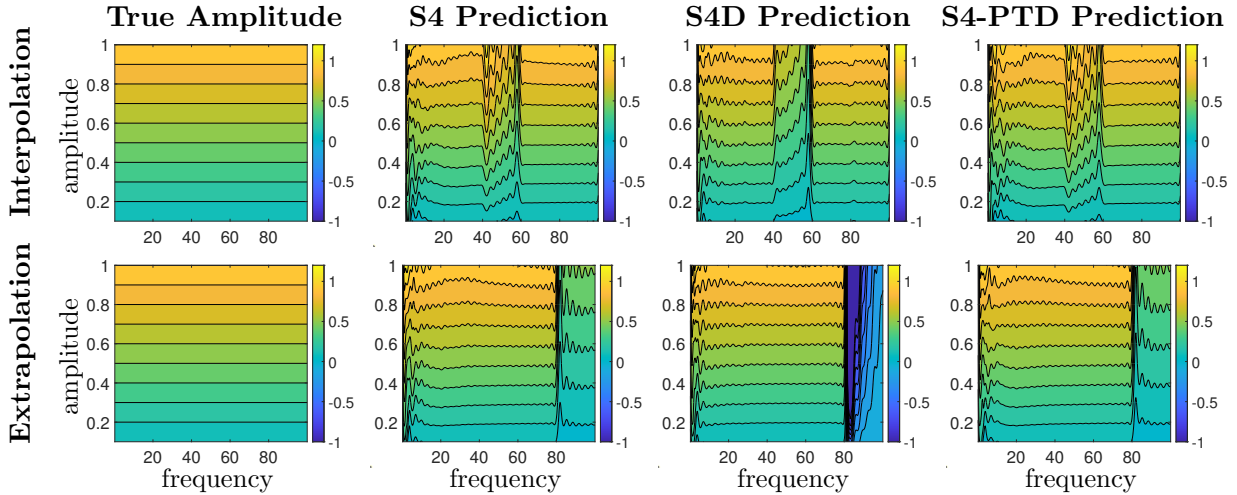
\begin{figure}
\centering
\hspace{0.8cm}
\begin{minipage}{0.22\textwidth}
\begin{overpic}[width = 1.1\textwidth]{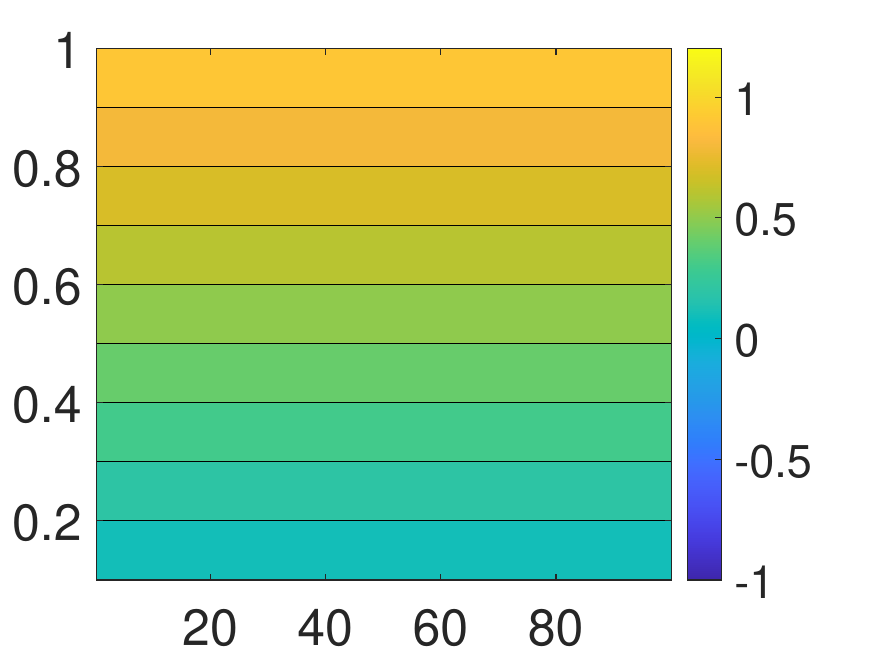}
\put(-8,18) {\rotatebox{90}{\small{amplitude}}}
\put(-25,10) {\rotatebox{90}{\textbf{Interpolation}}}
\put(8,76) {{\textbf{True Amplitude}}}
\end{overpic}
\end{minipage}
\hfill
\begin{minipage}{0.22\textwidth}
\begin{overpic}[width = 1.1\textwidth]{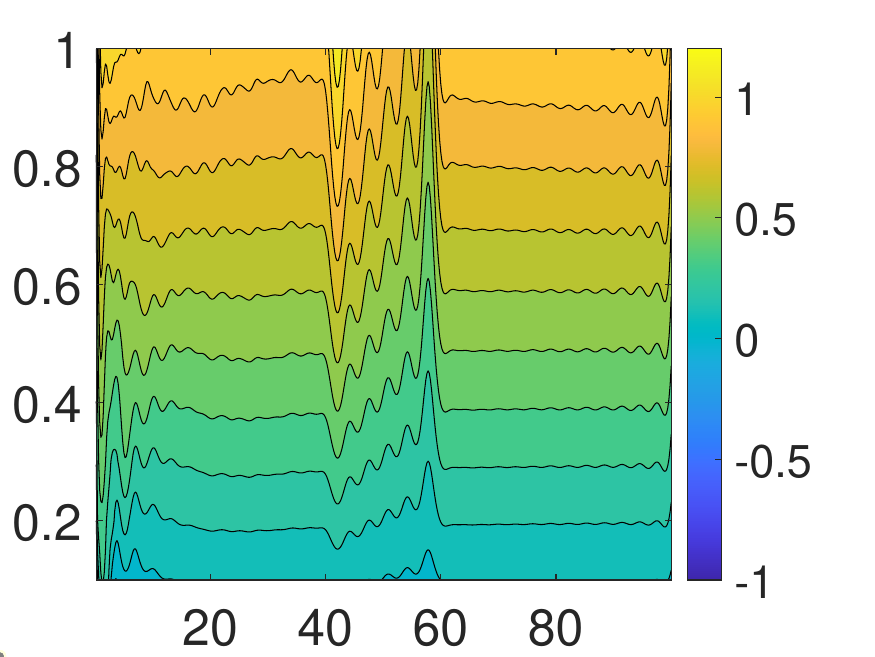}
\put(12,76) {{\textbf{S4 Prediction}}}
\end{overpic}
\end{minipage}
\hfill
\begin{minipage}{0.22\textwidth}
\begin{overpic}[width = 1.1\textwidth]{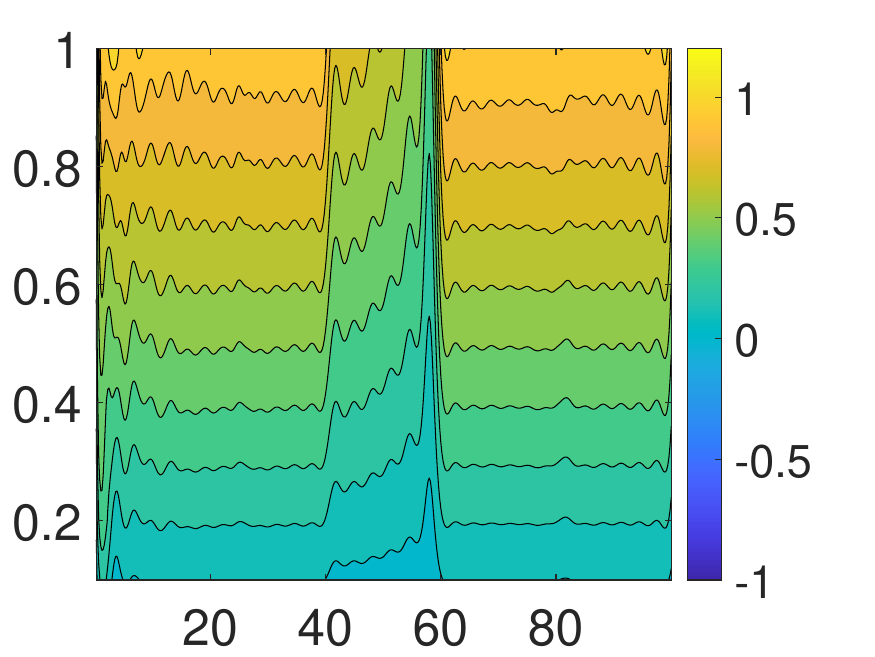}
\put(7,76) {{\textbf{S4D Prediction}}}
\end{overpic}
\end{minipage}
\hfill
\begin{minipage}{0.22\textwidth}
\begin{overpic}[width = 1.1\textwidth]{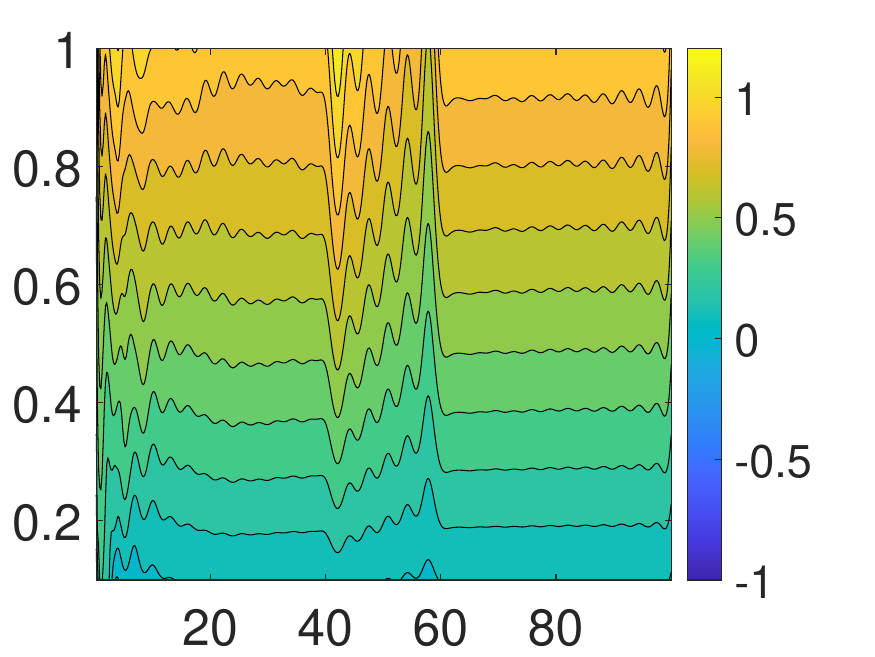}
\put(-5,76) {{\textbf{S4-PTD Prediction}}}
\end{overpic}
\end{minipage}

\hspace{0.8cm}
\begin{minipage}{0.22\textwidth}
\begin{overpic}[width = 1.1\textwidth]{./Figures/true_interp.pdf}
\put(-8,18) {\rotatebox{90}{\small{amplitude}}}
\put(27,-7) {{\small{frequency}}}
\put(-25,8) {\rotatebox{90}{\textbf{Extrapolation}}}
\end{overpic}
\end{minipage}
\hfill
\begin{minipage}{0.22\textwidth}
\begin{overpic}[width = 1.1\textwidth]{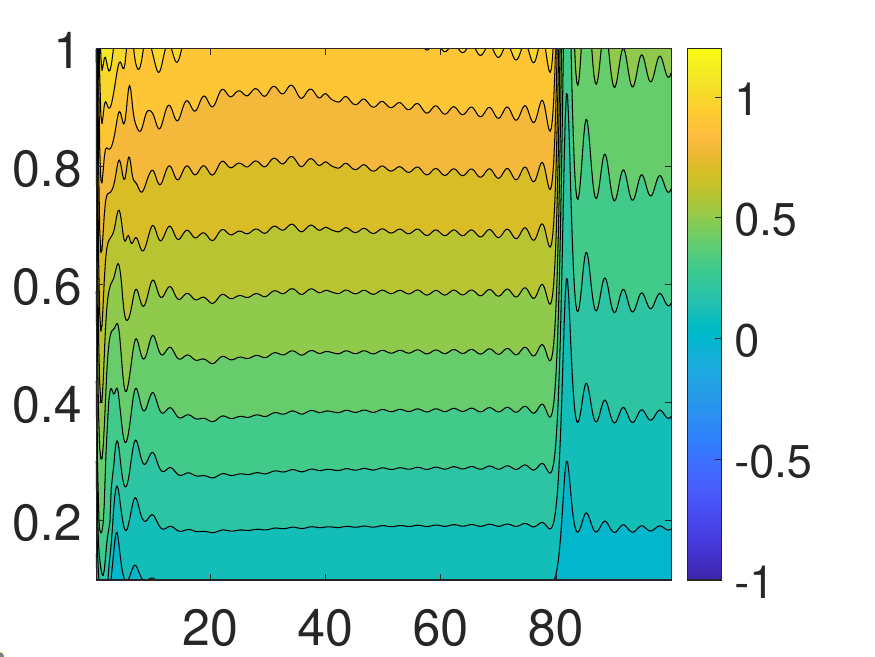}
\put(27,-7) {{\small{frequency}}}
\end{overpic}
\end{minipage}
\hfill
\begin{minipage}{0.22\textwidth}
\begin{overpic}[width = 1.1\textwidth]{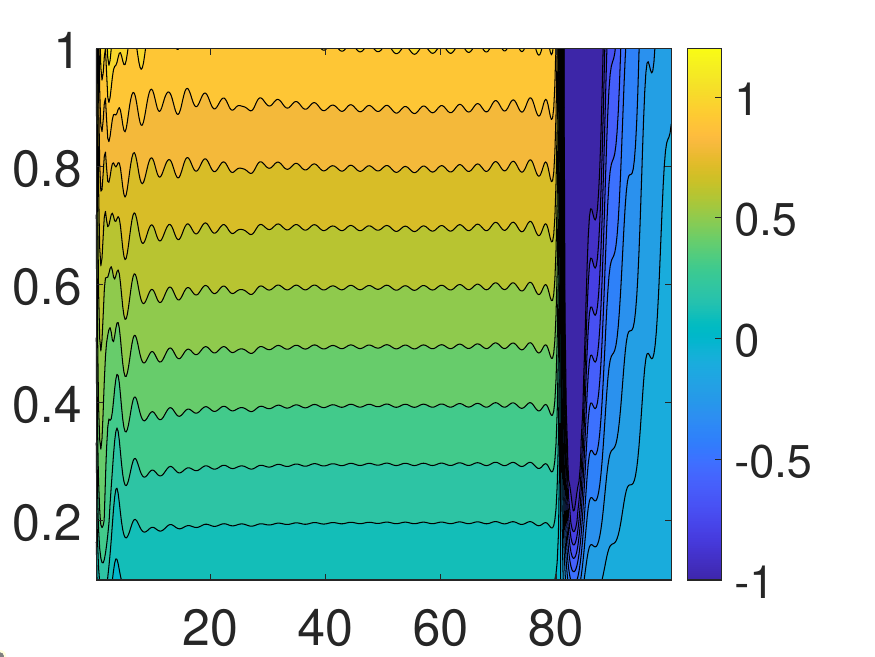}
\put(27,-7) {{\small{frequency}}}
\end{overpic}
\end{minipage}
\hfill
\begin{minipage}{0.22\textwidth}
\begin{overpic}[width = 1.1\textwidth]{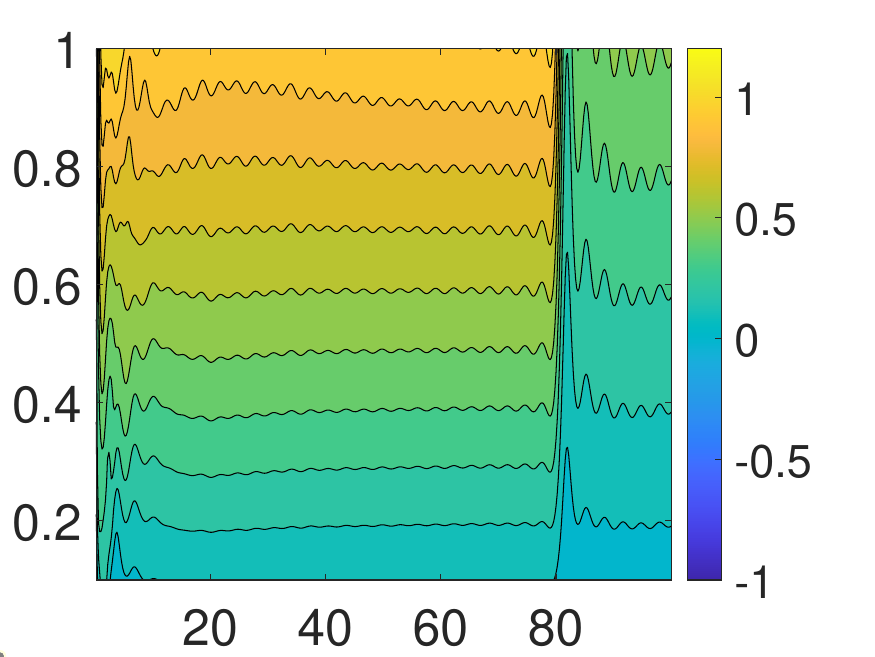}
\put(27,-7) {{\small{frequency}}}
\end{overpic}
\end{minipage}
\vspace{0.2cm}
\caption{The predicted amplitude $A$ of signals in the test set whose frequencies $s$ are sampled uniformly from $[0,100]$ and amplitudes from $[0,1]$. The top row shows the interpolation result, where the functions in the training datasets have frequencies only in $[0,40] \cup [60,100]$. The bottom row shows the extrapolation result, where the functions in the training datasets have frequencies only in $[0,80]$. The figure shows the interpolation and extrapolation results for the S4 model, the S4D model, and the S4-PTD model (see~\cref{sec:S4-PTD}). We observe that our S4-PTD model interpolates and extrapolates better than the S4D model. In particular, the S4D model is not stable around $s = 80$, where the predicted amplitude decreases to $-4$ when the true value increases from $0$ to $1$. More quantitative results for the interpolation and extrapolation errors can be found in Appendix~\ref{sec:quantitativeerr}.}
\label{fig:failuremodes}
\end{figure}

In~\Cref{fig:failuremodes}, we plot the model prediction of the amplitude $A$ over a test set of signals for which $s$ and $A$ are on a uniform grid on $[0,100] \times [0,1]$.
\Cref{fig:failuremodes} shows that while both the S4 and S4D models predict well on sampled domains (i.e., $S_{\text{interp}}$ and $S_{\text{extrap}}$), the S4 model is significantly better at interpolating and extrapolating on the unsampled domains. In particular, the S4D model suffers from an extrapolation disaster: for $s > 80$, as the true amplitude of the signal increases from $0$ to $1$, the predicted amplitude decreases monotonically with a minimum value less than $-4$. 
This happens because $\abs{G_{\text{Diag}}}$ has a spike around $\abs{s} = 83$, making the information of $s \in [0,80]$ impossible to transfer to $s \in [80,100]$. Hence, while the S4D initialization stabilizes the diagonalization process, making the computation more efficient, its underlying state-space model is unstable near certain Fourier modes, impairing its robustness (see also~\cref{sec:robustness}).

\section{Perturbing the HiPPO initialization: a new way of diagonalizing the state-space model}
\label{sec:S4-PTD}

In~\cref{sec:failuremodes}, we saw the instability of the S4D model at certain Fourier modes. 
Nevertheless, the diagonal structure of $\mathbf{A}$ is preferred over the DPLR one due to its training and inference efficiency and its adaptivity to the MIMO model (i.e., the S5 model)~\cite{smith2023simplified}. 
To avoid the instability in a diagonal model, we want to leverage the HiPPO initialization in~\cref{eq.HiPPO} instead of the one in~\cref{eq.initS4D} that discards the rank-$1$ part. 
One obvious solution is to diagonalize the HiPPO matrix $\mathbf{A}_H = \mathbf{V}_H \mathbf{\Lambda}_H \mathbf{V}_H^{-1}$ and conjugate $(\mathbf{A}_H, \mathbf{B}_H, \mathbf{C}, \mathbf{D})$ using $\mathbf{V}_H$. 
However, as shown in~\cite{gu2022parameterization}, the eigenvector matrix $\mathbf{V}_H$ is exponentially ill-conditioned with respect to $n$, making the spectral information meaningless. While the exact eigenvalues and eigenvectors of $\mathbf{A}_H$ are very ill-conditioned, since we only care about the backward error of diagonalization,
we propose the following initialization scheme. let $\mathbf{E} \in \C^{n \times n}$ be a perturbation matrix. 
We diagonalize the perturbed HiPPO matrix as
\begin{equation}
\label{eq.perturbedHiPPO}
\tilde{\mathbf{A}}_H = \mathbf{A}_H + \mathbf{E} = \tilde{\mathbf{V}}_H \tilde{\mathbf{\Lambda}}_H \tilde{\mathbf{V}}_H^{-1}.
\end{equation}
We then initialize the systems using $\Sigma_{\text{Pert}} = (\mathbf{A}_{\text{Pert}}, \mathbf{B}_{\text{Pert}}, \mathbf{C}_{\text{Pert}}, \mathbf{D}_{\text{Pert}}) = (\tilde{\boldsymbol{\Lambda}}_H, \tilde{\mathbf{V}}_H^{-1} \mathbf{B}_H, \mathbf{C}, \mathbf{D})$, where $\mathbf{C}$ and $\mathbf{D}$ are random matrices. Therefore, we approximately diagonalize the HiPPO initialization in the sense that although the diagonal entries in $\tilde{\boldsymbol\Lambda}$ do not approximate the eigenvalues of $\mathbf{A}_H$, the transfer function of $\Sigma_{\text{Pert}}$ is an approximation of that of $\Sigma_{\text{DPLR}}$ (see~\Cref{thm.perturbsys}). This is enough to guarantee a good initialization. Our proposed perturb-then-diagonalize method is not restricted to the HiPPO-LegS matrices in~\cref{eq.HiPPO}. 
This endows our method with adaptivity to any (dense) initialization scheme. 
This adaptivity was absent from the previous line of work on SSMs.

We call our model S4-PTD or S5-PTD, depending on whether the model architecture is adapted from the S4D or the S5 model, where ``PTD'' stands for ``perturb-then-diagonalize.'' 
Since our models are only different from the S4D and the S5 models in initialization, we refer interested readers to~\cite{gu2022parameterization,smith2023simplified} for a discussion of computation details and time/space complexity. 

Centered around the perturbed initialization scheme~\cref{eq.perturbedHiPPO} are two important questions:
\begin{enumerate}
    \item What is the difference between the perturbed initialization $(\mathbf{A}_{\text{Pert}}, \mathbf{B}_{\text{Pert}}, \mathbf{C}_{\text{Pert}}, \mathbf{D}_{\text{Pert}})$ and the HiPPO initialization $(\mathbf{A}_{\text{DPLR}}, \mathbf{B}_{\text{DPLR}}, \mathbf{C}_{\text{DPLR}}, \mathbf{D}_{\text{DPLR}})$?
    \item What is the condition number of $\tilde{\mathbf{V}}_H$?
The first question is important because it controls the deviation of our perturbed initialization from the successful and robust DPLR initialization.
\end{enumerate}
The second question is important because it shadows the numerical robustness of conjugating the LTI system by $\tilde{\mathbf{V}}_H$. 
Moreover, since the state vector $\mathbf{x}(t)$ is transformed by $\tilde{\mathbf{V}}_H$ via conjugation (see~\cref{sec:prelim}), a small condition number of $\tilde{\mathbf{V}}_H$ shows that its singular values are more evenly distributed.
Hence, the transformation $\tilde{\mathbf{V}}_H$ does not significantly magnify or compress $\mathbf{x}(t)$ onto some particular modes.

\subsection{Estimating the transfer function perturbation}

To study the first question, we define the transfer function of the perturbed system to be
\[
G_{\text{Pert}}(s) = \mathbf{C}_{\text{Pert}} (s\mathbf{I} - \mathbf{A}_{\text{Pert}})^{-1} \mathbf{B}_{\text{Pert}} + \mathbf{D}_{\text{Pert}}.
\]
We control the size of the transfer function perturbation by proving the following theorem.

\begin{thm}
\label{thm.perturbsys}
Assume ${\mathbf{C}_{\text{Pert}} \tilde{\mathbf{V}}_H^{-1}} = {\mathbf{C}_{\text{DPLR}} {\mathbf{V}}_H^{-1}}$ and $\mathbf{D}_{\text{Pert}} = \mathbf{D}_{\text{DPLR}}$. Suppose $\norm{\mathbf{E}}_2 \leq \epsilon$ and we normalize the matrices so that $\|{\tilde{\mathbf{V}}_H \mathbf{B}_\text{Pert}}\| = \|{{\mathbf{V}}_H \mathbf{B}_\text{DPLR}}\| = \|{\mathbf{C}_{\text{Pert}} \tilde{\mathbf{V}}_H^{-1}}\| = \|{\mathbf{C}_{\text{DPLR}} {\mathbf{V}}_H^{-1}}\| = 1$. For any $s$ on the imaginary axis, we have
\[
\abs{G_{\text{Pert}}(s) - G_{\text{DPLR}}(s)} = (2\ln(n)+4)\epsilon + \mathcal{O}(\sqrt{\log(n)}\epsilon^2).
\]
\end{thm}

While our perturb-then-diagonalize method works for a general initialization and a bound on the transfer function error can always be established, the proof of~\Cref{thm.perturbsys} leverages the structure of HiPPO matrices to improve this bound.
The error in~\Cref{thm.perturbsys} is the uniform error on the imaginary axis. 
Using H\"older's inequality, for any bounded and integrable input function $\mathbf{u}(\cdot)$, if $\mathbf{y}_{\text{Pert}}$ and $\mathbf{y}_{\text{DPLR}}$ are the outputs of $\Sigma_{\text{Pert}}$ and $\Sigma_{\text{DPLR}}$, respectively, then we have
\begin{equation}
\begin{aligned}
&\|\mathbf{y}_{\text{Pert}} - \mathbf{y}_{\text{DPLR}}\|_{L^2} = \|\hat{\mathbf{y}}_{\text{Pert}} - \hat{\mathbf{y}}_{\text{DPLR}}\|_{L^2} =  \|\hat{\mathbf{x}}(s) (G_{\text{Pert}}(is) - G_{\text{DPLR}}(is))\|_{L^2} \\
&\qquad\leq \|\hat{\mathbf{x}}(s)\|_{L^2} \|G_{\text{Pert}}(is) \!-\! G_{\text{DPLR}}(is)\|_{L^\infty} \!\leq \|\mathbf{x}\|_{L^2} \big((2\ln(n)\!+\!4)\epsilon \!+\! \mathcal{O}(\sqrt{\log(n)}\epsilon^2)\big),
\end{aligned}
\end{equation}
where the first and the last steps follow from Parseval's identity. 
Hence,~\Cref{thm.perturbsys} gives us an upper bound on the distance between $\Sigma_{\text{Pert}}$ and $\Sigma_{\text{DPLR}}$ in the operator norm topology. 
The theorem states that the error made by the perturbation is linear in the size of the perturbation. 
Moreover, the error depends only logarithmically on the dimension $n$ of the state space. 

\subsection{Perturbed initialization as an optimization problem}\label{sec:optimizeE}

Next, we consider the conditioning of $\tilde{\mathbf{V}}_H$, which affects the accuracy of computing $\tilde{\mathbf{V}}_H^{-1} \mathbf{B}_{\text{Pert}}$ and the scaling ratio of the states in $\mathbf{x}(\cdot)$ (see Appendix~\ref{sec:morebackground}). This problem was studied by some numerical analysts~\cite{davies2008approximate,davies2009perturbations,banks2021gaussian,banks2022pseudospectral}. The following theorem can be derived from their results on perturbing a general square matrix by a Ginibre matrix.
\begin{thm}\label{thm.averagecond}
	Given any matrix $\mathbf{A} \in \C^{n \times n}$, perturbation size $\epsilon \in (0,1)$, and spectral radius $R > 0$. Let $\mathbf{G}_n \in \C^{n \times n}$ be the Ginibre matrix and let $\Omega$ be the event that the spectrum of $\mathbf{A} + \epsilon \mathbf{G}_n$ is contained in $D_R(0)$, the disk centered at zero of radius $R$. Then, we have
	\[
	\mathbb{E} \left[\kappa_{\text{eig}}(\mathbf{A} + \epsilon \mathbf{G}_n)^2 \middle| \Omega \right] \leq \norm{\mathbf{A}}^2 \frac{R^2 n^3}{\epsilon^2 \mathbb{P}(\Omega)},
	\]
	where the eigenvector condition number of $\mathbf{A} + \epsilon \mathbf{G}_n$ is defined by
	\[
	\kappa_{\text{eig}}(\mathbf{A} + \epsilon \mathbf{G}_n) = \inf\{\kappa(\tilde{\mathbf{V}}) \mid \mathbf{A} + \epsilon \mathbf{G}_n = \tilde{\mathbf{V}} \tilde{\boldsymbol{\Lambda}} \tilde{\mathbf{V}}^{-1}, \tilde{\boldsymbol{\Lambda}} \text{ diagonal}\}.
	\]
\end{thm}
\Cref{thm.averagecond} states that we can get around the exponentially growing condition number of $\tilde{\mathbf{V}}_H$ by adding a small Gaussian perturbation to $\mathbf{A}_H$, which justifies our S4-PTD and S5-PTD models. While it shows an upper bound on the eigenvector condition number of the perturbed HiPPO matrix, in most circumstances, perturbation by the Ginibre matrix does not give us the smallest eigenvector condition number.
The following theorem
provides a deterministic estimate of the eigenvector condition number for the ``best perturbation scheme.''

\begin{thm}[{\cite[Thm.~1.1.]{banks2021gaussian}}]\label{thm.bestcond}
Given any $\mathbf{A} \in \C^{n \times n}$ and $\epsilon \in (0,1)$, there exists a matrix $\mathbf{E} \in \C^{n \times n}$ with $\norm{\mathbf{E}} \leq \epsilon$ and an eigenvector matrix $\tilde{\mathbf{V}}$ of $\mathbf{A} + \mathbf{E}$ such that
\[
\kappa(\tilde{\mathbf{V}}) \leq 4n^{3/2}\big(1+\epsilon^{-1}\norm{\mathbf{A}}\big).
\]
\end{thm}

\Cref{thm.bestcond} shows the promise of finding a good perturbation matrix to reduce the eigenvector condition number.
In this paper, we propose to compute $\mathbf{E}$ by solving the following optimization problem with a soft constraint:
\begin{equation}
\label{eq.optobjective}
\text{minimize } \Phi(\mathbf{E}) = \kappa(\tilde{\mathbf{V}}_H) + \gamma \|\mathbf{E}\| \qquad \text{s.t.} \qquad \mathbf{A}_H + \mathbf{E} = \tilde{\mathbf{V}}_H \tilde{\boldsymbol{\Lambda}} \tilde{\mathbf{V}}_H^{-1},\quad \tilde{\boldsymbol{\Lambda}} \text{ diagonal},
\end{equation}
where $\gamma > 0$ is a hyperparameter that controls the trade-off between $\kappa(\tilde{\mathbf{V}}_H)$ and $\|\mathbf{E}\|$. We implement a solver to this optimization problem using gradient descent.
As $\gamma$ increases, it is harder to recover the original states $\mathbf{x}(\cdot)$ from the transformed states $\tilde{\mathbf{V}}_H \mathbf{x}(\cdot)$ because $\kappa(\tilde{\mathbf{V}}_H)$ increases, but $\|\mathbf{E}\|$ decreases, resulting in a more robust SSM that is closer to the flawless HiPPO initialization.

\section{Empirical evaluation and discussion}
\label{sec:experiments}

In this section, we present empirical evaluations of our proposed S4-PTD and S5-PTD models. 
In~\cref{sec:LRA} we compare the performance of our full model with the existing ones in the Long Range Arena (LRA).
In~\cref{sec:robustness}, we perform a sensitivity analysis using the CIFAR-10 dataset to provide real-world evidence that our perturbed initialization scheme is more robust than the one in the S4D/S5 model. 
Finally, in~\cref{sec:ablation}, we study the relationship between the size of the perturbation matrix $\mathbf{E}$ and the performance of our models.

\subsection{Performance in the Long-Range Arena}
\label{sec:LRA}

The LRA benchmark comprises six tasks with sequential data~\cite{tay2020long}. 
This collection, with its sequence lengths ranging from $1024$ to $16000$, is designed to measure the model's capability of processing the long-range inputs. 
We train an S4-PTD model and an S5-PTD model to learn these tasks, respectively. 
We adopt the same SSM architectures, and thus the same number of parameters, from the original S4D~\cite{gu2022parameterization} and S5 papers~\cite{smith2023simplified}. Results are reported in~\Cref{tab:LRA}, along with the accuracies of other sequential models, including the Liquid-S4 model which is built upon S4~\cite{hasani2022liquid}.
We report details of hyperparameters in~\Cref{sec:experimentdetail}. 
While the perturbation matrix $\mathbf{E}$ is also tunable, we restrict its size to be less than $10\%$ of that of the HiPPO matrix $\mathbf{A}_H$, promoting the worst-case robustness of our model (see~\cref{sec:robustness}). 
We note that the S4-PTD model outperforms the S4D model (and even the S4 model with the DPLR structure for most tasks), while the S5-PTD model matches the performance of the S5 model.

\begin{table}
\centering
\begin{tabular}{c c c c c c c c}
\specialrule{.1em}{.05em}{.05em}
{Model} & \!\!\!\texttt{ListOps}\!\!\! & \texttt{Text} & \!\!\!\texttt{Retrieval}\!\!\! & \texttt{Image} & \!\!\!\!\texttt{Pathfinder}\!\!\!\! & \texttt{Path-X} & \!\!\!Avg. \\
\hline
Transformer & $36.37$ & $64.27$ & $57.56$ & ${{42.44}}$ & $71.40$ & \ding{55} & $53.66$  \\
Luna-256 & $37.25$ & $64.57$ & $79.29$ & ${{47.38}}$ & $77.72$ & \ding{55} & $59.37$  \\
H-Trans.-1D & $49.53$ & $78.69$ & $63.99$ & ${{46.05}}$ & $68.78$ & \ding{55} & $61.41$  \\
CCNN & $43.60$ & $84.08$ & \ding{55} & ${{88.90}}$ & $91.51$ & \ding{55} & $68.02$  \\
\hline
S4 & $59.60$ & $86.82$ & $90.90$ & ${{88.65}}$ & $94.20$ & $96.35$ & $86.09$  \\
Liquid-S4 & $\underline{\boldsymbol{62.75}}$ & $\underline{89.02}$ & $\underline{91.20}$ & $\underline{\boldsymbol{89.50}}$ & $\underline{94.80}$ & $\underline{96.66}$ & $\underline{87.32}$  \\
\hline
S4D & $60.47$ & $86.18$ & $89.46$ & $88.19$ & $93.06$ & $91.95$ & $84.89$ \\
S4-PTD (ours) & $\underline{60.65}$ & $\underline{88.32}$ & $\underline{91.07}$ & $\underline{88.27}$ & $\underline{94.79}$ & $\underline{96.39}$ & $\underline{86.58}$ \\
\hline
S5 & $62.15$ & $89.31$ & $91.40$ & $\underline{88.00}$ & $95.33$ & $\underline{\boldsymbol{98.58}}$ & $87.46$ \\
S5-PTD (ours) & $\underline{\boldsymbol{62.75}}$ & $\underline{\boldsymbol{89.41}}$ & $\underline{\boldsymbol{91.51}}$ & $87.92$ & $\underline{\boldsymbol{95.54}}$ & $98.52$ & $\underline{\boldsymbol{87.61}}$ \\
\specialrule{.1em}{.05em}{.05em}
\end{tabular}
\caption{Test accuracies on LRA, where \ding{55} means the model isn't outperforming random guessing. We use the boldface number to indicate the highest test accuracy among all models for each task. We use the underlined number to indicate the highest test accuracy within the comparable group.}
\label{tab:LRA}
\vspace{-0.2cm}
\end{table}

\subsection{Robustness of our perturbed model over the diagonal model}
\label{sec:robustness}

\begin{figure}[!t]
\centering
\begin{subfigure}{.32\textwidth}
\begin{overpic}[width=.93\textwidth]{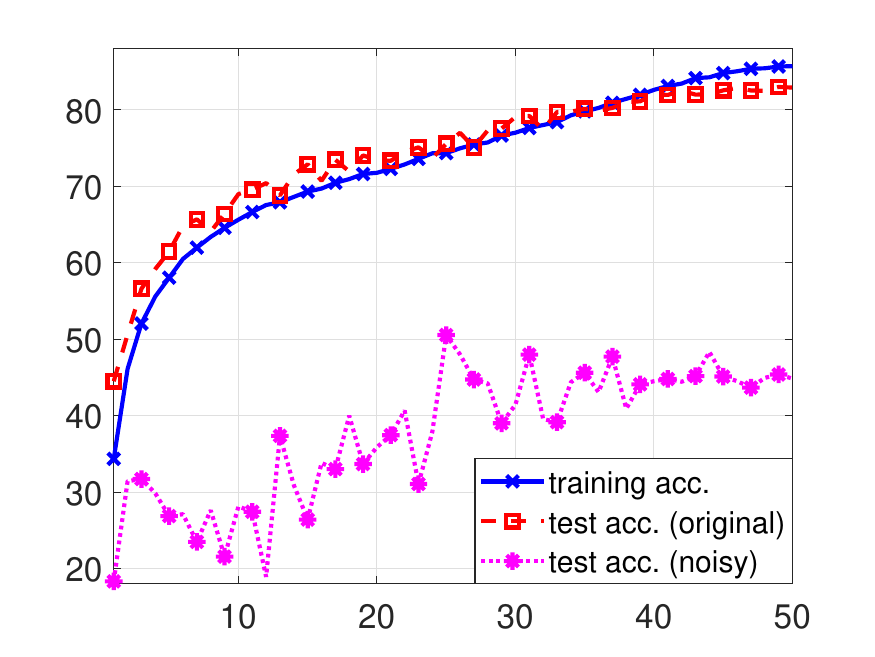}
\put(41,-4) {epochs}
\put(0,22) {\rotatebox{90}{accuracy}}
\end{overpic} 
\vspace{0.2cm}
\subcaption{Accuracies for S4D}
\label{fig:expS4D}
\end{subfigure}
\hspace{-0.3cm}
\begin{subfigure}{.32\textwidth}
\begin{overpic}[width=.93\textwidth]{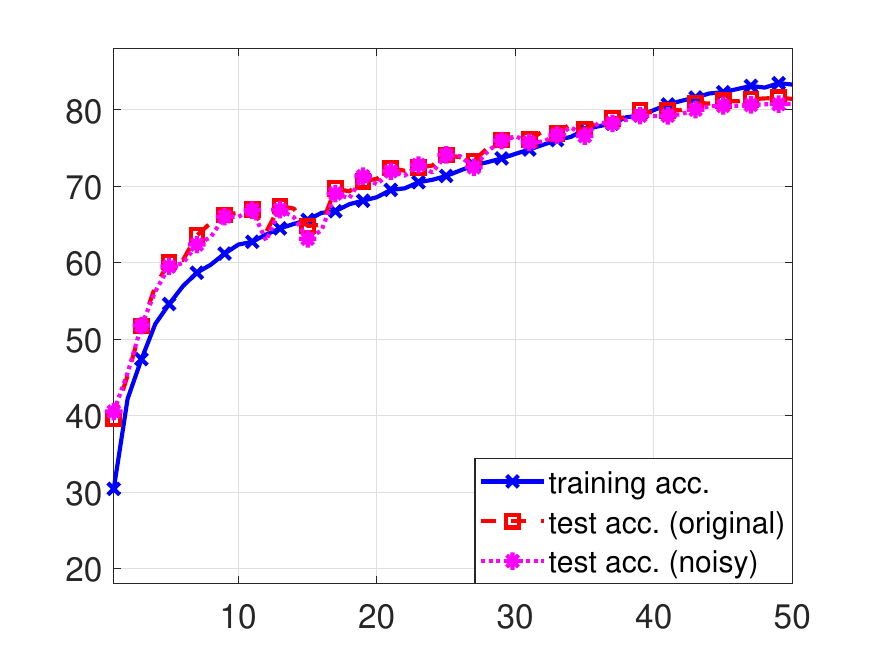}
\put(41,-4) {epochs}
\put(0,22) {\rotatebox{90}{accuracy}}
\end{overpic} 
\vspace{0.2cm}
\subcaption{Accuracies for S4-PTD}
\label{fig:expS4P}
\end{subfigure}
\begin{subfigure}{.32\textwidth}
\begin{overpic}[width=0.93\textwidth]{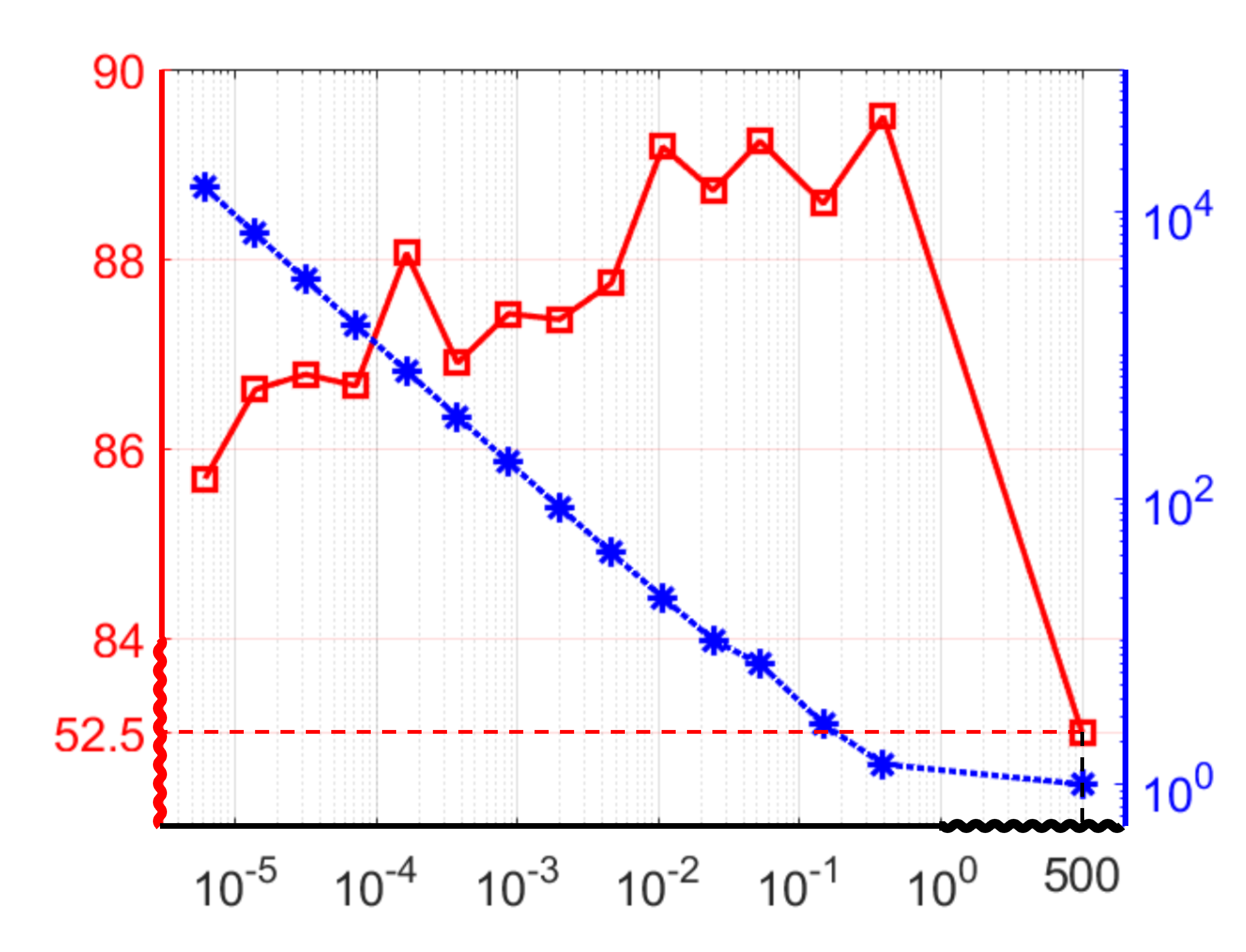}
\put(32,-3.5) {\small{$\|\mathbf{E}\| / \|\mathbf{A}_H\|$}}
\put(0,28) {\small{\rotatebox{90}{\textcolor{red}{accuracy}}}}
\put(98,50.5) {\small{\rotatebox{270}{\textcolor{blue}{$\kappa(\tilde{\mathbf{V}}_H)$}}}}
\end{overpic}
\vspace{0.2cm}
\subcaption{Ablation study}
\label{fig:expablation}
\end{subfigure}
\vspace{-0.1cm}
\caption{(\subref{fig:expS4D}) and (\subref{fig:expS4P}): the training and test accuracies of the S4D model and the S4-PTD model on contaminated and uncontaminated CIFAR-10 dataset (see~\cref{sec:robustness}). (\subref{fig:expablation}): The effect of the perturbation size on the accuracy (shown in red) of the S4-PTD model and the eigenvector condition number (shown in blue) of the perturbed HiPPO matrix (see~\cref{sec:ablation}).
}
\label{fig:experiments}
\end{figure}
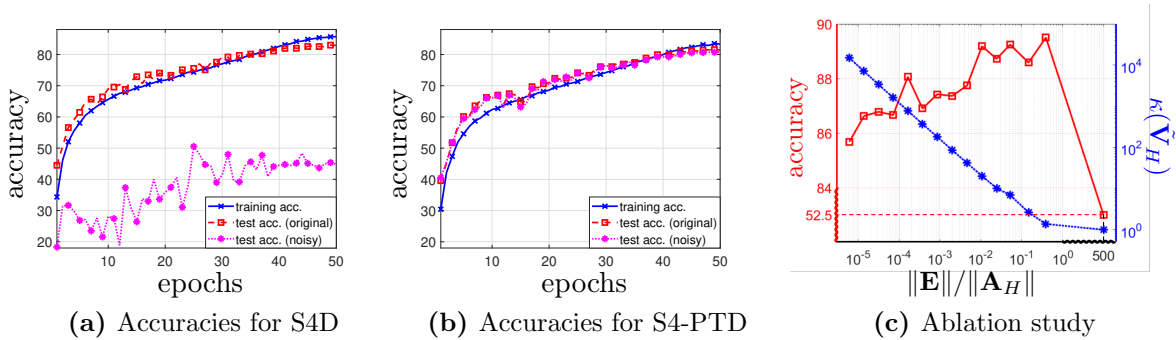

In~\cref{sec:failuremodes}, we see the instability of the S4D model using a synthetic example. 
Here, we demonstrate its practical implication regarding the robustness of the model. 
We train an S4D model and an S4-PTD model (with $\|\mathbf{E}\| / \|\mathbf{A}_H\| \approx 10^{-1}$) to learn the sCIFAR task, where the images in the CIFAR-10 dataset~\cite{krizhevsky2009learning} are flattened into sequences of pixels. 
We test the two models against two different test sets: one is taken from the original CIFAR-10 dataset while the other one is contaminated by $10\%$ of sinusoidal noises whose frequencies are located near the spikes of $\abs{G_{\text{Diag}}}$. 
We plot the training and test accuracies of the two models in~\Cref{fig:experiments}\subref{fig:expS4D}~and~\subref{fig:expS4P}. 
Whereas the two models both achieve high accuracies on the uncontaminated test set, the S4D model does not generalize to the noisy dataset as the S4-PTD model does. 
That is, the S4D model is not robust to these noises. 
In comparison, since the S4-PTD initialization is uniformly close to the S4 initialization (see~\Cref{thm.perturbsys}) when $\|\mathbf{E}\|$ is small, the S4-PTD model is robust to noises with any mode. 
We remark, nevertheless, that the noises in this experiment are the ``worst-case'' noises and intentionally made to fail the S4D model; in practice, the distribution of sensitive modes of S4D in the frequency domain gets sparser as $n$ increases (see~\Cref{fig:transfer}), which improves its ``average-case'' robustness.

\subsection{Ablation study of our model}
\label{sec:ablation}

As mentioned in~\cref{sec:S4-PTD}, the size of the perturbation plays a key role in the performance of our S4-PTD and S5-PTD models. 
When $\mathbf{E} = 0$, the eigenvector condition number of $\mathbf{A}_H$ is exponential in $n$, making it numerically impossible to diagonalize when $n$ is moderately large. On the other hand, when $\mathbf{E}$ overshadows $\mathbf{A}_H$, the initialization scheme becomes a random one, often leading to poor performance~\cite{gu2021combining}. 
In this section, we train an S4-PTD model to learn the sequential CIFAR (sCIFAR) task. 
We control the size of the perturbation $\|\mathbf{E}\|$ by changing the hyperparameter $\gamma$ in the optimization problem~\cref{eq.optobjective}. 
For each perturbation matrix $\mathbf{E}$, we then initialize our S4-PTD model by diagonalizing $\mathbf{A}_H + \mathbf{E}$. 
In~\Cref{fig:experiments}\subref{fig:expablation}, we plot (in red) the test accuracies with respect to different perturbation sizes. 
We see that our S4-PTD model achieves its best performance when the ratio between the perturbation size and the size of the HiPPO matrix is between $10^{-2}$ and $1$, while the accuracy drops when this ratio gets too small or too large. 
This aligns with our expectations. 
In addition, the (blue) curve of the eigenvector condition number admits a straight-line pattern with a slope of roughly $-1$, corroborating the factor $\epsilon^{-1}$ in~\Cref{thm.bestcond}.

%% file: supplement.tex
\newpage 

\appendix

\section*{Appendix}

The Appendix is organized as follows. 
In~\Cref{sec:morebackgroundill}, we survey the background of ill-posed problems, including conditioning, stability, and backward and forward errors. In~\Cref{sec:morebackground}, we provide more background information on LTI systems, including the derivation of the transfer function, system diagonalization, and system discretization, which provides insights into our theory and models. 
In~\Cref{sec:proofdifftransfer}, we prove~\Cref{lem.difftransfer} on the difference between the two transfer functions. 
Using this result, we prove~\Cref{thm.noconverge} and~\ref{thm.weakstarconverge} on the uniform divergence and the input-wise convergence in~\Cref{sec:proofnoconverge} and~\ref{sec:proofconverge}, respectively. 
We then present in~\Cref{sec:experimentconverge} some numerical experiments corroborating these two theorems and in~\Cref{sec:quantitativeerr} some plots to show the quantitative interpolation and extrapolation errors in our synthetic example in~\cref{sec:failuremodes}.
In~\Cref{sec:proofperturb}, we prove the results in~\cref{sec:S4-PTD} that are related to perturbing the HiPPO matrix, which are then verified in~\Cref{sec:experimentperturb} by a numerical experiment. 
Finally, in~\Cref{sec:experimentdetail} we give the details of our experiments in~\cref{sec:experiments}.

\section{More background information of ill-posed problems}\label{sec:morebackgroundill}

To make the phrase ``ill-posed" precise, we need to introduce the idea of condition numbers. The conditioning is a property of a problem and it does not depend on the algorithm that we use. Abstractly, we let the problem space $\mathcal{X}$ and the solution space $\mathcal{Y}$ be two normed vector spaces with the norms $\|\cdot\|_{\mathcal{X}}$ and $\|\cdot\|_{\mathcal{Y}}$, respectively. Each element in $x \in \mathcal{X}$ is considered as an instance of the problem and its solution in the solution space $\mathcal{Y}$ is defined by a map $f: \mathcal{X} \rightarrow \mathcal{Y}$. For example, if we want to solve a system $\mathbf{A}\mathbf{x} = \mathbf{b}_0$ with different matrices $\mathbf{A}$ and a fixed vector $\mathbf{b}_0$, then we can make $\mathcal{X}$ the space of $n$-by-$n$ matrices and $\mathcal{Y}$ the space of vectors of length $n$. In that case, we have $f(\mathbf{A}) = \mathbf{A}^{-1}\mathbf{b}_0$.\footnote{Of course, this function is not defined at singular matrices, but since they form a Lebesgue null set, $f$ is still defined almost everywhere.} Likewise, consider the problem of finding eigenvalues. We can make $\mathcal{X}$ and $\mathcal{Y}$ both equal to the space of $n$-by-$n$ matrices and $f(\mathbf{A}) = \boldsymbol{\Lambda}$, the eigenvalue matrix of $\mathbf{A}$. Now, given an instance $x \in \mathcal{X}$, we define the (absolute) condition number of problem $f$ at $x$ to be
\[
    \kappa(x;f) = \lim_{\epsilon \rightarrow 0} \sup_{\|\delta x\|_\mathcal{X} \leq \epsilon} \frac{\|f(x) - f(x+\delta x)\|_{\mathcal{Y}}}{\|\delta x\|_{\mathcal{X}}}.
\]
Intuitively, a large condition number means that if we perturb the problem by a little bit, then the solution may become drastically different. Hence, in general, we do not expect that a solution of an ill-conditioned problem can be found accurately using floating-point arithmetic because a small rounding error has a large effect on the computed solution.

Unlike the conditioning of a problem $x$, stability is a property of an algorithm. Let $\tilde{f}:\mathcal{X} \rightarrow \mathcal{Y}$ be an algorithm that solves $f$. We are particularly interested in the ``backward stability" of $\tilde{f}$. That is, if for any $x \in \mathcal{X}$, there exists an element $\tilde{x} \in \mathcal{X}$ so that $\tilde{f}(x) = f(\tilde{x})$ and 
\[
    E_b(x) = \frac{\|x - \tilde{x}\|_{\mathcal{X}}}{\|x\|_{\mathcal{X}}}
\]
is small, then we say that $\tilde{f}$ is backward stable. Intuitively, this is saying that our algorithm is computing the solution to a nearby problem $\tilde{x}$. Note that this is different from saying that 
\[
    E_f(x) = \frac{\|f(x) - \tilde{f}(x)\|_\mathcal{Y}}{\|f(x)\|_{\mathcal{Y}}}
\]
is small. This error measures how accurately we solved our problem. The error $E_f$ is called a forward error, while $E_b$ is called a backward error. We can control the forward error using the backward error, and this bound is established through the condition number of the problem. (Intuitively, this says that if a problem is well-conditioned, then a small perturbation to the problem does not change the solution by too much. Hence, a small backward error leads to a small forward error.) The advantage of studying backward stability is two-fold. First, the backward error is decoupled from the conditioning of the problem. Hence, backward stable algorithms are much more common than forward stable algorithms, because in many cases, the ill-conditioned problems essentially prevent an algorithm from being forward stable. On the other hand, if an algorithm is backward stable, then we know that its forward error must be small on well-conditioned problems.

\begin{figure}
\centering
    \begin{overpic}[width=0.7\textwidth]{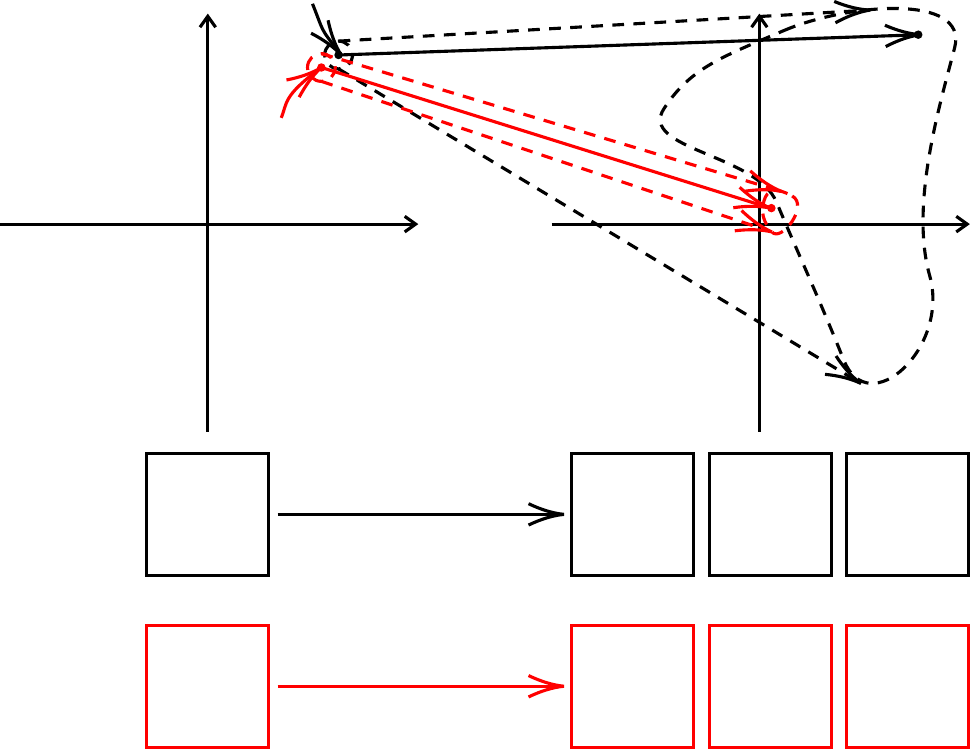}
        \put(10.5,83) {\textbf{Problem Space}}
        \put(67.5,83) {\textbf{Solution Space}}
        \put(16,78) {\small{An ill-conditioned problem}}
        \put(26,62) {\small{\textcolor{red}{A close but}}}
        \put(23,59) {\small{\textcolor{red}{well-conditioned}}}
        \put(28,56) {\small{\textcolor{red}{problem}}}
        \put(19,23.2) {$\mathbf{A}_H$}
        \put(19,5) {\textcolor{red}{$\tilde{\mathbf{A}}_H$}}
        \put(20,14) {\rotatebox{90}{$\approx$}}
        \put(63,23.2) {$\mathbf{V}_H$}
        \put(63,5) {\textcolor{red}{$\tilde{\mathbf{V}}_H$}}
        \put(63,14) {\rotatebox{90}{$\neq$}}
        \put(77,23.2) {$\boldsymbol{\Lambda}_H$}
        \put(77,5) {\textcolor{red}{$\tilde{\boldsymbol{\Lambda}}_H$}}
        \put(77.5,14) {\rotatebox{90}{$\neq$}}
        \put(90,23.2) {$\mathbf{V}_H^{-1}$}
        \put(90,5) {\textcolor{red}{$\tilde{\mathbf{V}}_H^{-1}$}}
        \put(92,14) {\rotatebox{90}{$\neq$}}
    \end{overpic}
    \caption{An illustration of the perturbation of an ill-posed problem.}
    \label{fig:ptdfigure}
\end{figure}
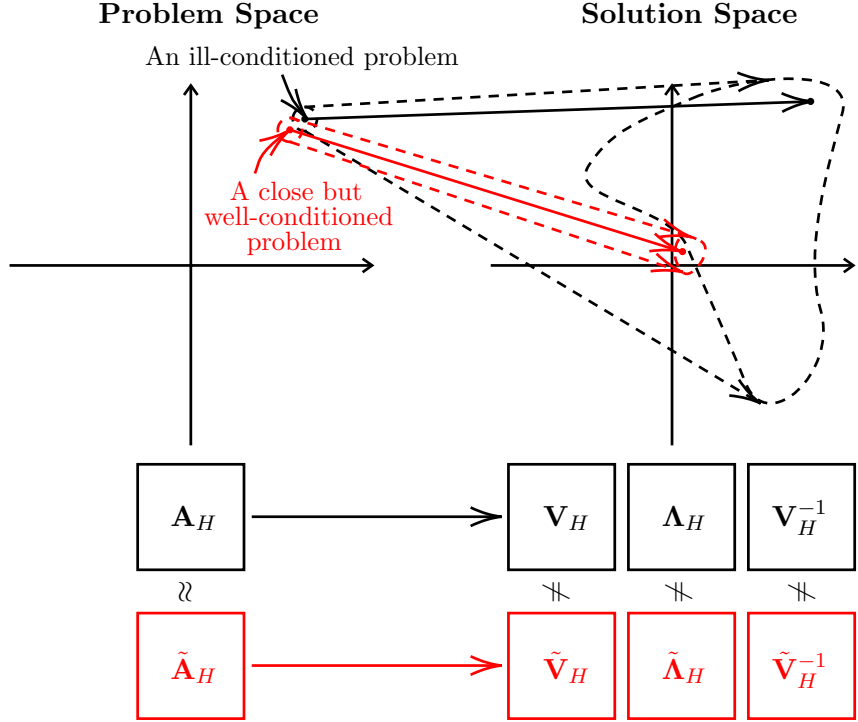

In our paper, we consider the case where we are forced to solve an ill-conditioned problem $x$. We propose to use a backward stable algorithm $\tilde{f}$ to solve it. Since the problem is ill-conditioned, we do not have that $f(x)$ is close to $\tilde{f}(x)$, i.e., we cannot find the solutions accurately. However, we know that $\tilde{f}(x)$ is the solution to $\tilde{x}$, where $x \approx \tilde{x}$. In many machine learning applications, this is enough to guarantee an acceptable solution. (See~\Cref{fig:ptdfigure}.)


\section{More background information of state-space models}\label{sec:morebackground}

Recall that an LTI system $\Sigma = (\mathbf{A}, \mathbf{B}, \mathbf{C}, \mathbf{D})$ is given by
\begin{equation}\label{eq.LTIappend}
    \begin{aligned}
        \mathbf{x}'(t) &= \mathbf{A} \mathbf{x}(t) + \mathbf{B} \mathbf{u}(t), \\
        \mathbf{y}(t) &= \mathbf{C} \mathbf{x}(t) + \mathbf{D} \mathbf{u}(t).
    \end{aligned}
\end{equation}
Assume the initial condition is given by $\mathbf{x}(0) = \boldsymbol{0}$ and the input function $\mathbf{u}(\cdot)$ is bounded and integrable. Suppose the system is asymptotically stable, i.e., $\Lambda(\mathbf{A})$ is contained in the left half-plane. Then, we have $\mathbf{x}(\cdot)$ and $\mathbf{y}(\cdot)$ are also bounded and integrable. By taking the Fourier transform of the LTI system, we have
\begin{equation}\label{eq.fourierLTI}
    \begin{aligned}
        si\hat{\mathbf{x}}(s) &= \mathbf{A} \hat{\mathbf{x}}(s) + \mathbf{B} \hat{\mathbf{u}}(s), \\
        \hat{\mathbf{y}}(s) &= \mathbf{C} \hat{\mathbf{x}}(s) + \mathbf{D} \hat{\mathbf{u}}(s).
    \end{aligned}
\end{equation}
Rearranging the first equation gives us
\[
    \hat{\mathbf{x}}(s) = (si\mathbf{I} - \mathbf{A})^{-1} \mathbf{B}\hat{\mathbf{u}}(s), \qquad s \in \R.
\]
Plugging it into the second equation of~\cref{eq.fourierLTI}, we can derive the transfer function on the imaginary axis:
\[
    \hat{\mathbf{y}}(s) = \underbrace{\left[\mathbf{C} (si\mathbf{I} - \mathbf{A})^{-1} \mathbf{B} + \mathbf{D}\right]}_{G(si)} \hat{\mathbf{u}}(s), \qquad s \in \R.
\]
Let $\mathbf{V} \in \C^{n \times n}$ be an invertible matrix. Consider the system $\tilde \Sigma = (\mathbf{V}^{-1}\mathbf{A}\mathbf{V}, \mathbf{V}^{-1}\mathbf{B}, \mathbf{C}\mathbf{V}, \mathbf{D})$:
\begin{equation}\label{eq.LTIconj}
    \begin{aligned}
        \mathbf{x}'(t) &= \mathbf{V}^{-1} \mathbf{A} \mathbf{V} \mathbf{x}(t) + \mathbf{V}^{-1} \mathbf{B} \mathbf{u}(t), \\
        \mathbf{y}(t) &= \mathbf{C} \mathbf{V}\mathbf{x}(t) + \mathbf{D} \mathbf{u}(t).
    \end{aligned}
\end{equation}
By multiplying the first equation by $\mathbf{V}$, we have
\[
	\mathbf{V}\mathbf{x}(t) = \mathbf{A} \mathbf{V}\mathbf{x}(t) + \mathbf{B}\mathbf{u}(t),
\]
and defining the new state variable $\boldsymbol{\xi}(t) = \mathbf{V}\mathbf{x}(t)$, we can write $\tilde{\Sigma}$ into
\begin{equation*}
    \begin{aligned}
        \boldsymbol{\xi}(t) &= \mathbf{A} \boldsymbol{\xi}(t) + \mathbf{B}\mathbf{u}(t), \\
        \mathbf{y}(t) &= \mathbf{C} \boldsymbol{\xi}(t) + \mathbf{D} \mathbf{u}(t).
    \end{aligned}
\end{equation*}
Hence~\cref{eq.LTIappend} and~(\ref{eq.LTIconj}) are equivalent with their states connected via $\mathbf{V}$. We also can verify this by computing the transfer function of $\tilde{\Sigma}$:
\[
	\tilde{G}(s) := \mathbf{C}\mathbf{V} (s\mathbf{I} - \mathbf{V}^{-1}\mathbf{A} \mathbf{V})^{-1} \mathbf{V}^{-1} \mathbf{B} + \mathbf{D} = \mathbf{C} (s\mathbf{I} - \mathbf{A})^{-1} \mathbf{B} + \mathbf{D} = G(s).
\]
The LTI system $\Sigma$ is continuous-time. In order to apply it to sequential input, we need to discretize the system. Given a step size $\Delta t$, there are two common ways of discretizing the system:
\begin{align*}
	\text{Bilinear}&: \overline{\mathbf{A}} = \left(\mathbf{I} -\! \frac{\Delta t}{2} \mathbf{A}\right)^{-1}\! \left(\mathbf{I} +\! \frac{\Delta t}{2} \mathbf{A}\right), \quad \overline{\mathbf{B}} = \Delta t \left(\mathbf{I} -\! \frac{\Delta t}{2} \mathbf{A}\right)^{-1}\! \mathbf{B}, \quad (\overline{\mathbf{C}}, \overline{\mathbf{D}}) = (\mathbf{C}, \mathbf{D}), \\
	\text{ZOH}&: \overline{\mathbf{A}} = \text{exp}(\Delta t \mathbf{A}), \quad \overline{\mathbf{B}} = \mathbf{A}^{-1} (\text{exp}(\Delta t \mathbf{A}) - \mathbf{I}) \mathbf{B}, \quad (\overline{\mathbf{C}}, \overline{\mathbf{D}}) = (\mathbf{C}, \mathbf{D}).
\end{align*}
Then, the discrete system
\begin{equation}\label{eq.discreteLTI}
\begin{aligned}
	\mathbf{x}_{t} &= \overline{\mathbf{A}} \mathbf{x}_{t-1} + \overline{\mathbf{B}} \mathbf{u}_{t-1}, \\
	\mathbf{y}_t &= \overline{\mathbf{C}} \mathbf{x}_{t} + \overline{\mathbf{D}} \mathbf{u}_t
\end{aligned}
\end{equation}
takes the discrete sequential input $(\mathbf{u}_{0}, \mathbf{u}_1, \ldots)$. The discrete system~\cref{eq.discreteLTI} mimics the continuous system~\cref{eq.LTIappend} by sampling the continuous input signal $\mathbf{u}(\cdot)$ at time intervals $\Delta t$: $(\mathbf{u}_0, \mathbf{u}_1, \ldots) = (\mathbf{u}(0\Delta t), \mathbf{u}(1 \Delta t), \ldots)$. The SSMs store the continuous LTI systems. When evaluating on a discrete input, they discretize the continuous systems using a trainable step size $\Delta t$ and either the Bilinear or the ZOH descritization.

\section{Proof of~\Cref{lem.difftransfer}}\label{sec:proofdifftransfer}

In this section, we prove~\Cref{lem.difftransfer} on the difference between the transfer functions $G_{\text{DPLR}}$ and $G_{\text{Diag}}$. The starting point is to use the Woodbury matrix identity to separate out the rank-$1$ part in the resolvent that appears in $G_{\text{DPLR}}$. In~\cref{sec:theorydiag}, we let $\mathbf{C} = \mathbf{e}_{\ell}^\top \mathbf{V}_H$ for some fixed $\ell$. Since we will reserve the letter $\ell$ as an index in the proof, in the appendices, we change the notation and assume $\mathbf{C} = \mathbf{e}_{p}^\top \mathbf{V}_H$. While this introduces a notation collision with the length of the output vector $\mathbf{y}$, it does not cause any confusion in the proofs.

\begin{proof}[Proof of~\Cref{lem.difftransfer}]
For notational cleanliness, in this proof, we define $\mathbf{A} = \mathbf{A}_H$, $\mathbf{A}^\perp = \mathbf{A}_H^\perp$, and $\mathbf{B} = \mathbf{B}_H$. To begin with, we expand $(s\mathbf{I} - \mathbf{A}_H^\perp)^{-1} \mathbf{B}_H$ using the Woodbury matrix identity~\cite{woodbury1950inverting}:
\begin{align*}
    &(s\mathbf{I} - \mathbf{A}_H^\perp)^{-1} \mathbf{B}_H = (s\mathbf I - \mathbf A - \mathbf B \mathbf B^\top)^{-1} \mathbf B \\
    &\qquad= \big[(s\mathbf I - \mathbf A)^{-1} + (s \mathbf I - \mathbf A)^{-1} \mathbf B (1- \mathbf B^\top(s \mathbf I - \mathbf A)^{-1} \mathbf B)^{-1} \mathbf B^\top(s \mathbf I - \mathbf A)^{-1}\big] \mathbf B \\
    &\qquad= (s\mathbf I - \mathbf A)^{-1} \mathbf B + \frac{\mathbf B^\top(s \mathbf I - \mathbf A)^{-1} \mathbf B}{1-\mathbf B^\top(s \mathbf I - \mathbf A)^{-1}\mathbf B}(s \mathbf I - \mathbf A)^{-1} \mathbf B \\
    &\qquad= (s\mathbf I - \mathbf A)^{-1}\mathbf B + \left(1 + \frac{2\mathbf B^\top(s\mathbf I - \mathbf A)^{-1}\mathbf B-1}{1-\mathbf B^\top(s \mathbf I - \mathbf A)^{-1}\mathbf B}\right) (s\mathbf I - \mathbf A)^{-1}\mathbf B \\
    &\qquad= 2 (s \mathbf I - \mathbf A)^{-1} \mathbf B + \frac{2 \mathbf B^\top(s \mathbf I - \mathbf A)^{-1} \mathbf B-1}{1-\mathbf B^\top(s\mathbf I - \mathbf A)^{-1}\mathbf B} (s \mathbf I - \mathbf A)^{-1} \mathbf B.
\end{align*}
Hence, when $\mathbf{C}_{\text{DPLR}} = \mathbf{C}_{\text{Diag}} = \mathbf{I}$, the difference between $G_{\text{DPLR}}$ and $G_{\text{Diag}}$ can be written as
\begin{equation}\label{eq.error}
   \frac{1}{2} (s\mathbf{I} - \mathbf{A}_H^\perp)^{-1} \mathbf{B}_H - (s\mathbf{I} - \mathbf{A}_H)^{-1} \mathbf{B}_H = \frac{2\mathbf B^\top(s\mathbf I - \mathbf A)^{-1}\mathbf B-1}{2-2\mathbf B^\top(s\mathbf I - \mathbf A)^{-1}\mathbf B} (s\mathbf I - \mathbf A)^{-1}\mathbf B.
\end{equation}
Our next step is to study $\mathbf B^\top(s\mathbf I - \mathbf A)^{-1}\mathbf B$ that appears in~\cref{eq.error}. To wit, we use Hua's identity~\cite{cohn2003further} to obtain
\begin{align*}
    \mathbf B^\top(s\mathbf I - \mathbf A)^{-1}\mathbf B &= \mathbf B^\top \left(-\mathbf A^{-1} + \left(\mathbf A - \frac{1}{s}\mathbf A^2\right)^{-1}\right) \mathbf B \\
    &= \mathbf B^\top \left(-\mathbf A^{-1} + s \left(\mathbf A(s\mathbf I - \mathbf A)\right)^{-1}\right) \mathbf B.
\end{align*}
It is easy to see that $\mathbf B^\top \mathbf A^{-1} \mathbf B = -1/2$. Hence, we have
\[
    \mathbf B^\top(s\mathbf I - \mathbf A)^{-1}\mathbf B = \frac{1}{2} + s\mathbf B^\top (s\mathbf I-\mathbf A)^{-1} \mathbf A^{-1} \mathbf B.
\]
Note that when $s = 0$, the second term in the expression above vanishes, and therefore we already have that $(\mathbf A^\perp)^{-1} \mathbf B/2 = \mathbf A^{-1} \mathbf B$. To deal with the general case when $s$ is a purely imaginary number, we first note that $\mathbf A^{-1} \mathbf B = -\mathbf e_1/\sqrt{2}$ because $\mathbf{B}$ is $-1/\sqrt{2}$ times the first column of $\mathbf{A}$. Hence, $s\mathbf B^\top (s\mathbf I- \mathbf A)^{-1} \mathbf A^{-1} \mathbf B$ is equal to $s$ times the first coordinate of $\mathbf B^\top (s\mathbf I- \mathbf A)^{-1}$, which we now compute using Cramer's rule. The first coordinate of $\mathbf B^\top (s \mathbf I - \mathbf A)^{-1}$ can be written as
\begin{align*}
    \mathbf B^\top (s\mathbf I - \mathbf A)^{-1} \mathbf e_1 &= \overline{\mathbf e_1^\top (s\mathbf I - \mathbf A)^{-*} \mathbf B} = \overline{\mathbf e_1^\top (-s\mathbf I - \mathbf A^*)^{-1} \mathbf B},
\end{align*}
where $\overline{s} = -s$ since $s$ is purely imaginary. By Cramer's rule, we have that
\begin{equation*}
    \mathbf e_1^\top (-s\mathbf I - \mathbf A^*)^{-1} \mathbf B = \frac{\text{det}\begin{bmatrix}
        1 & \sqrt{3} & \sqrt{5} & \cdots & \sqrt{2n-1} \\
        \sqrt{3} & 2-s & \sqrt{15} & \cdots & \sqrt{3(2n-1)} \\
        \sqrt{5} & 0 & 3-s & \cdots & \sqrt{5(2n-1)} \\
        \vdots & \vdots & \vdots & \ddots & \vdots \\
        \sqrt{2n-1} & 0 & 0 & \cdots & n-s
    \end{bmatrix}}{\sqrt{2}
    \text{det}\begin{bmatrix}
        1-s & \sqrt{3} & \sqrt{5} & \cdots & \sqrt{2n-1} \\
        0 & 2-s & \sqrt{15} & \cdots & \sqrt{3(2n-1)} \\
        0 & 0 & 3-s & \cdots & \sqrt{5(2n-1)} \\
        \vdots & \vdots & \vdots & \ddots & \vdots \\
        0 & 0 & 0 & \cdots & n-s
    \end{bmatrix}
    }.
\end{equation*}
Obviously, the denominator is $\sqrt{2}\prod_{j=1}^n (j-s)$. We compute the numerator by solving a recurrence. We use $D_n$ to denote this determinant. Hence, we have $D_1 = 1$ and $D_2 = -1-s$. To compute $D_n$, we expand the last row and obtain
\begin{align*}
    D_n \!=\! (-1)^{n+1} \sqrt{2n\!-\!1}\, \text{det}\!\!\begin{bmatrix}
        \sqrt{3} \!\!&\!\! \sqrt{5} \!\!&\!\! \cdots \!\!&\!\! \sqrt{2n\!-\!3} \!\!&\!\! \sqrt{2n\!-\!1} \\
        2\!-\!s \!\!&\!\! \sqrt{15} \!\!&\!\! \cdots \!\!&\!\! \sqrt{3(2n\!-\!3)} \!\!&\!\! \sqrt{3(2n\!-\!1)} \\
        0 \!\!&\!\! 3\!-\!s \!\!&\!\! \cdots \!\!&\!\! \sqrt{5(2n\!-\!3)} \!\!&\!\! \sqrt{5(2n\!-\!1)} \\
        \vdots \!\!&\!\! \vdots \!\!&\!\! \ddots \!\!&\!\! \vdots \!\!&\!\! \vdots \\
        0 \!\!&\!\! 0 \!\!&\!\! \cdots \!\!&\!\! n\!-\!1\!-\!s \!\!&\!\! \sqrt{(2n\!-\!3)(2n\!-\!1)}
    \end{bmatrix} \!\!+\! (n\!-\!s)D_{n-1}.
\end{align*}
To compute the determinant of this submatrix, we have
\begin{align*}
    &\text{det}\!\begin{bmatrix}
        \sqrt{3} & \sqrt{5} & \cdots & \sqrt{2n-3} & \sqrt{2n-1} \\
        2-s & \sqrt{15} & \cdots & \sqrt{3(2n-3)} & \sqrt{3(2n-1)} \\
        0 & 3-s & \cdots & \sqrt{5(2n-3)} & \sqrt{5(2n-1)} \\
        \vdots & \vdots & \ddots & \vdots & \vdots \\
        0 & 0 & \cdots & n-1-s & \sqrt{(2n-3)(2n-1)}
    \end{bmatrix} \\
    &\qquad = (-1)^{n-2} \text{det}\!\begin{bmatrix}
        \sqrt{2n\!-\!1} & \sqrt{3} & \sqrt{5} & \cdots & \sqrt{2n\!-\!3} \\
        \sqrt{3(2n\!-\!1)} & 2\!-\!s & \sqrt{15} & \cdots & \sqrt{3(2n\!-\!3)} \\
        \sqrt{5(2n\!-\!1)} & 0 & 3\!-\!s & \cdots & \sqrt{5(2n\!-\!3)} \\
        \vdots & \vdots & \vdots & \ddots & \vdots \\
        \sqrt{(2n\!-\!3)(2n\!-\!1)} & 0 & 0 & \cdots & n\!-\!1\!-\!s
    \end{bmatrix} \\
    &\qquad= (-1)^{n-1} \sqrt{2n\!-\!1} D_{n-1}.
\end{align*}
Hence, combining the two equations above, we obtain the following recurrence:
\begin{equation*}
    D_n = -(2n-1)D_{n-1} + (n-s)D_{n-1} = (-n+1-s)D_{n-1}.
\end{equation*}
It is then easy to show that
\[
    D_n = (-1-s)(-2-s) \cdots (-(n-1)-s) = (-1)^{n-1} \prod_{j=1}^{n-1} (j+s).
\]
Putting everything together, we have
\begin{equation}\label{eq.errorsub1}
    \mathbf B^\top (s\mathbf I-\mathbf A)^{-1}\mathbf B = \frac{1}{2} - s\,\text{conj} \left(\frac{(-1)^{n-1} \prod_{j=1}^{n-1} (j+s)}{2\prod_{j=1}^{n} (j-s)}\right) = \frac{1}{2} - s \frac{(-1)^{n-1} \prod_{j=1}^{n-1} (j-s)}{2\prod_{j=1}^{n} (j+s)}.
\end{equation}
Now, it remains to study the term $(s\mathbf I-\mathbf A)^{-1}\mathbf B$ in~\cref{eq.error}. Since it is a vector of length $n$, we study it component-wise, and the derivation is similar to the one above. To begin with, we fix a component $p$ that we wish to study. Then, by Cramer's rule, we have
\begin{align*}
    &\mathbf e_p^\top (s\mathbf I-\mathbf A)^{-1}\mathbf B \\
    &\qquad = \frac{\text{det}\!\!\begin{bmatrix}
        s+1 & 0 & 0 & \cdots & 1 & \cdots & 0 \\
        \sqrt{3} & s+2 & 0 & \cdots & \sqrt{3} & \cdots & 0 \\
        \sqrt{5} & \sqrt{15} & s+3 & \cdots & \sqrt{5} & \cdots & 0 \\
        \sqrt{7} & \sqrt{21} & \sqrt{35} & \cdots & \sqrt{7} & \cdots & 0 \\
        \vdots & \vdots & \vdots & \ddots & \vdots & \ddots & \vdots \\
        \sqrt{2n-1} & \sqrt{3(2n-1)} & \sqrt{5(2n-1)} & \cdots & \sqrt{2n-1} & \cdots & s+n  
    \end{bmatrix}}{\sqrt{2}\text{det}\!\!\begin{bmatrix}
        s+1 \!&\! 0 \!&\! 0 \!&\! \cdots \!&\! 0 \!&\! \cdots \!&\! 0 \\
        \sqrt{3} \!&\! s+2 \!&\! 0 \!&\! \cdots \!&\! 0 \!&\! \cdots \!&\! 0 \\
        \sqrt{5} \!&\! \sqrt{15} \!&\! s+3 \!&\! \cdots \!&\! 0 \!&\! \cdots \!&\! 0 \\
        \sqrt{7} \!&\! \sqrt{21} \!&\! \sqrt{35} \!&\! \cdots \!&\! 0 \!&\! \cdots \!&\! 0 \\
        \vdots \!&\! \vdots \!&\! \vdots \!&\! \ddots \!&\! \vdots \!&\! \ddots \!&\! \vdots \\
        \sqrt{2n\!-\!1} \!&\! \sqrt{3(2n\!-\!1)} \!&\! \sqrt{5(2n\!-\!1)} \!&\! \cdots \!&\! \sqrt{(2p\!-\!1)(2n\!-\!1)} \!&\! \cdots \!&\! s\!+\!n  
    \end{bmatrix}}.
\end{align*}
Clearly, we have that the denominator is equal to $\sqrt{2}\prod_{j=1}^n(j+s)$. To compute the numerator, we first subtract the $p$th column from the first column. This shows that the numerator is equal to
\begin{align}
    s\,\text{det}\begin{bmatrix}
        s+2 & 0 & \cdots & \sqrt{3} & \cdots & 0 \\
        \sqrt{15} & s+3 & \cdots & \sqrt{5} & \cdots & 0 \\
        \sqrt{21} & \sqrt{35} & \cdots & \sqrt{7} & \cdots & 0 \\
        \vdots & \vdots & \ddots & \vdots & \ddots & \vdots \\
        \sqrt{3(2n-1)} & \sqrt{5(2n-1)} & \cdots & \sqrt{2n-1} & \cdots & s+n  
    \end{bmatrix}.
\end{align}
We can then subtract $\sqrt{3}$ times the $(p-1)$th column of the submatrix from the first column, showing that the numerator is equal to
\begin{align*}
    s(s-1)\,\text{det}\begin{bmatrix}
        s+3 & \cdots & \sqrt{5} & \cdots & 0 \\
        \sqrt{35} & \cdots & \sqrt{7} & \cdots & 0 \\
        \vdots & \ddots & \vdots & \ddots & \vdots \\
        \sqrt{5(2n-1)} & \cdots & \sqrt{2n-1} & \cdots & s+n  
    \end{bmatrix}.
\end{align*}
Continuing in this manner, we have that the numerator is equal to
\begin{align*}
    &s(s\!-\!1)\cdots(s\!-\!p\!+\!2) \, \text{det}\begin{bmatrix}
        \sqrt{2p\!-\!1} & 0 & 0 & \cdots & 0 \\
        \sqrt{2p\!+\!1} & s\!+\!p\!+\!1 & 0 & \cdots & 0 \\
        \sqrt{2p\!+\!3} & \sqrt{(2p\!+\!3)(2p\!+\!1)} & s\!+\!p\!+\!2 & \cdots & 0 \\
        \vdots & \vdots & \vdots & \ddots & \vdots \\
        \sqrt{2n\!-\!1} & \sqrt{(2n\!-\!1)(2p\!+\!1)} & \sqrt{(2n\!-\!1)(2p\!+\!2)} & \cdots & s\!+\!n
    \end{bmatrix} \\
    &\qquad= \sqrt{2p-1} s(s-1)\cdots(s-p+2) (s+p+1)(s+p+2) \cdots (s+n).
\end{align*}
Hence, we have
\begin{equation}\label{eq.resolvantsol}
    \mathbf e_p^\top (s\mathbf I-\mathbf A)^{-1}\mathbf B = \frac{\sqrt{2p-1} \prod_{j=0}^{p-2} (s-j)}{\sqrt{2}\prod_{j=1}^p (s+j)}.
\end{equation}
Note that the expression above does not depend on $n$. Combining~\cref{eq.error},~(\ref{eq.errorsub1}),~(\ref{eq.resolvantsol}), when $\mathbf{C}_{\text{DPLR}} = \mathbf{C}_{\text{Diag}} = \mathbf{e}_p^\top$, we have
\begin{equation}\label{eq.functionG}
    \begin{aligned}
        G_{\text{DPLR}}(s) - G_{\text{Diag}}(s) &= \mathbf e_p^\top \left[\frac{1}{2}(s\mathbf I - \mathbf A^\perp)^{-1} \mathbf B - (s\mathbf I - \mathbf A)^{-1} \mathbf B\right] \\
        &= \frac{2\left(\frac{1}{2} - s\frac{(-1)^{n-1} \prod_{j=1}^{n-1} (j-s)}{2\prod_{j=1}^n (j+s)}\right)-1}{2-2\left(\frac{1}{2} - s\frac{(-1)^{n-1} \prod_{j=1}^{n-1}(j-s)}{2\prod_{j=1}^n (j+s)}\right)} \frac{\sqrt{2p-1}\prod_{j=0}^{p-2} (s-j)}{\sqrt{2} \prod_{j=1}^p (s+j)} \\
        &= \frac{-s\frac{(-1)^{n-1} \prod_{j=1}^{n-1} (j-s)}{\prod_{j=1}^n (j+s)} \sqrt{2p-1} \frac{\prod_{j=0}^{p-2}(s-j)}{\prod_{j=1}^p (s+j)}}{\sqrt{2} \left(1 + s\frac{(-1)^{n-1} \prod_{j=1}^{n-1}(j-s)}{\prod_{j=1}^n (j+s)}\right)}.
    \end{aligned}
\end{equation}
This completes the proof of the lemma.
\end{proof}

\section{Proof of~\Cref{thm.noconverge}}\label{sec:proofnoconverge}

In this section, we prove~\Cref{thm.noconverge}. The idea is to locate the last spike in the figure of $G_{\text{Diag}}$ (see~\Cref{fig:transfer}) and control the height of its peak by lower-bounding the denominator of~\cref{eq.difftransfer}.

\begin{proof}[Proof of~\Cref{thm.noconverge}]
Fix an $n \geq p$. Define $s_n$ by
    \begin{equation}
        s_n = \max\left\{s \geq 0 \middle| A(s) := si\frac{(-1)^{n-1} \prod_{j=1}^{n-1}(j-si)}{\prod_{j=1}^n (j+si)} \text{ is real and } \leq 0\right\}.
    \end{equation}
    Note that this set is finite because $A(s) \rightarrow 1$ as $s \rightarrow \infty$; thus, its supremum is attained. Therefore, we have that
    \begin{equation}\label{eq.worstcasedenom}
        \begin{aligned}
            \abs{1 + s_ni\frac{(-1)^{n-1} \prod_{j=1}^{n-1}(j-s_ni)}{\prod_{j=1}^n (j+s_ni)}} &= 1 - \abs{s_ni\frac{(-1)^{n-1} \prod_{j=1}^{n-1}(j-s_ni)}{\prod_{j=1}^n (j+s_ni)}} = \frac{\abs{n+s_ni} - s_n}{\abs{n+s_ni}}.
        \end{aligned}
    \end{equation}
    In what follows, we show that $s_n = \Omega(n^2)$\footnote{We say $f(n) = \Omega(g(n))$ if there exists a constant $C > 0$ such that $f(n) \geq Cg(n)$ for all $n \in \N$.}
    Then, combined with~\Cref{lem.difftransfer}, we have that as $n \rightarrow \infty$,
    \begin{align*}
        \abs{G_{\text{DPLR}}(s_ni) \!-\! G_{\text{Diag}}(s_ni)} &= \frac{s_n^2\sqrt{2p-1}}{\sqrt{2}\abs{p\!-\!1\!+\!s_ni}\abs{p\!+\!s_ni}(\abs{n\!+\!s_ni}\!-\!s_n)} = \Theta(1) \frac{1}{\sqrt{n^2\!+\!s_n^2}-s_n} \\
        &= \Theta(1) \frac{\sqrt{n^2\!+\!s_n^2}+s_n}{n^2}.
    \end{align*}
    If we can show that $s_n = \Omega(n^2)$, then we have that $\abs{G_{\text{DPLR}}(s_ni) \!-\! G_{\text{Diag}}(s_ni)} = \Omega(1)$ and does not converge to zero. To this end, we first rewrite the expression into
    \begin{equation}\label{eq.factordenom}
        A(s_n) \!=\! s_ni\frac{(-1)^{n-1} \prod_{j=1}^{n-1}(j\!-\!s_ni)}{\prod_{j=1}^n (j\!+\!s_ni)} \!=\! \frac{s_ni}{n\!+\!s_ni} \prod_{j=1}^{n-1} \frac{s_ni\!-\!j}{s_ni\!+\!j} \!=\! \frac{s_ni}{n\!+\!s_ni} \text{exp}\!\left(\!\!-i2\sum_{j=1}^{n-1} \arctan\frac{j}{s_n}\!\!\right).
    \end{equation}
    Since $\arctan x = \Theta(x)$ as $x \rightarrow 0$, if we assume, for a contradiction, that $s_{n_k} = o({n_k}^2)$ for a subsequence $s_{n_k}$ of $s_n$, then we must have that
    \[
        \sum_{j=1}^{{n_k}-1} \arctan\frac{j}{s_{n_k}} - \sum_{j=1}^{{n_k}-1} \arctan\frac{j}{\max\{{n_k},2s_{n_k}\}} \rightarrow \infty \qquad \text{as} \qquad k \rightarrow \infty.
    \]
    We pick some index ${n_k} \geq p$ large enough such that $\sum_{j=1}^{{n_k}-1} \arctan(j/s_{n_k}) - \sum_{j=1}^{{n_k}-1} \arctan(j/\max\{{n_k},2s_{n_k}\}) \geq 2\pi$. Hence, as $s$ increases from $s_{n_k}$ to $\max\{{n_k},2s_{n_k}\}$, the angle of the unit imaginary number
    \[
        \text{exp}\left(-i2\sum_{j=1}^{{n_k}-1} \arctan\frac{j}{s}\right)
    \]
    changes by at least $4\pi$ whereas the angle of ${si}/{(n+si)}$ changes by at most $\pi/2$. Hence, the winding number of the curve
    \[
    \Gamma: s \mapsto \frac{si}{{n_k}\!+\!si} \text{exp}\!\left(\!\!-i2\sum_{j=1}^{{n_k}-1} \arctan\frac{j}{s}\!\!\right), \qquad s \in [s_{n_k}, \max\{{n_k},2s_{n_k}\}],
    \]
    is non-zero. That is, we must have an $s \in (s_{n_k}, \max\{{n_k},2s_{n_k}\})$ such that the angle of $A(s)$ is equal to $\pi$ modulo $2\pi$, but this is a contradiction because $s_{n_k} < s$. Hence, we have $s_n = \Omega(n^2)$.
\end{proof}

\section{Proof of~\Cref{thm.weakstarconverge}}\label{sec:proofconverge}

In this section, we prove~\Cref{thm.weakstarconverge}. 
Since the proof is very involved, we provide some intuition here. 
In~\Cref{fig:transfer}, we observe that for a sufficiently large $n$, as $|s|$ increases, the difference between the two transfer functions, $G_{\text{DPLR}} - G_{\text{Diag}}$, goes through three stages.
In the first stage (i.e., the pre-spike stage), the large spikes have yet developed. 
In this stage, as $n$ increases, $|G_{\text{DPLR}} - G_{\text{Diag}}|$ decreases uniformly. 
In the second stage (i.e., the spike stage), the spikes start to occur. 
This is the stage in which we do not get uniform convergence. 
However, by carefully controlling the locations and the total measure of the spikes, we can show that when the Fourier transform of a fixed input function with a sufficient decay is multiplied with $G_{\text{DPLR}} - G_{\text{Diag}}$, its integral on the second stage vanishes linearly as $n \rightarrow \infty$. 
Finally, after the last spike, we enter the third stage (i.e., the post-spike stage).
In this stage, $|G_{\text{DPLR}} - G_{\text{Diag}}|$ enjoys rapid decay. 
In what follows, we carefully analyze the three stages separately to prove~\Cref{thm.weakstarconverge}.

\begin{figure}
    \centering
    \includegraphics[width=0.7\textwidth]{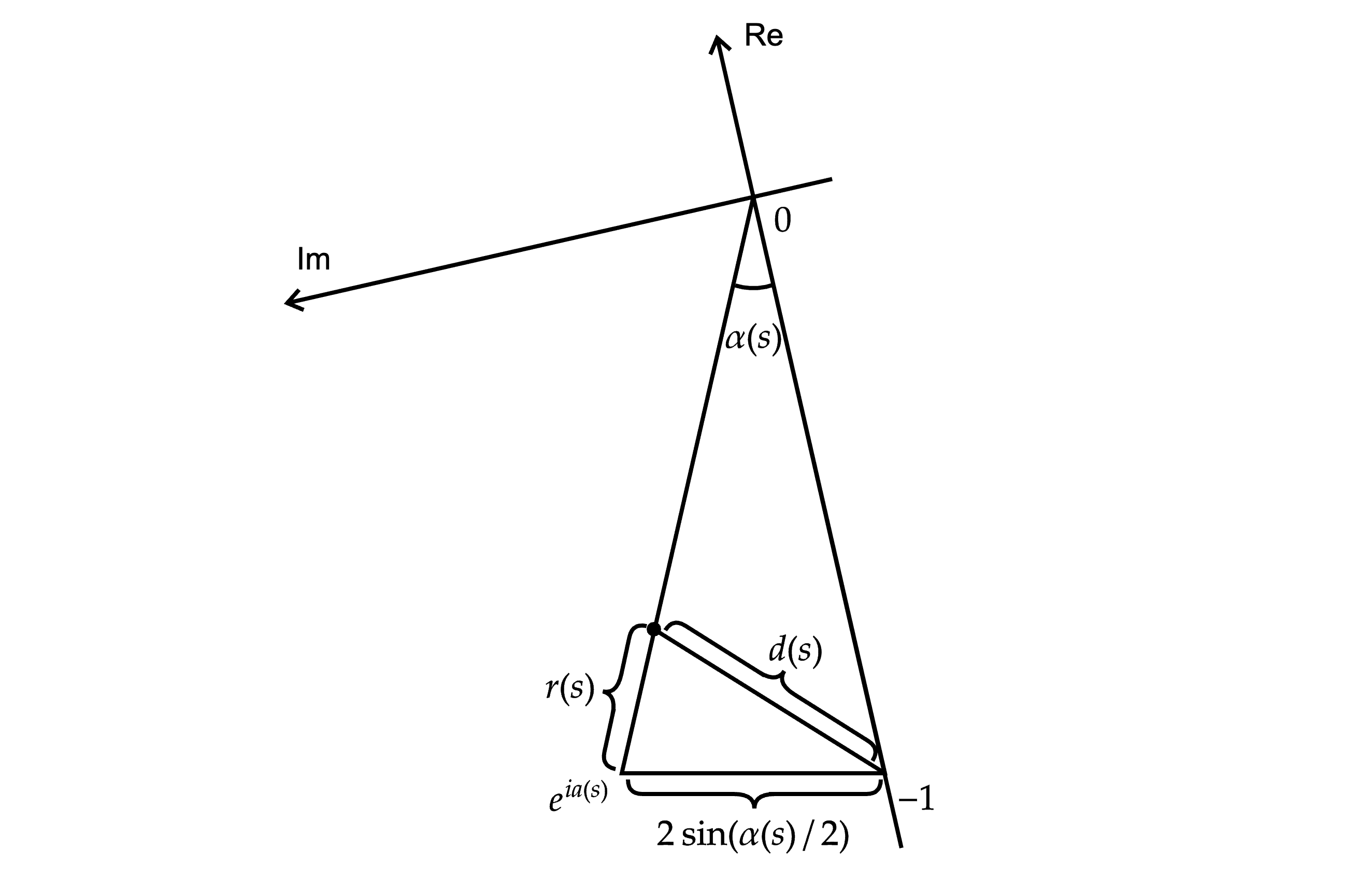}
    \caption{Illustration of the proof of~\Cref{thm.weakstarconverge}.}
    \label{fig:triangle}
\end{figure}

\begin{proof}[Proof of~\Cref{thm.weakstarconverge}]

Let $\mathbf{u}$ satisfy the assumptions in~\Cref{thm.weakstarconverge}. Without loss of generality, we assume $\hat{\mathbf{u}}(s)$ vanishes on $(-\infty,0]$ because the argument would be symmetric for a negative $s$. Let
\[
	H_n(s) = G_{\text{DPLR}}(s) - G_{\text{Diag}}(s) = \frac{-s\frac{(-1)^{n-1} \prod_{j=1}^{n-1} (j-s)}{\prod_{j=1}^n (j+s)} \sqrt{2p-1} \frac{\prod_{j=0}^{p-2}(s-j)}{\prod_{j=1}^p (s+j)}}{\sqrt{2} \left(1 + s\frac{(-1)^{n-1} \prod_{j=1}^{n-1}(j-s)}{\prod_{j=1}^n (j+s)}\right)}.
\]
We set $s^{(1)}_n = cn$, where $c$ is a universal constant determined later on, and 
\begin{equation}
        s^{(2)}_n = \max\left\{s \geq 0 \middle| A(s) := si\frac{(-1)^{n-1} \prod_{j=1}^{n-1}(j-si)}{\prod_{j=1}^n (j+si)} \text{ is purely imaginary and Im}(A(s)) \geq 0\right\}.
\end{equation}
By the same argument as in the proof of~\Cref{thm.noconverge}, we have $s^{(2)}_n = \mathcal{O}(n^2)$. To compute the integral of $\abs{H_{n}(si)\hat{\mathbf{u}}(s)}^2$ on $[0,\infty)$, we do so on each of the three stages, marked by $[0,s^{(1)}_n)$, $[s^{(1)}_n,s^{(2)}_n)$, and $[s^{(2)}_n,\infty)$, respectively. Since $2p-1$ is a constant appearing unanimously in $\abs{H_{n}(si)\hat{\mathbf{u}}(s)}^2$ for all $n$, we absorb it into the asymptotic notations in this proof. Unless otherwise stated, the constants in the asymptotic bounds in this proof are universal constants depending only on $p$; in particular, they do not depend on $n$ or $s$.

\noindent \textbf{Integrate on the pre-spike stage:} Since
\[
	\abs{s\frac{(-1)^{n-1} \prod_{j=1}^{n-1}(j-s)}{\prod_{j=1}^n (j+s)}} = \frac{\abs{s}}{\abs{n+s}}
\]
and $s = \mathcal{O}(n)$ whenever $0 \leq s \leq s_n^{(1)}$, the denominator of $H_{n}$ is lower-bounded by a constant independent of $n$. Hence, we have $\abs{H_{n}(si)} = \mathcal{O}(n^{-1})$ on $[0,s^{(1)}_n)$. 
Using H\"older's inequality, we~have
\begin{equation}\label{eq.controlflat}
    \int_0^{s^{(1)}_n} \abs{H_{n}(si) \hat{\mathbf{u}}(s)}^2 ds \leq \norm{H_{n}(si)}_{L^\infty([0,s^{(1)}_n))}^2 \norm{\hat{\mathbf{u}}(s)}_{L^2([0,s^{(1)}_n))}^2 = \mathcal{O}(n^{-2}).
\end{equation}

\noindent \textbf{Integrate on the post-spike stage:} For $s \geq s^{(2)}_n$, the denominator of $H_{n}$ is lower-bounded by a constant independent of $n$. Hence, we have $\abs{H_{n}(si)} = \mathcal{O}(s^{-1})$, where the constant does not depend on $n$. Hence, we have
\begin{equation}\label{eq.controldecay}
    \int_{s^{(2)}_n}^\infty \abs{H_{n}(si) \hat{\mathbf{u}}(s)}^2 ds = \int_{s^{(2)}_n}^\infty \mathcal{O}(s^{-2-2q}) ds = \mathcal{O}(n^{-2-4q})
\end{equation}
because $s_n^{(2)} = \mathcal{O}(n^2)$.

\noindent \textbf{Integrate on the spike stage:} To integrate $\abs{H_{n}(si) \hat{\mathbf{u}}(s)}^2$ on $[s_n^{(1)}, s_n^{(2)}]$, we first define the angle function by
\begin{align*}
    a(s) &:= \text{arg}\left(\frac{si}{n+si}\right) + 2\sum_{j=1}^{n-1} \arctan\left(\frac{j}{s}\right) = \arctan\left(\frac{n}{s}\right) + 2\sum_{j=1}^{n-1} \arctan\left(\frac{j}{s}\right) \\
     &\equiv \text{arg}\left(si\frac{(-1)^{n-1} \prod_{j=1}^{n-1}(j-si)}{\prod_{j=1}^n (j+si)}\right) \qquad (\text{mod } 2\pi).
\end{align*}
The importance of $a(s)$ is that when $a(s)$ is close to $(2k+1)\pi$ for some integer $k$, we get a spike in the figure of $\abs{H_{n}}$. We therefore partition the oscillation stage into two parts:
\[
    S_1 = \{s \in [s_n^{(1)},s_n^{(2)}) \mid \abs{a(s) - (2k+1)\pi} < \pi/4 \text{ for some } k \in \N\}, \qquad S_2 = [s_n^{(1)},s_n^{(2)}) \setminus S_1.
\]
The integral on $S_2$ is studied in the same way as the decay stage:
\begin{equation}\label{eq.oscillationS2}
    \int_{S_2} \abs{H_{n}(si) \hat{\mathbf{u}}(s)}^2 ds \leq \int_{\mathcal{O}(n)}^{\mathcal{O}(n^2)} \mathcal{O}(s^{-2-2q}) ds = \mathcal{O}(n^{-1-2q}).
\end{equation}
To study the spikes, we first need to derive a simplified expression of the denominator. Fix an $s \in S_1$. We let 
\[
\alpha(s) = \min_k \abs{a(s) - (2k+1)\pi}
\]
and
\[
    d(s) = \abs{1 + si\frac{(-1)^{n-1} \prod_{j=1}^{n-1}(j-si)}{\prod_{j=1}^n (j+si)}}.
\]
Since
\[
    r(s) := 1 - \abs{si\frac{(-1)^{n-1} \prod_{j=1}^{n-1}(j-si)}{\prod_{j=1}^n (j+si)}} = 1-\frac{s}{\sqrt{s^2+n^2}},
\]
by the cosine law, we have (see~\Cref{fig:triangle})
\begin{align*}
    &\cos\left(\frac{\pi}{2} - \frac{\alpha(s)}{2}\right) = \frac{-d^2 + r(s)^2 + 4\sin^2(\alpha(s)/2)}{4r(s)\sin(\alpha(s)/2)} \\
    &\qquad\Rightarrow d(s)^2 = r(s)^2 + 4\sin^2\left(\frac{\alpha(s)}{2}\right) - 4r(s)\sin^2\left(\frac{\alpha(s)}{2}\right) \geq r(s)^2 + \sin^2\left(\frac{\alpha(s)}{2}\right),
\end{align*}
where the last inequality follows from the fact that $r(s) < 1/2$ for a sufficiently large constant $c$ in the definition of $s_n^{(1)}$.\footnote{This is our only requirement of the universal constant $c$ appearing in the definition of $s_n^{(1)}$.} Therefore, we have
\begin{equation}\label{eq.Gintermediate}
    \abs{H_{n}(si) \hat{\mathbf{u}}(s)}^2 = \mathcal{O}(s^{-2-2q}) \frac{1}{d(s)^2} \leq \mathcal{O}(s^{-2-2q}) \frac{1}{r(s)^2 + \alpha(s)^2},
\end{equation}
where we used the fact that $x/\pi \leq \sin(x) \leq x$ for all $0 \leq x \leq \pi/2$. Clearly, we have
\begin{equation}\label{eq.controlrs}
    r(s)^2 = \left(\frac{\sqrt{s^2+n^2}-s}{\sqrt{s^2+n^2}}\right)^2 = \left(\frac{s^2 + n^2 - s^2}{\sqrt{s^2+n^2}(\sqrt{s^2+n^2} + s)}\right)^2 = \mathcal{O}\left(\frac{n^4}{s^4}\right)
\end{equation}
because $s = \Omega(n)$. To study $\alpha(s)$, we first need to compute $a(s)$. To this end, note that since we assume $s = \Omega(n)$, there exist two universal constants $C_1, C_2 > 0$, independent of $n$, such that
\[
    C_1 \frac{j}{s} \leq \arctan\left(\frac{j}{s}\right) \leq C_2 \frac{j}{s}, \qquad 1 \leq j \leq n-1.
\]
Hence, we have
\[
    a(s) = \Theta\left(\frac{n^2}{s}\right).
\]
By the intermediate value theorem and monotonicity of $a$, there are $k_n = \mathcal{O}(n)$ frequencies $s_1, \ldots, s_{k_n}$ between $s = s_n^{(1)}$ and $s = s_n^{(2)}$ such that $a(s_j) \equiv \pi \; (\text{mod } 2\pi)$ for all $1 \leq j \leq {k_n}$. Each $s_j$ is contained in a connected component $S_1^{(j)} = (\xi_j,\zeta_j)$ of $S_1$ and $S_1 = \bigcup_{j=1}^{k_n} S_1^{(j)}$. That is, we have
\[
    s_n^{(1)} < \xi_{k_n} < s_{k_n} < \zeta_{k_n} < \xi_{k_n-1} < s_{k_n-1} < \zeta_{k_n-1} < \cdots < \xi_1 < s_1 < \zeta_1 < s_n^{(2)}.
\]
Moreover, there are two universal constants $C_1, C_2 > 0$, independent of $n$ or $j$, such that
\[
    C_1 j^{-1} n^2 \leq s_j \leq C_2 j^{-1} n^2.
\]
Combined with~\cref{eq.controlrs}, we have
\[
    r(s)^2 = \mathcal{O}\left(\frac{j^4}{n^4}\right), \qquad s \in S_1^{(j)}, \quad 1 \leq j \leq k_n,
\]
where the constant is universal and does not depend on $n$ or $j$. To integrate $\abs{H_{n}(si)\hat{\mathbf{u}}(s)}^2$ on $S_1$, we integrate it on each of $(\xi_j,\zeta_j)$. To do so, we study the value of $\alpha(s)$ using the Mean Value Theorem. First, we note that for any given $s_j$, we have
\begin{align*}
    \frac{d}{ds} a(s_j) &= -\Theta(1) \sum_{k=1}^{n-1} \frac{1}{1+\frac{k^2}{s_j^2}} \frac{k}{s_j^2} = -\Theta(1) \frac{1}{s_j^2} \sum_{k=1}^{n-1} \frac{k}{\frac{s_j^2+k^2}{s_j^2}} \\
    &= -\Theta(1) \sum_{k=1}^{n-1} \frac{k}{s_j^2} = -\Theta(1)\frac{n^2}{s_j^2} = -\Theta(1) \frac{j^2}{n^2},
\end{align*}
where the constant in the $\Theta$-notation does not depend on $n$ or $j$. Hence, fixing a $1 \leq j \leq k_n$ and choosing $s \in (\xi_j,\zeta_j)$, by the Mean Value Theorem, we have
\[
    \alpha(s) = \abs{a(s) - a(s_j)} = \Theta(1) \frac{j^2}{n^2} \abs{s-s_j}.
\]
This shows $\zeta_j - s_j, s_j - \xi_j = \Theta(n^2/j^2)$. Hence, we have
\begin{equation}
    \begin{aligned}
        &\int_{\xi_j}^{\zeta_j} \abs{H_{n}(si)\hat{\mathbf{u}}(s)}^2 ds = \mathcal{O}((j^{-1}n^2)^{-2-2q}) \int_{\xi_j}^{\zeta_j} \frac{1}{r(s)^2 + \alpha(s)^2} ds \\
        &\qquad\leq \mathcal{O}(j^{2+2q}n^{-4-4q}) \left(\int_{s_j-1}^{s_j+1} \frac{1}{r(s)^2} ds + \int_{\xi_j}^{s_j-1} \frac{1}{\alpha(s)^2} ds + \int_{s_j+1}^{\zeta_j} \frac{1}{\alpha(s)^2} ds\right) \\
        &\qquad\leq \mathcal{O}(j^{2+2q}n^{-4-4q}) \left(\frac{n^4}{j^4} + \frac{n^4}{j^4} \int_1^{\Theta(n^2/j^2)} \delta^{-2} d\delta\right) = \mathcal{O}(j^{-2+2q} n^{-4q}).
    \end{aligned}
\end{equation}
Suppose $q > 1/2$ and let $q' = q-1/2$. Then, we have
\begin{equation}\label{eq.oscillationS1}
    \begin{aligned}
        \int_{S_1} \abs{H_{n}(si)\hat{\mathbf{u}}(s)}^2 ds &= \sum_{j=1}^{k_n} \int_{\xi_j}^{\zeta_j} \abs{H_{n}(si)\hat{\mathbf{u}}(s)}^2 ds = \mathcal{O}(1) \sum_{j=1}^{k_n} j^{-1+2q'} n^{-2-4q'} \\
        &\leq \mathcal{O}(n^{-2}) \sum_{j=1}^{k_n} j^{-1-2q'} = \mathcal{O}(n^{-2}),
    \end{aligned}
\end{equation}
where the constant in the last $\mathcal{O}$-notation only depends on $p$. Combining~\cref{eq.oscillationS2} and~(\ref{eq.oscillationS1}), we have that when $q > 1/2$, it holds that
\begin{equation}\label{eq.controloscillation}
    \int_{s_n^{(1)}}^{s_n^{(2)}} \abs{H_{n}(si)\hat{\mathbf{u}}(s)}^2 ds = \mathcal{O}(n^{-2}).
\end{equation}

\noindent \textbf{Put everything together:} Combining~\cref{eq.controlflat},~(\ref{eq.controldecay}), and~(\ref{eq.controloscillation}) and applying Parseval's identity, we obtain
\begin{align*}
&\norm{\mathbf{y}_{\text{DPLR}} \!-\! \mathbf{y}_{\text{Diag}}}_{L^2} = \norm{\hat{\mathbf{y}}_{\text{DPLR}} \!-\! \hat{\mathbf{y}}_{\text{Diag}}}_{L^2} = \norm{H_n(si)\hat{\mathbf{u}}(s)}_{L^2} \\
&\qquad= \sqrt{\int_0^{s_n^{(1)}} \abs{H_{n}(si)\hat{\mathbf{u}}(s)}^2 ds + \int_{s_n^{(1)}}^{s_n^{(2)}} \abs{H_{n}(si)\hat{\mathbf{u}}(s)}^2 ds + \int_{s_n^{(2)}}^\infty \abs{H_{n}(si)\hat{\mathbf{u}}(s)}^2 ds} = \mathcal{O}(n^{-1}).
\end{align*}
This completes the proof.
\end{proof}

\section{Numerical experiments on~\Cref{thm.weakstarconverge} and~\ref{thm.noconverge}}\label{sec:experimentconverge}

In this section, we present three numerical experiments that elaborate our theory in~\cref{sec:theorydiag}. The first experiment examines the behaviors of the DPLR system and the diagonal system given a single Fourier mode as an input. By doing so, we observe the ``numerical unstable modes" of the S4D model. This corroborates~\Cref{thm.noconverge}. Then, we compare the two systems using two different input functions: an exponentially decaying function and the unit impulse. We will show that the smoothness condition in~\Cref{thm.weakstarconverge} is necessary and the linear convergence rate is tight.

In each of these experiments, we simulate LTI systems on some continuous input signals. It is done as follows: given an input signal $u(t)$\footnote{Since $u$ will be scalar-valued in this section, we do not make it boldface.} and an LTI system $\Sigma$, we fix the step size to be $\Delta t = 10^{-3}$. For some final time step $N$, we discretize our input function to obtain a vector $\mathbf{u} = (u(0), u(\Delta t), \ldots, u(N\Delta t))$. We then discretize the LTI system bilinearly (see~\Cref{sec:morebackground}) and compute its output $\mathbf{y}$ on the input $\mathbf{u}$. We call this procedure ``simulate", i.e., 
\[
	\mathbf{y} = \texttt{simulate}(u,\Sigma,N).
\]
In this section, we let $\Sigma_{\text{DPLR},n}$ to be the DPLR system with state size $n$ of S4 and $\Sigma_{\text{Diag},n}$ to be the diagonal system with state size $n$ of S4D, where we always take $\mathbf{C} = \mathbf{e}_1$ and $\mathbf{D} = \boldsymbol{0}$.

\subsection{The diagonal system behaves differently for distinct Fourier modes}

\begin{figure}
    \centering
    \begin{minipage}{.32\textwidth}
    \begin{overpic}[width=0.95\textwidth]{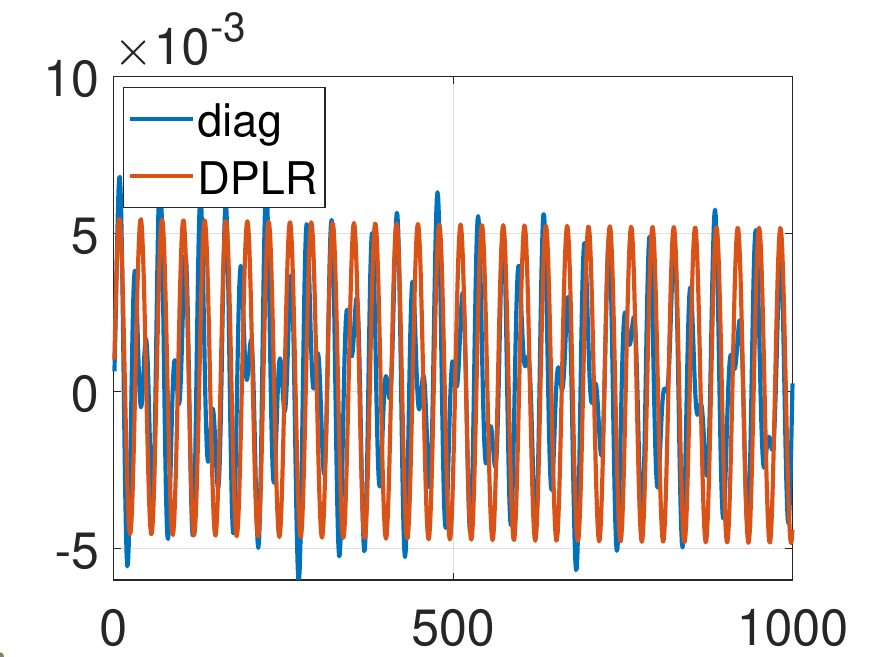}
        \put(40,-6) {time step}
        \put(-2,18) {\rotatebox{90}{output value}}
        \put(37,75) {$\bf{s = 200}$}
    \end{overpic}
    \end{minipage}
    \begin{minipage}{.32\textwidth}
    \begin{overpic}[width=.95\textwidth]{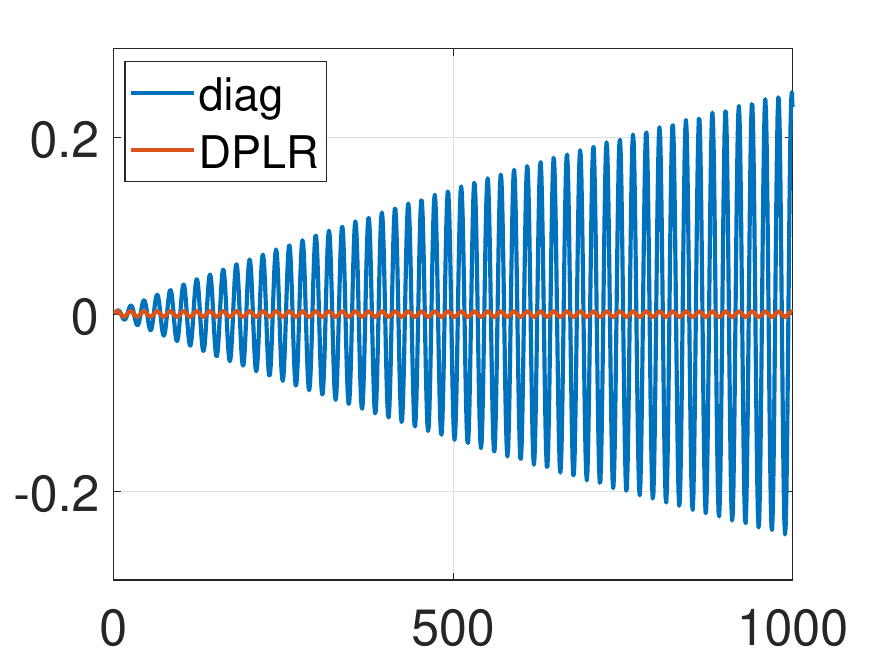}
    	\put(35,75) {$\bf{s = 322.5}$}
        \put(40,-6) {time step}
        \put(-5,18) {\rotatebox{90}{output value}}
    \end{overpic} 
    \end{minipage}
    \begin{minipage}{.32\textwidth}
    \begin{overpic}[width=.95\textwidth]{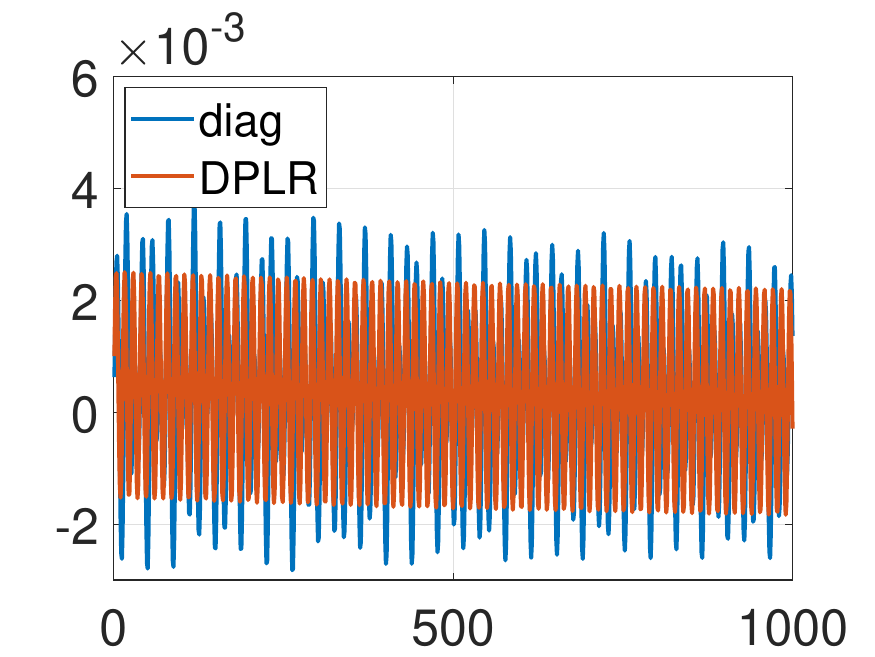}
    	\put(37,75) {$\bf{s = 500}$}
        \put(40,-6) {time step}
        \put(-2,18) {\rotatebox{90}{output value}}
    \end{overpic}
    \end{minipage}
    \vspace{0.3cm}
    \caption{Simulated outputs of the DPLR and diagonal systems with cosine wave inputs of different frequencies $s$. Note the scale of the $y$-axis when $s = 322.5$.}
    \label{fig:simulatecos}
\end{figure}
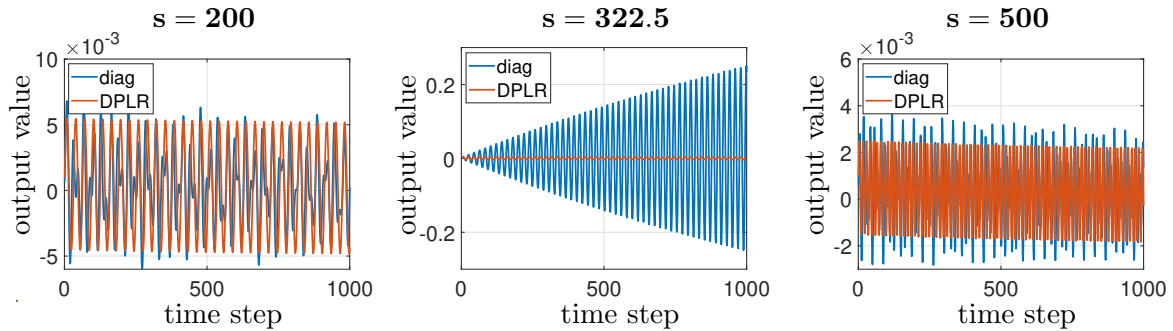

Our first experiment considers the outputs of $\Sigma_{\text{DPLR},n}$ and $\Sigma_{\text{Diag},n}$ when the input is a cosine wave
\[
	u_s(t) = \cos(st).
\]
This function has a dense Fourier mode at frequency $s$. We fix $n = 32$ and let $s$ change. In~\Cref{fig:simulatecos}, we plot $\texttt{simulate}(u_s,\Sigma_{\text{DPLR},32}, 10^3)$ and $\texttt{simulate}(u_s,\Sigma_{\text{Diag},32}, 10^3)$ with $s = 200, 322.5$, and $500$, respectively. We see that when $s = 200$ or $500$, the outputs of the two systems are close to each other - at least, they are on the same order of magnitude. However, when $s = 322.5$, the output of the diagonal system blows up. In fact, this value of $s$ is exactly where the spike in the plot of $\|G_{\text{Diag}}\|$ occurs when $n = 32$. Hence, we visualize the counter-example that shows the divergence in~\Cref{thm.noconverge}.

\subsection{The DPLR and diagonal systems converge on the exponentially decaying function}

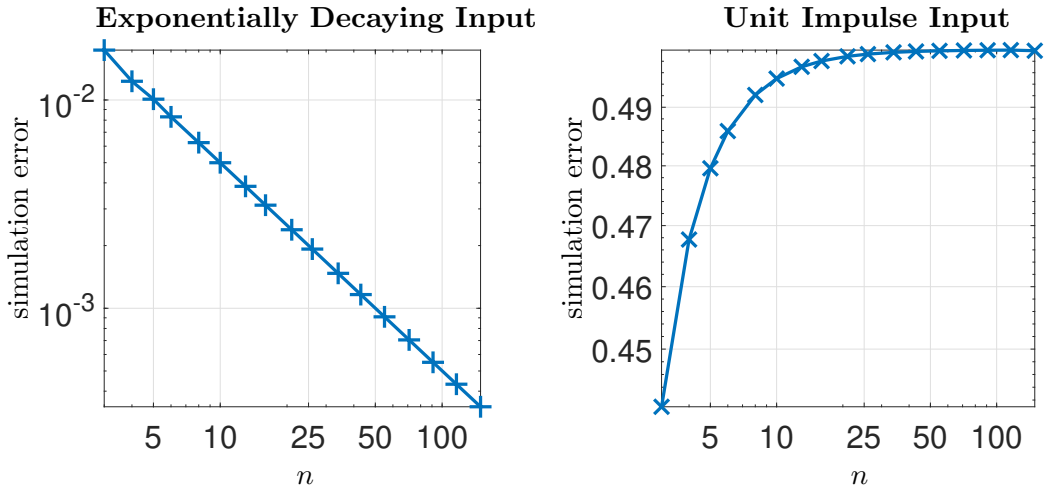
\begin{figure}[!htb]
	\centering
	\begin{minipage}{.45\textwidth}
	\begin{overpic}[width=.9\textwidth]{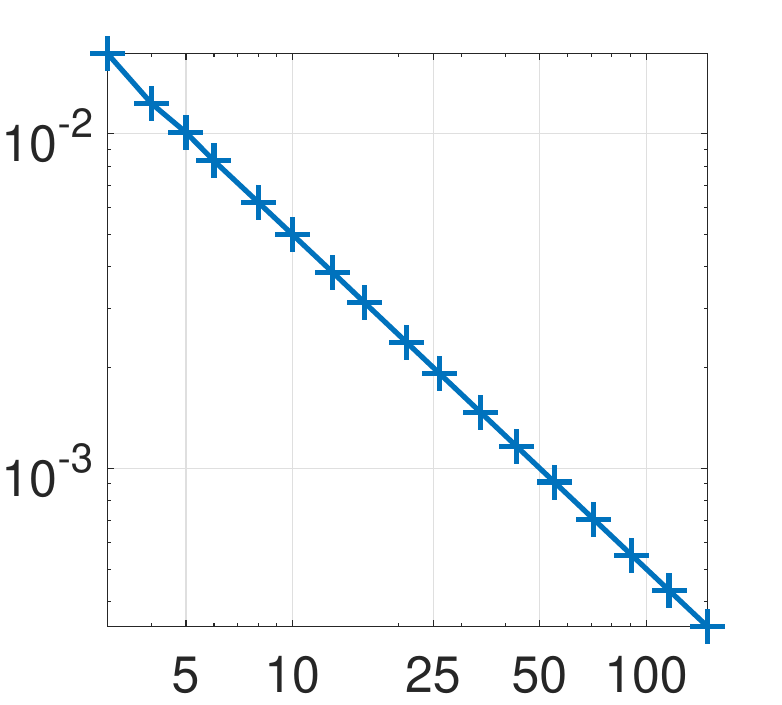}
	\put(12,88) {\textbf{Exponentially Decaying Input}}
        \put(53,-6) {$n$}
        \put(-6,28) {\rotatebox{90}{simulation error}}
        \end{overpic}
        \end{minipage}
        \begin{minipage}{.45\textwidth}
	\begin{overpic}[width=.9\textwidth]{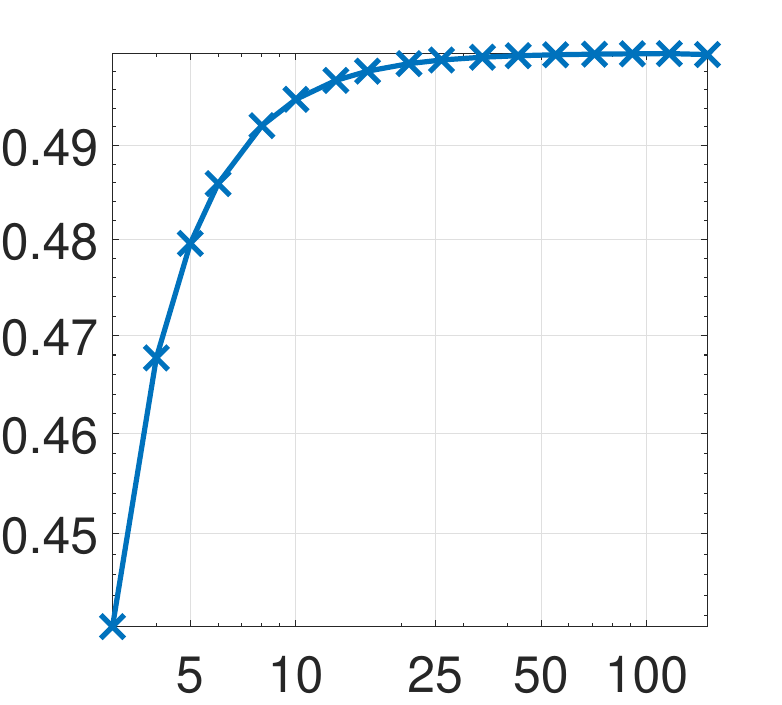}
	\put(27,88) {\textbf{Unit Impulse Input}}
        \put(53,-6) {$n$}
        \put(-6,28) {\rotatebox{90}{simulation error}}
        \end{overpic}
        \end{minipage}
        \vspace{0.3cm}
        \caption{The difference between the outputs $\|\texttt{simulate}(u,\Sigma_{\text{DPLR},n}, 10^4) - \texttt{simulate}(u,\Sigma_{\text{Diag},n}, 10^4)\|$ for difference values of $n$ when $u$ is the exponentially decaying function $u_e$ (left) and the unit impulse signal $\delta_0$ (right).}
        \label{fig:convergesingle}
\end{figure}

To test the function-wise convergence of the diagonal system to the DPLR system (see~\Cref{thm.weakstarconverge}), we consider the following exponentially decaying function:
\[
	u_e(t) = e^{-t} H(t),
\]
where $H = \mathbbm{1}_{[0,\infty)}$ is the Heaviside function. The Fourier transform of this function is
\[
	\hat{u}_e(s) = \frac{1}{1+is}.
\]
Hence, it is a function that satisfies the assumptions of~\Cref{thm.weakstarconverge}. In the left panel of \Cref{fig:convergesingle}, we show the difference between the two simulated outputs $\|\texttt{simulate}(u_e,\Sigma_{\text{DPLR},n}, 10^4) - \texttt{simulate}(u_e,\Sigma_{\text{Diag},n}, 10^4)\|$ as $n$ increases. We see that as $n$ increases, $\texttt{simulate}(u_e,\Sigma_{\text{Diag},n}$ converges to $\texttt{simulate}(u_e,\Sigma_{\text{DPLR},n}, 10^4)$. Moreover, the slope of the curve is roughly $-1$, indicating a linearly convergence rate as $n \rightarrow \infty$. This matches the theoretical statement in~\Cref{thm.weakstarconverge}.

\begin{figure}
    \centering
    \begin{minipage}{.32\textwidth}
    \begin{overpic}[width=0.95\textwidth]{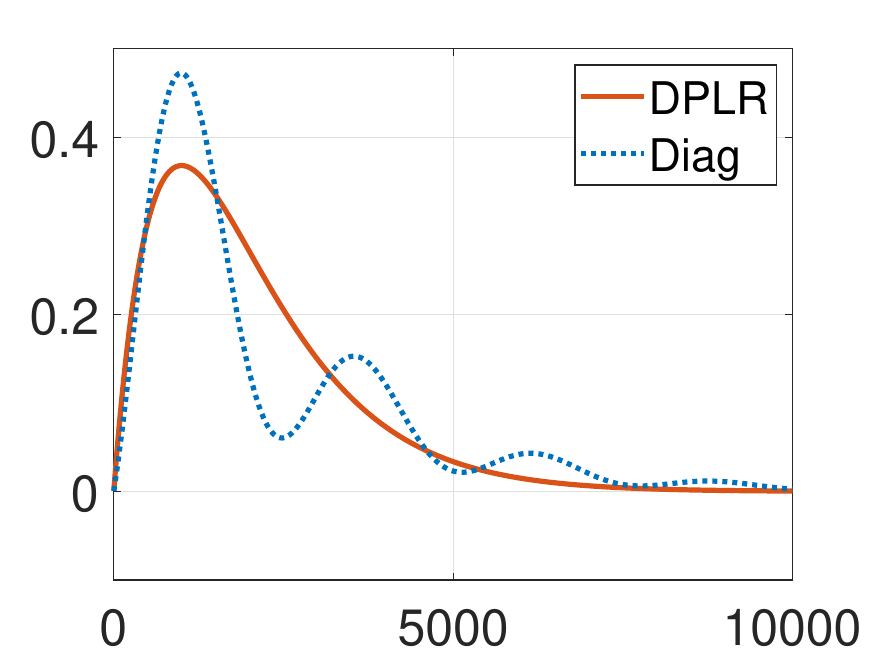}
        \put(40,-6) {time step}
        \put(-5,18) {\rotatebox{90}{output value}}
        \put(40.5,75) {$\bf{n = 3}$}
    \end{overpic}
    \end{minipage}
    \begin{minipage}{.32\textwidth}
    \begin{overpic}[width=.95\textwidth]{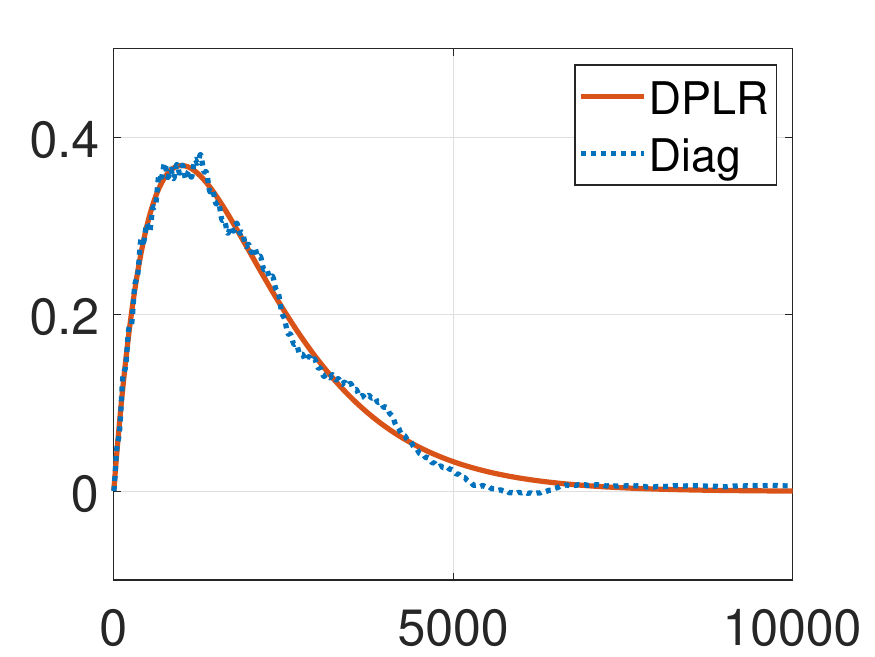}
    	\put(38.5,75) {$\bf{n = 15}$}
        \put(40,-6) {time step}
        \put(-5,18) {\rotatebox{90}{output value}}
    \end{overpic} 
    \end{minipage}
    \begin{minipage}{.32\textwidth}
    \begin{overpic}[width=.95\textwidth]{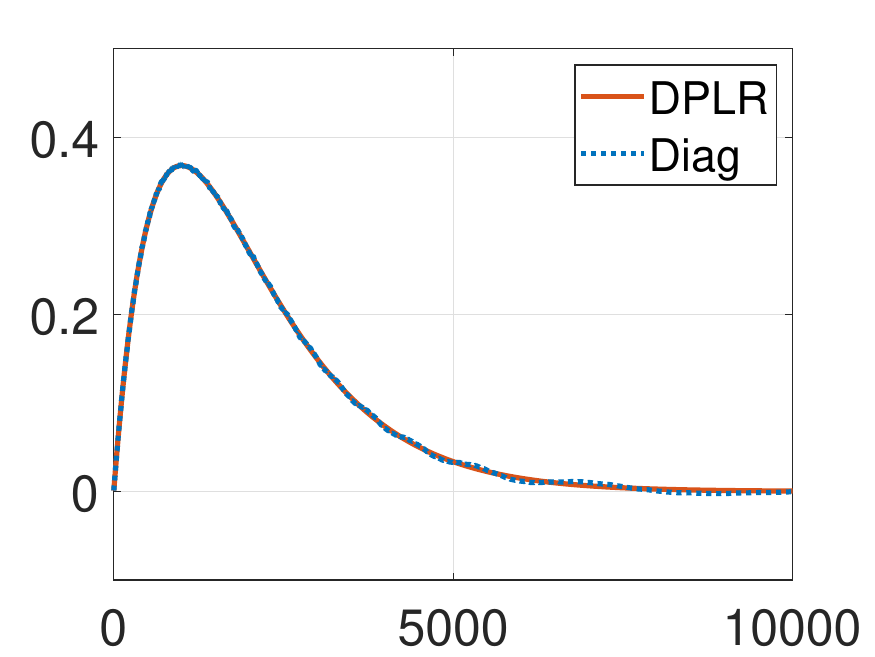}
    	\put(38.5,75) {$\bf{n = 75}$}
        \put(40,-6) {time step}
        \put(-5,18) {\rotatebox{90}{output value}}
    \end{overpic}
    \end{minipage}
    \vspace{0.3cm}
    \caption{Simulated outputs of the DPLR and diagonal systems with the exponentially decaying input function and varying state-space dimension $n$.}
    \label{fig:simulateexp}
\end{figure}
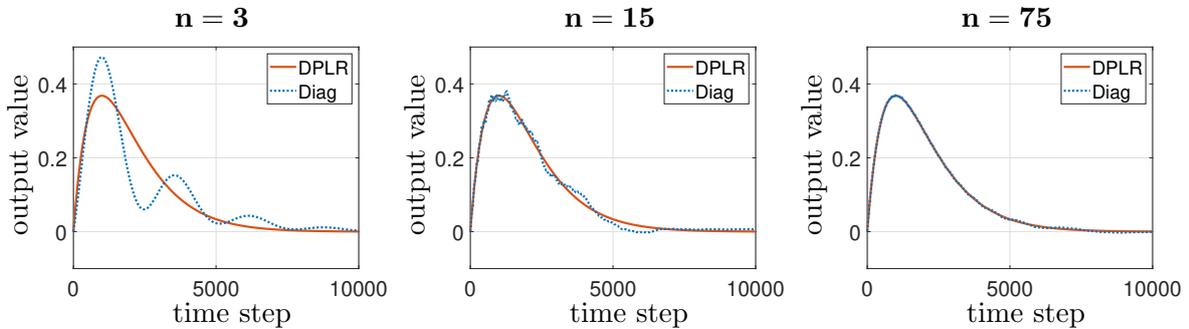

In~\Cref{fig:simulateexp}, we show the behaviors of the two simulated outputs as $n$ increases. For the exponentially decaying input function $u_e$, the outputs demonstrate a clear pattern of convergence as $n \rightarrow \infty$.

\subsection{The DPLR and diagonal systems diverge on the unit impulse}

\begin{figure}
    \centering
    \begin{minipage}{.32\textwidth}
    \begin{overpic}[width=0.95\textwidth]{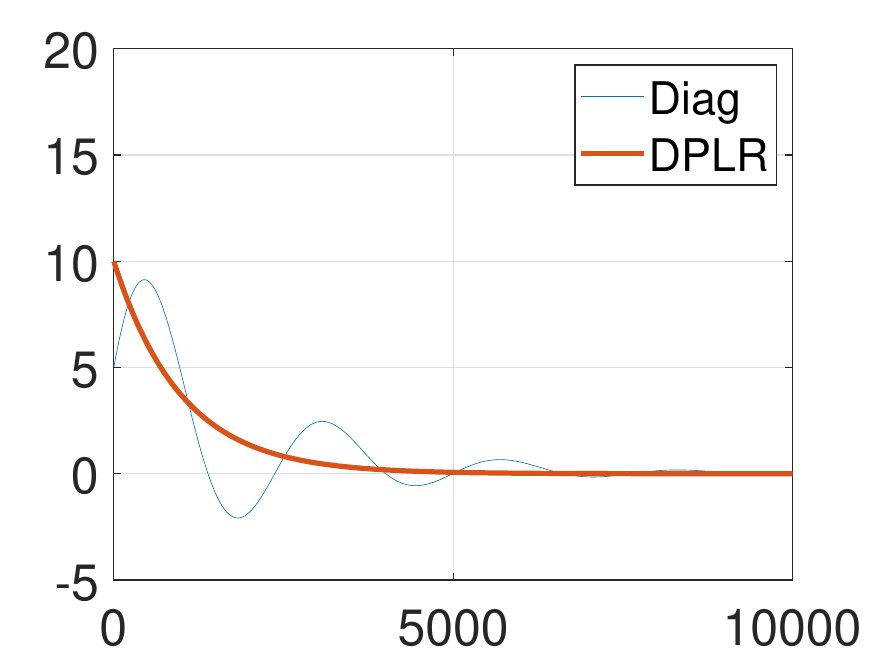}
        \put(40,-6) {time step}
        \put(-5,18) {\rotatebox{90}{output value}}
        \put(40.5,75) {$\bf{n = 3}$}
    \end{overpic}
    \end{minipage}
    \begin{minipage}{.32\textwidth}
    \begin{overpic}[width=.95\textwidth]{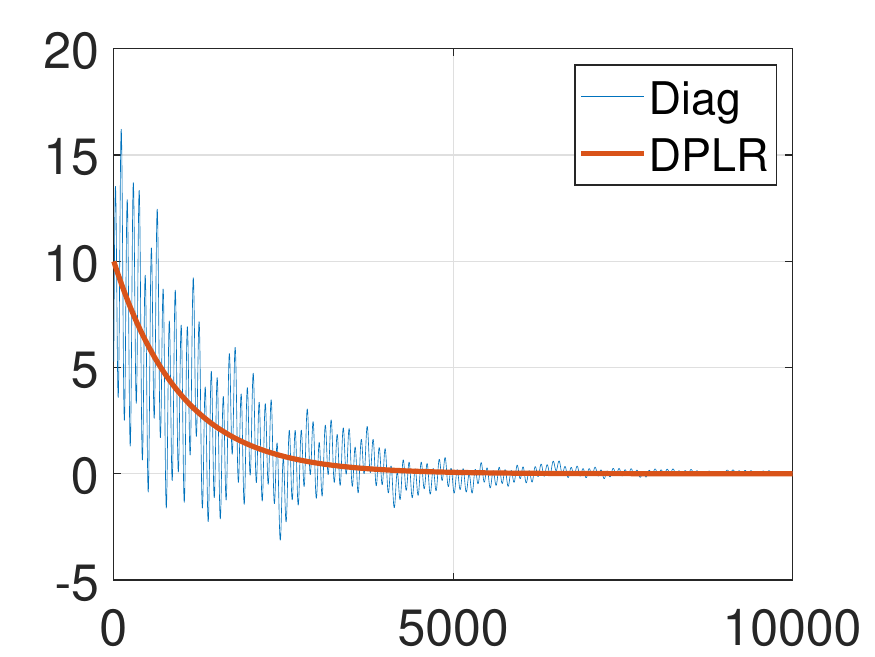}
    	\put(38.5,75) {$\bf{n = 15}$}
        \put(40,-6) {time step}
        \put(-5,18) {\rotatebox{90}{output value}}
    \end{overpic} 
    \end{minipage}
    \begin{minipage}{.32\textwidth}
    \begin{overpic}[width=.95\textwidth]{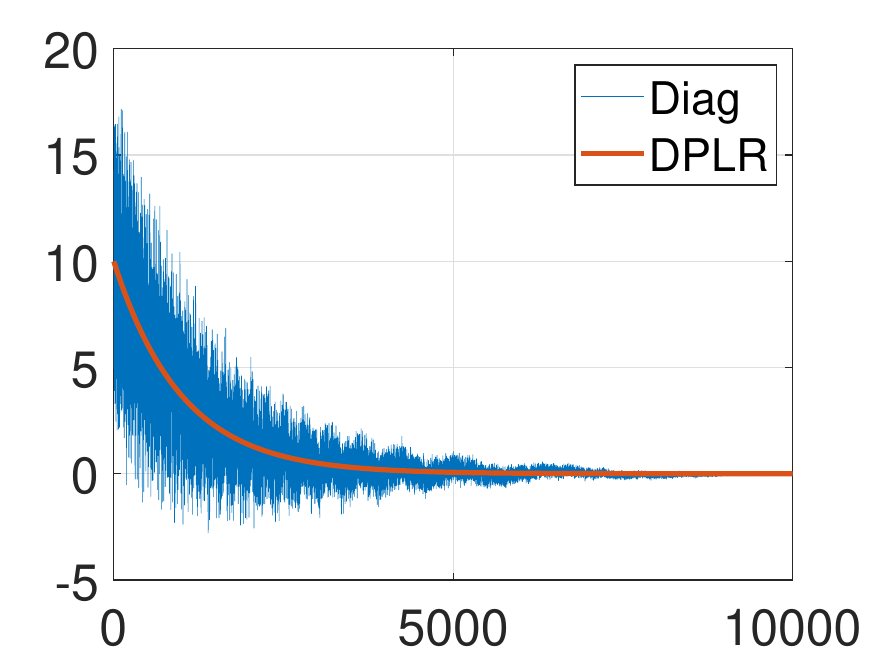}
    	\put(38.5,75) {$\bf{n = 75}$}
        \put(40,-6) {time step}
        \put(-5,18) {\rotatebox{90}{output value}}
    \end{overpic}
    \end{minipage}
    \vspace{0.3cm}
    \caption{Simulated outputs of the DPLR and diagonal systems with the unit impulse input and varying state-space dimension $n$.}
    \label{fig:simulatedelta}
\end{figure}
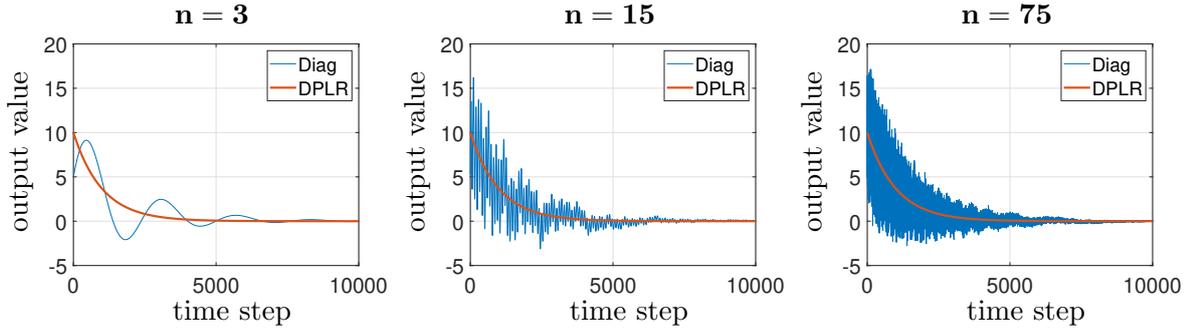

Our experiment with the exponentially decaying input shows that the DPLR and diagonal systems converge on a sufficiently smooth input function. One may wonder, however, if the smoothness condition is necessary. To show that a mild one is indeed required, we consider the Dirac delta function $\delta_0$. It is well-known that the Fourier transform of it is constantly one:
\[
	\hat{\delta}_0(s) = 1, \qquad s \in \R.
\]
In that sense, $\delta_0$ is highly non-smooth as its Fourier transform does not decay at all. Since $\delta_0$ is a distribution rather than a classical function, we cannot sample it directly. However, we can mimic it by setting the discrete input to be the unit impulse $(1,0,0,\ldots,0)$. In the right panel of \Cref{fig:convergesingle}, we see that $\|\texttt{simulate}(\delta_0,\Sigma_{\text{DPLR},n}, 10^4) - \texttt{simulate}(\delta_0,\Sigma_{\text{Diag},n}, 10^4)\|$ does not decay as $n$ increases. We can take a closer look in~\Cref{fig:simulatedelta}, where we plot the two output functions with different state-space dimensions $n$. In particular, we see that as $n$ increases, the output of the DPLR system remains the same, whereas the output of the diagonal system becomes more oscillatory. We do not have convergence. The oscillatory behavior can be explained by our observation in~\Cref{fig:transfer}: the larger the $n$, the later the spike emerges. This means that for a larger $n$, the outputs of two systems differ at a higher frequency (i.e., a more oscillatory mode).

\section{Quantitative interpolation and extrapolation errors in~\cref{sec:failuremodes}}\label{sec:quantitativeerr}

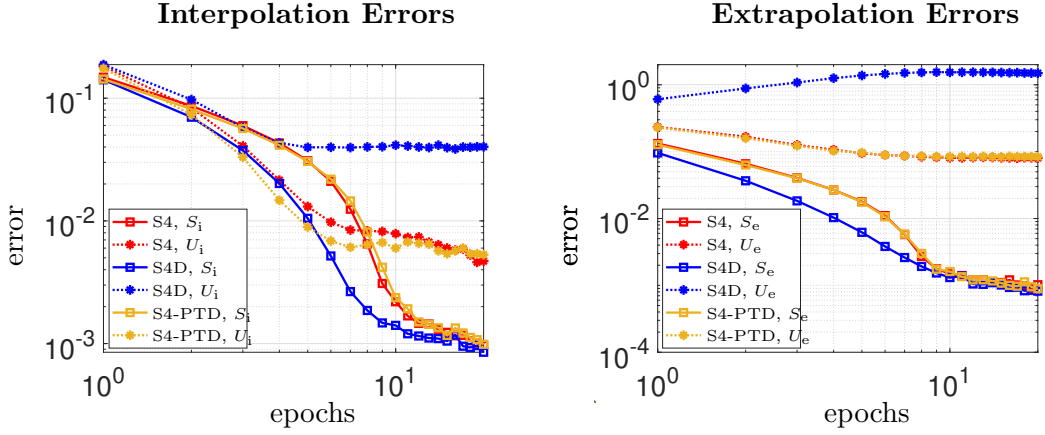
\begin{figure}
	\centering
	\begin{minipage}{.45\textwidth}
	\begin{overpic}[width=.9\textwidth]{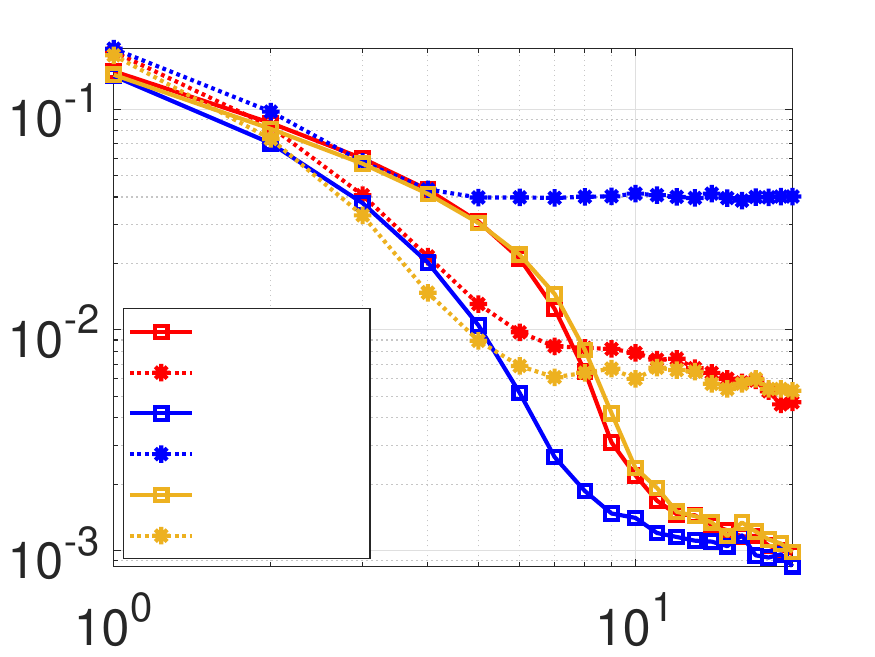}
	\put(24,78) {\textbf{Interpolation Errors}}
        \put(47,-4) {epochs}
        \put(-6,28) {\rotatebox{90}{error}}
        \put(23,36) {\tiny{S4, $S_{\text{i}}$}}
        \put(23,31.3) {\tiny{S4, $U_{\text{i}}$}}
        \put(23,26.6) {\tiny{S4D, $S_{\text{i}}$}}
        \put(23,21.9) {\tiny{S4D, $U_{\text{i}}$}}
        \put(23,17.2) {\tiny{S4-PTD, $S_{\text{i}}$}}
        \put(23,12.5) {\tiny{S4-PTD, $U_{\text{i}}$}}
        \end{overpic}
        \end{minipage}
        \begin{minipage}{.45\textwidth}
	\begin{overpic}[width=.9\textwidth]{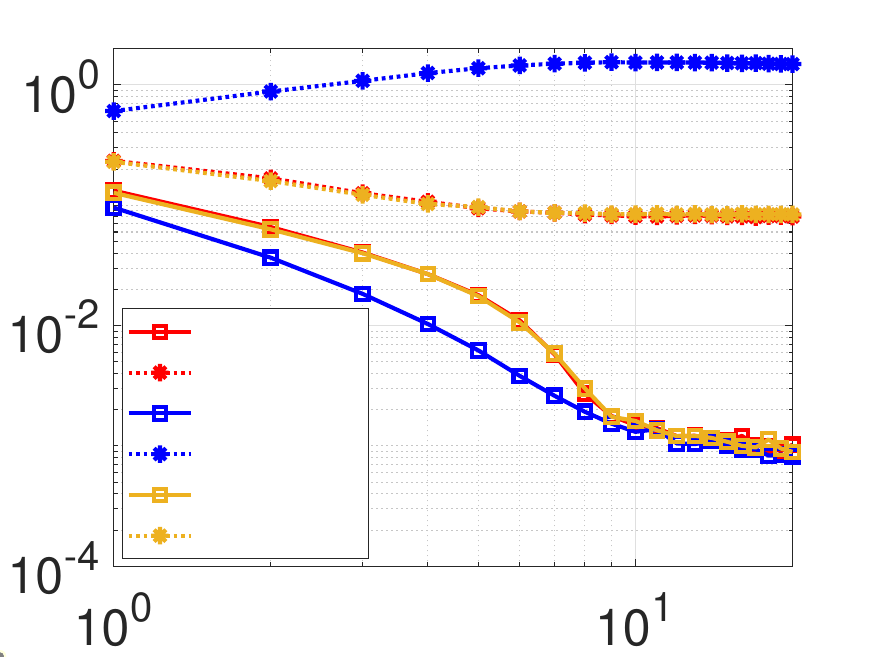}
	\put(24,78) {\textbf{Extrapolation Errors}}
        \put(47,-4) {epochs}
        \put(-6,28) {\rotatebox{90}{error}}
        \put(-6,28) {\rotatebox{90}{error}}
        \put(23,36) {\tiny{S4, $S_{\text{e}}$}}
        \put(23,31.3) {\tiny{S4, $U_{\text{e}}$}}
        \put(23,26.6) {\tiny{S4D, $S_{\text{e}}$}}
        \put(23,21.9) {\tiny{S4D, $U_{\text{e}}$}}
        \put(23,17.2) {\tiny{S4-PTD, $S_{\text{e}}$}}
        \put(23,12.5) {\tiny{S4-PTD, $U_{\text{e}}$}}
        \end{overpic}
        \end{minipage}
        \vspace{0.3cm}
        \caption{The interpolation and extrapolation errors of predicting the amplitude of a sinusoidal signal (see~\cref{sec:failuremodes}) made by the S4, S4D, and S4-PTD models. Each curve shows the mean-squared test error of one model on either the seen domain, $S_\text{i}$ or $S_\text{e}$, or the unseen domain, $U_\text{i}$ or $U_\text{e}$. The yellow curve for the S4-PTD model and the red curve for the S4 model almost overlap in the extrapolation problem.}
        \label{fig:interpextraperrs}
\end{figure}

In~\cref{sec:failuremodes}, we show a task of predicting the amplitude of a sinusoidal wave
\[
    \mathbf{u}(t) = A\sin(st).
\]
We define four domains by
\begin{align*}
    S_{\text{interp}} &= [0,40] \cup [60,100], \qquad S_{\text{extrap}} = [0,80],\\
    U_{\text{interp}} &= [0,100] \setminus S_{\text{interp}} = (40,60), \qquad U_{\text{extrap}} = [0,100] \setminus S_{\text{extrap}} = (80,100].
\end{align*}
We are given training samples only with $s \in S_{\text{interp}}$ (resp. $s \in S_{\text{extrap}}$) and we test the sequential model on the entire domain $[0,100]$. Given a test set, we measure the mean-squared error of our model. We uniformly sample the test set from $s \in S_{\text{interp}}$ (resp. $s \in S_{\text{extrap}}$) and $s \in U_{\text{interp}}$ (resp. $s \in U_{\text{extrap}}$) to evaluate our models' performance on generalization to unseen data in the seen domain, and their performance on interpolation (resp. extrapolation). The results are shown in~\Cref{fig:interpextraperrs}. We see that the S4D model performs even better on the seen domain (i.e., $S_{\text{interp}}$ or $S_{\text{extrap}}$), but its interpolation and extrapolation capabilities are much worse than those of the S4 and our S4-PTD models.

\section{Proof of results in~\cref{sec:S4-PTD}}\label{sec:proofperturb}

In this section, we present the proof of~\Cref{thm.perturbsys}. In addition, we introduce a probabilistic statement of the eigenvector condition number of a matrix perturbed by a random Gaussian matrix.

The proof of~\Cref{thm.perturbsys} is a classical forward error analysis, but to maintain the best result, we need to explicitly compute the resolvent of $\mathbf{A}_H$.

\begin{proof}[Proof of~\Cref{thm.perturbsys}]
For notational cleanliness, in this proof, we define $\mathbf{A} = \mathbf{A}_H$, $\mathbf{B} = \mathbf{B}_H$, and $\mathbf{C} = \mathbf{C}_{\text{DPLR}} \mathbf{V}_H^{-1}$. We have
    \begin{align*}
        \abs{G_{\text{Pert}}(s) - G_{\text{DPLR}}(s)} &= \abs{\mathbf B(s\mathbf I-\mathbf A)^{-1}\mathbf C - \mathbf B(s\mathbf I-\mathbf A-\mathbf E)^{-1}\mathbf C} \\
        &= \abs{\mathbf B\big((s\mathbf I-\mathbf A)^{-1}-(s\mathbf I-\mathbf A-\mathbf E)^{-1}\big)\mathbf C} \\
        &\leq \norm{\mathbf B}_2 \norm{(s\mathbf I-\mathbf A)^{-1}-(s\mathbf I-\mathbf A-\mathbf E)^{-1}}_2 \norm{\mathbf C}_2,
    \end{align*}
    where, by a result in~\cite{demmel1992componentwise}, we have
    \begin{equation}\label{eq.perturbinverse}
        \norm{(s\mathbf I-\mathbf A)^{-1}-(s\mathbf I-\mathbf A-\mathbf E)^{-1}}_2 \leq \norm{\mathbf E}_2 \norm{(s\mathbf I-\mathbf A)^{-1}}^2_2 + \mathcal{O}(\norm{\mathbf E}^2_2) \norm{(s\mathbf I-\mathbf A)^{-1}}_2.
    \end{equation}
    We set
    \[
        \begin{bmatrix}
            c_{1,1} \!&\! 0 \!&\! 0 \!&\! \cdots \!&\! 0 \\
            c_{2,1} \!&\! c_{2,2} \!&\! 0 \!&\! \cdots \!&\! 0 \\
            c_{3,1} \!&\! c_{3,2} \!&\! c_{3,3} \!&\! \cdots \!&\! 0 \\
            \vdots \!&\! \vdots \!&\! \vdots \!&\! \ddots \!&\! \vdots \\
            c_{n,1} \!&\! c_{n,2} \!&\! c_{n,3} \!&\! \cdots \!&\! c_{n,n}
        \end{bmatrix}
        \!\!=\!
        (-s\mathbf I+\mathbf A)^{-1}
        \!=\!\!\!
        \begin{bmatrix}
            1-s \!&\! 0 \!&\! 0 \!&\! \cdots \!&\! 0 \\
            \sqrt{3} \!&\! 2-s \!&\! 0 \!&\! \cdots \!&\! 0 \\
            \sqrt{5} \!&\! \sqrt{15} \!&\! 3-s \!&\! \cdots \!&\! 0 \\
            \vdots \!&\! \vdots \!&\! \vdots \!&\! \ddots \!&\! \vdots \\
            \sqrt{2n\!-\!1}\! \!&\! \!\sqrt{3(2n\!-\!1)}\! \!&\! \!\sqrt{5(2n\!-\!1)}\! \!&\! \cdots \!&\! \!n\!-\!s
        \end{bmatrix}^{\!-1}\!\!\!\!.
    \]
    Then, fixing a column $i$ and a row $j \geq i$, we have
    \begin{numcases}{}
            \sum_{k=i}^{j-1} c_{k,i} \sqrt{2j-1} \sqrt{2k-1} + c_{j,i}(j-s) &= 0,\label{eq.inverseeq1} \\
            \sum_{k=i}^{j} c_{k,i} \sqrt{2j+1} \sqrt{2k-1} + c_{j+1,i}(j+1-s) &= 0. \label{eq.inverseeq2}
    \end{numcases}
    Multiplying~\cref{eq.inverseeq1} by $\sqrt{2j+1}/\sqrt{2j-1}$, we have
    \begin{equation}\label{eq.inverseeq3}
        \sum_{k=i}^{j-1} c_{k,i} \sqrt{2j+1} \sqrt{2k-1} + \frac{\sqrt{2j+1}}{\sqrt{2j-1}}c_{j,i}(j-s) = 0.
    \end{equation}
    Subtracting~\cref{eq.inverseeq3} from~\cref{eq.inverseeq2}, we have
    \[
        c_{j,i}\sqrt{2j+1}\sqrt{2j-1} - c_{j,i}\frac{\sqrt{2j+1}}{\sqrt{2j-1}}(j-s) + c_{j+1,i}(j+1-s) = 0.
    \]
    After simplifying, we get the recurrence relation
    \begin{align*}
        &c_{i,i} = \frac{1}{i-s}, \qquad c_{i+1,i} = -\frac{\sqrt{2i-1}\sqrt{2i+1}}{(i-s)(i+1-s)}, \\
        &c_{j+1,i} = -\frac{(j+s-1)\sqrt{2j+1}}{(j-s+1)\sqrt{2j-1}} c_{j,i}, \qquad j \geq i+1.
    \end{align*}
    Solving this recurrence relation gives us
    \[ 
        c_{k,i} = (-1)^{k-i} \frac{\sqrt{2i-1}\sqrt{2i+1}}{(i-s)(i+1-s)} \frac{\sqrt{2k-1}}{\sqrt{2i+1}} \frac{\prod_{\ell=i}^{k-2}(\ell+s)}{\prod_{\ell=i+2}^{k}(\ell-s)}, \qquad k \geq i+1.
    \]
    Since $s$ is purely imaginary, we have
    \[
        \abs{\frac{\ell+s}{\ell-s}} = 1.
    \]
    Therefore, we can control the size of $c_{k,i}$ by\footnote{With a slight abuse of notation, the letter $i$ here stands for a real-valued index instead of the imaginary unit.}
    \begin{align*}
        \abs{c_{k,i}} &= \frac{\sqrt{2i-1}\sqrt{2k-1}}{\abs{i-s} \abs{i+1-s}} \frac{\abs{i+s} \abs{i+1+s}}{\abs{k-1-s} \abs{k-s}} = \frac{\sqrt{2i-1}\sqrt{2k-1}}{\abs{k-1-s}\abs{k-s}}, \qquad k \geq i+2.
    \end{align*}
    Clearly, this value is maximized when $s = 0$. Hence, we have
    \[
        \abs{c_{k,i}}^2 \leq \frac{(2i-1)(2k-1)}{(k-1)^2k^2} \leq \frac{4i}{(k-1)^2 k}.
    \]
    Note that this inequality holds also for the case when $k = i+1$. Now, we have
    \begin{align*}
        \norm{(s\mathbf I-\mathbf A)^{-1}}_2^2 &\leq \norm{(s\mathbf I-\mathbf A)^{-1}}_F^2 \leq \sum_{k=2}^n \sum_{i=1}^{k-1} \frac{4i}{(k-1)^2k} + \sum_{i=1}^n \frac{1}{i^2} \\
        &\leq \sum_{k=2}^n \frac{2(k-1)k}{(k-1)^2k} + 2 \leq 2\ln(n)+4.
    \end{align*}
    The result follows from~\cref{eq.perturbinverse}.
\end{proof}

Next, we prove~\Cref{thm.averagecond}. The proof is heavily based on~\cite[Thm.~1.5]{banks2021gaussian}.

\begin{proof}[Proof of~\Cref{thm.averagecond}]
By~\cite[Thm.~1.5]{banks2021gaussian}, we have that
\[
	\mathbb{E} \left[\sum_{j=1}^n \kappa(\lambda_i)^2 \mathbbm{1}_{\{\lambda_i \in D_R(0)\}}\right] \leq \norm{\mathbf{A}}^2 \frac{R^2 n^2}{\epsilon^2},
\]
where $\lambda_1, \ldots, \lambda_n$ are eigenvalues of $\mathbf{A} + \epsilon\mathbf{G}_n$ and $\kappa(\lambda_i)$ is defined in~\cite{banks2021gaussian}. When $\lambda_j \in D_R(0)$ for all $1 \leq j \leq n$, we have
\[
	\kappa_{\text{eig}}(\mathbf{A} + \epsilon\mathbf{G}_n)^2 \leq n\sum_{j=1}^n \kappa(\lambda_i)^2.
\]
Hence, this shows
\begin{align*}
	\frac{1}{n}\mathbb{E} \left[\kappa_{\text{eig}}(\mathbf{A} + \epsilon \mathbf{G}_n)^2 \middle| \Omega \right] \mathbb{P}(\Omega) + \mathbb{E} \left[\sum_{j=1}^n \kappa(\lambda_i)^2 \mathbbm{1}_{\{\lambda_i \in D_R(0)\}} \middle| \Omega^C \right] \mathbb{P}(\Omega^C) \leq \norm{\mathbf{A}}^2 \frac{R^2 n^2}{\epsilon^2}.
\end{align*}
We are done.
\end{proof}

Comparing~\Cref{thm.averagecond} to~\Cref{thm.bestcond}, we note that the bound in~\Cref{thm.bestcond} is slightly better than that in~\Cref{thm.averagecond}. However, the Gaussian perturbation in~\Cref{thm.averagecond} is problem-independent and can be generically implemented, whereas it is not necessarily easy to identify the perturbation in~\Cref{thm.bestcond}.

\section{Numerical experiments on~\Cref{thm.bestcond}}\label{sec:experimentperturb}

The performance of our perturbed model is heavily based on two things: the perturbation size $\|\mathbf{E}\|$ and the condition number of $\tilde{\mathbf{V}}_H$. The former value controls the difference between our initialization to the known-to-be-good HiPPO initialization, whereas the latter one controls the unfairness when transforming the states via $\tilde{\mathbf{V}}_H$. By~\Cref{thm.bestcond}, the condition number $\kappa(\tilde{\mathbf{V}}_H)$ should depend linearly on $\|E\|^{-1}$ and depend sub-quadratically on $n$, the state space size. 

\begin{figure}[!htb]
    \centering
    \begin{overpic}[width=0.52\textwidth]{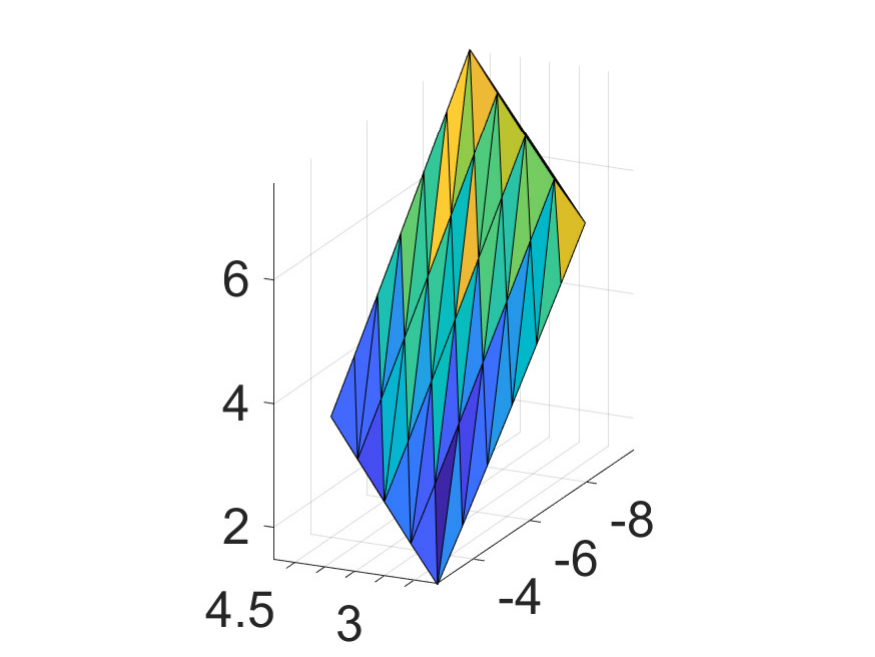}
        \put(29,-1) {\rotatebox{-10}{$\ln(n)$}}
        \put(60,-3) {\rotatebox{40}{$\ln(\|\mathbf{E}\|/\|\mathbf{A}_H\|)$}}
        \put(18,20) {\rotatebox{90}{$\ln(\kappa_{\text{eig}}(\tilde{\mathbf{A}}_H))$}}
    \end{overpic}
    \vspace{.3cm}
    \caption{The relationship among the state space dimension $n$, the relative perturbation size $\|\mathbf{E}\|/\|\mathbf{A}_H\|$, and the eigenvector condition number $\kappa_{\text{eig}}(\tilde{\mathbf{A}}_H)$.}
    \label{fig:eigcond}
\end{figure}

In this section, we present a numerical experiment that investigates the relationship between these three values. To do so, we solve the optimization problem in~\cref{eq.optobjective} with different state space dimensions $n$ and values of $\gamma$. We then record the size of the perturbation and the eigenvector condition number of the perturbed matrix. In~\Cref{fig:eigcond}, we see that the eigenvector condition number $\kappa_{\text{eig}}(\tilde{\mathbf{A}}_H)$ depends polynomially on both the state space dimension $n$ and the relative perturbation size $\|\mathbf{E}\|/\|\mathbf{A}_H\|$. Numerical values are reported in~\Cref{table:cond} and~\ref{table:perturbation}. Using the data, one can compute that we have $\kappa_{\text{eig}}(\tilde{\mathbf{A}}_H) = \mathcal{O}((\|\mathbf{E}\|/\|\mathbf{A}_H\|)^{-p})$, where $p \approx 0.87$. Hence, we are doing slightly better than the theory of~\Cref{thm.bestcond}. Another surprising observation that can be made with a little bit computation is that if we normalize $\mathbf{A}_H$ to have a spectral norm of $1$, then the eigenvector condition number $\kappa_{\text{eig}}(\tilde{\mathbf{A}}_H)$ does not depend on $n$ at all. This is much better than the bound proposed in~\Cref{thm.bestcond}.

\begin{table}
\centering
\begin{tabular}[b]{|c|c|c|c|c|c|c|c|}
    \hline
      \diagbox{$n$}{$\gamma$} & $10$ & $10^2$ & $10^3$ & $10^4$ & $10^5$ & $10^6$ & $10^7$ \\ \hline
      $8$ & $4.40\texttt{e}0$ & $8.62\texttt{e}0$ & $1.73\texttt{e}1$ & $3.51\texttt{e}1$ & $7.12\texttt{e}1$ & $1.45\texttt{e}2$ & $2.96\texttt{e}2$ \\ \hline
      $16$ & $6.59\texttt{e}0$ & $1.32\texttt{e}1$ & $2.69\texttt{e}1$ & $5.53\texttt{e}1$ & $1.14\texttt{e}2$ & $2.35\texttt{e}2$ & $4.86\texttt{e}2$ \\ \hline
      $32$ & $9.98\texttt{e}0$ & $2.02\texttt{e}1$ & $4.16\texttt{e}1$ & $8.63\texttt{e}1$ & $1.79\texttt{e}2$ & $3.72\texttt{e}2$ & $7.75\texttt{e}2$ \\ \hline
      $64$ & $1.52\texttt{e}1$ & $3.12\texttt{e}1$ & $6.45\texttt{e}1$ & $1.34\texttt{e}2$ & $2.80\texttt{e}2$ & $5.84\texttt{e}2$ & $1.22\texttt{e}3$ \\ \hline
      $128$ & $2.34\texttt{e}1$ & $4.82\texttt{e}1$ & $1.00\texttt{e}2$ & $2.09\texttt{e}2$ & $4.37\texttt{e}2$ & $9.14\texttt{e}2$ & $1.91\texttt{e}3$ \\
      \hline
    \end{tabular}
    \caption{The eigenvector condition number $\kappa_{\text{eig}}(\tilde{\mathbf{A}}_H)$ when the optimization problem~\cref{eq.optobjective} is solved with different values of $n$ and $\gamma$.}
     \label{table:cond}
\end{table}

\begin{table}
\centering
\begin{tabular}[b]{|c|c|c|c|c|c|c|c|}
    \hline
      \diagbox{$n$}{$\gamma$} & $10$ & $10^2$ & $10^3$ & $10^4$ & $10^5$ & $10^6$ & $10^7$ \\ \hline
      $8$ & $2.81\texttt{e}0$ & $1.16\texttt{e}0$ & $4.78\texttt{e}$-$1$ & $1.98\texttt{e}$-$1$ & $8.24\texttt{e}$-$2$ & $3.45\texttt{e}$-$2$ & $1.45$\texttt{e}-$2$ \\ \hline
      $16$ & $6.77\texttt{e}0$ & $2.86\texttt{e}0$ & $1.22\texttt{e}0$ & $5.18\texttt{e}$-$1$ & $2.22\texttt{e}$-$1$ & $9.50\texttt{e}$-$2$ & $4.09\texttt{e}$-$2$ \\ \hline
      $32$ & $1.62\texttt{e}1$ & $6.96\texttt{e}0$ & $3.00\texttt{e}0$ & $1.30\texttt{e}0$ & $5.62\texttt{e}$-$1$ & $2.45\texttt{e}$-$1$ & $1.07\texttt{e}$-$1$ \\ \hline
      $64$ & $3.89\texttt{e}1$ & $1.68\texttt{e}1$ & $7.32\texttt{e}0$ & $3.19\texttt{e}0$ & $1.39\texttt{e}0$ & $6.11\texttt{e}$-$1$ & $2.69\texttt{e}$-$1$ \\ \hline
      $128$ & $9.37\texttt{e}1$ & $4.07\texttt{e}1$ & $1.78\texttt{e}1$ & $7.80\texttt{e}0$ & $3.42\texttt{e}0$ & $1.51\texttt{e}0$ & $6.65\texttt{e}$-$1$ \\
      \hline
    \end{tabular}
    \vspace{.2cm}
    \caption{The perturbation size $\|\mathbf{E}\|$ when the optimization problem~\cref{eq.optobjective} is solved with different values of $n$ and $\gamma$.}
    \label{table:perturbation}
\end{table}

\section{Details of experiments in~\cref{sec:experiments}}\label{sec:experimentdetail}

In this section, we provide the details of the experiments presented in~\cref{sec:experiments}.

\subsection{Details of the evaluation of our model in the Long Range Arena}\label{sec:detailsLRA}

To compare our perturbed models with the diagonal S4D and S5 models, we adopt the same model parameters used in~\cite{gu2022parameterization,smith2023simplified} but possibly change the training parameters, such as the learning rate, number of epochs, batch size, and weight decay rate. For choosing the perturbation matrix, we again solve the optimization problem in~\cref{eq.optobjective}. Instead of allowing $\gamma$ to be an unbounded positive tuning parameter, we require that $\gamma$ is large enough so that $\|\mathbf{E}\| / \|\mathbf{A}_H\| \leq 0.1$. This improves the worst-case robustness of our model (see~\cref{sec:robustness}). We provide the detailed configuration of our S4-PTD model in~\Cref{tab:S4P} and that of our S5-PTD model in~\Cref{tab:S5P}. In particular, we note that the first two columns of~\Cref{tab:S4P} are almost the same as those in~\cite{gu2022parameterization}\footnote{The only exception is that in the Path-X task, we half the number of features in order to reduce the computation time. This only simplifies our perturbed model.} and the first four columns of~\Cref{tab:S5P} match those in~\cite{smith2023simplified} --- these are model parameters. The only remaining non-trivial thing is that in the Path-X task, we start with a batch size of $32$. We half the batch size after epoch $30$ and epoch $60$. By making the batch size smaller, we improve the generalization power of our model.

\begin{table}
    \centering
    \begin{tabular}{c c c c c c c c c c c}
         \specialrule{.1em}{.05em}{.05em}
         \footnotesize{Task} & \footnotesize{Depth} & \footnotesize{\#Features} & \footnotesize{Norm} & \footnotesize{Prenorm} & \footnotesize{DO} & \footnotesize{LR} & \footnotesize{BS} & \footnotesize{Epochs} & \footnotesize{WD} & \footnotesize{$\Delta$ Range} \\
         \hline
         \footnotesize{ListOps} & \footnotesize{8} & \footnotesize{256} & \footnotesize{BN} & \footnotesize{False} & \footnotesize{0.} & \footnotesize{0.002} & \footnotesize{50} & \footnotesize{80} & \footnotesize{0.05} & \footnotesize{(1e-3,1e0)} \\
         \footnotesize{Text} & \footnotesize{6} & \footnotesize{256} & \footnotesize{BN} & \footnotesize{True} & \footnotesize{0.} & \footnotesize{0.01} & \footnotesize{16} & \footnotesize{80} & \footnotesize{0.05} & \footnotesize{(1e-3,1e-1)} \\
         \footnotesize{Retrieval} & \footnotesize{6} & \footnotesize{128} & \footnotesize{BN} & \footnotesize{True} & \footnotesize{0.} & \footnotesize{0.004} & \footnotesize{64} & \footnotesize{40} & \footnotesize{0.03} & \footnotesize{(1e-3,1e-1)} \\
         \footnotesize{Image} & \footnotesize{6} & \footnotesize{128} & \footnotesize{LN} & \footnotesize{False} & \footnotesize{0.1} & \footnotesize{0.01} & \footnotesize{128} & \footnotesize{2000} & \footnotesize{0.01} & \footnotesize{(1e-3,1e-1)} \\
         \footnotesize{Pathfinder} & \footnotesize{6} & \footnotesize{512} & \footnotesize{BN} & \footnotesize{True} & \footnotesize{0.} & \footnotesize{0.004} & \footnotesize{64} & \footnotesize{300} & \footnotesize{0.03} & \footnotesize{(1e-2,1e0)} \\
         \footnotesize{Path-X} & \footnotesize{6} & \footnotesize{128} & \footnotesize{BN} & \footnotesize{True} & \footnotesize{0.} & \footnotesize{0.001} & \footnotesize{20} & \footnotesize{100} & \footnotesize{0.03} & \footnotesize{(1e-4,1e-1)} \\
         \specialrule{.1em}{.05em}{.05em}
    \end{tabular}
    \vspace{.2cm}
    \caption{Configurations of the S4-PTD model, where DO, LR, BS, and WD stand for dropout rate, learning rate, batch size, and weight decay, respectively.}
    \label{tab:S4P}
\end{table}

\begin{table}
    \centering
    \begin{tabular}{c c c c c c c c c c c}
         \specialrule{.1em}{.05em}{.05em}
         \footnotesize{Task} & \footnotesize{Depth} & \footnotesize{H} & \footnotesize{P} & \footnotesize{J} & \footnotesize{DO} & \footnotesize{LR} & \footnotesize{SSM LR} & \footnotesize{BS} & \footnotesize{Epochs} & \footnotesize{WD} \\
         \hline
         \footnotesize{ListOps} & \footnotesize{8} & \footnotesize{128} & \footnotesize{16} & \footnotesize{8} & \footnotesize{0.} & \footnotesize{0.003} & \footnotesize{0.001} & \footnotesize{50} & \footnotesize{35} & \footnotesize{0.05} \\
         \footnotesize{Text} & \footnotesize{6} & \footnotesize{256} & \footnotesize{192} & \footnotesize{12} & \footnotesize{0.1} & \footnotesize{0.004} & \footnotesize{0.001} & \footnotesize{50} & \footnotesize{40} & \footnotesize{0.07} \\
         \footnotesize{Retrieval} & \footnotesize{6} & \footnotesize{128} & \footnotesize{256} & \footnotesize{16} & \footnotesize{0.} & \footnotesize{0.002} & \footnotesize{0.001} & \footnotesize{32} & \footnotesize{20} & \footnotesize{0.05} \\
         \footnotesize{Image} & \footnotesize{6} & \footnotesize{512} & \footnotesize{384} & \footnotesize{3} & \footnotesize{0.1} & \footnotesize{0.0055} & \footnotesize{0.001} & \footnotesize{50} & \footnotesize{250} & \footnotesize{0.07} \\
         \footnotesize{Pathfinder} & \footnotesize{6} & \footnotesize{192} & \footnotesize{256} & \footnotesize{8} & \footnotesize{0.05} & \footnotesize{0.0045} & \footnotesize{0.0009} & \footnotesize{64} & \footnotesize{230} & \footnotesize{0.05} \\
         \footnotesize{Path-X} & \footnotesize{6} & \footnotesize{128} & \footnotesize{256} & \footnotesize{16} & \footnotesize{0.} & \footnotesize{0.0018} & \footnotesize{0.0006} & \footnotesize{32} & \footnotesize{90} & \footnotesize{0.06} \\
         \specialrule{.1em}{.05em}{.05em}
    \end{tabular}
    \vspace{.2cm}
    \caption{Configurations of the S5-PTD model. See~\cite{smith2023simplified} for the meaning of the parameter labels.}
    \label{tab:S5P}
\end{table}

\subsection{Details of the robustness test of the diagonal model and our model}

In the robustness test presented in~\cref{sec:robustness}, we train both an S4D model and an S4-PTD model. Our models have $4$ layers, $128$ channels, and each layer contains an SSM with $n = 32$ states. The perturbation matrix in the S4-PTD model is computed by setting $\gamma = 0.03$ in~\cref{eq.optobjective}. From~\Cref{fig:experiments}\subref{fig:expablation}, it can be seen that the perturbation thence computed has a magnitude of roughly $10\%$ of the magnitude of $\mathbf{A}_H$. We leave the training dataset and the validation dataset unchanged, but we add $10\%$ of noises in the form of $\cos(325.4 t)$ to the test dataset. The frequency $325.4$ is chosen at one of the sensitivity regions of the diagonal SSM when $n = 32$. We train both models for $50$ epochs and report the evolution of the training accuracy, the test accuracy on uncontaminated data, and that on noisy data.

\subsection{Details of the ablation study of our model}

In~\cref{sec:ablation}, we train models with different perturbation sizes to solve the sCIFAR task~\cite{krizhevsky2009learning}. Our models have the same architecture as those in the sensitivity test (see~\cref{sec:robustness}). We set the batch size to be $64$ and the learning rate to be $0.001$ for SSM parameters and $0.01$ for other model parameters. These are common setups that are adapted from the original S4 and S4D papers. We use the parameter $\gamma$ in~\cref{eq.optobjective} to control the size of the perturbation $\|\mathbf{E}\|$. We set $\gamma = 10^{-4}, 10^{-3}, \ldots, 10^9$ and train the S4-PTD model for $200$ epochs to learn a classifier. These correspond to the first $14$ points in~\Cref{fig:experiments}\subref{fig:expablation}, where we report both the test accuracy of the model and the eigenvector condition number at initialization. Since setting $\gamma$ small does not help reducing $\kappa_{\text{eig}}(\tilde{\mathbf{A}}_H)$ all the way down to $1$, the smallest condition number possible, to obtain the rightmost point, we perturb $\mathbf{A}_H$ by a random symmetric matrix with a large norm.

%% file: manuscript.bbl
\begin{thebibliography}{10}

\bibitem{antoulas1986scalar}
Athanasios~C. Antoulas and Brian~D.O. Anderson.
\newblock On the scalar rational interpolation problem.
\newblock {\em IMA Journal of Mathematical Control and Information},
  3(2-3):61--88, 1986.

\bibitem{arjovsky2016unitary}
Martin Arjovsky, Amar Shah, and Yoshua Bengio.
\newblock Unitary evolution recurrent neural networks.
\newblock In {\em International Conference on Machine Learning}, pages
  1120--1128. PMLR, 2016.

\bibitem{aumann2023practical}
Quirin Aumann and Ion~Victor Gosea.
\newblock Practical challenges in data-driven interpolation: dealing with
  noise, enforcing stability, and computing realizations.
\newblock {\em arXiv preprint arXiv:2301.04906}, 2023.

\bibitem{bai2018empirical}
Shaojie Bai, J~Zico Kolter, and Vladlen Koltun.
\newblock An empirical evaluation of generic convolutional and recurrent
  networks for sequence modeling.
\newblock {\em arXiv preprint arXiv:1803.01271}, 2018.

\bibitem{banks2022pseudospectral}
Jess Banks, Jorge Garza-Vargas, Archit Kulkarni, and Nikhil Srivastava.
\newblock Pseudospectral shattering, the sign function, and diagonalization in
  nearly matrix multiplication time.
\newblock {\em Foundations of Computational Mathematics}, pages 1--89, 2022.

\bibitem{banks2021gaussian}
Jess Banks, Archit Kulkarni, Satyaki Mukherjee, and Nikhil Srivastava.
\newblock Gaussian regularization of the pseudospectrum and davies’
  conjecture.
\newblock {\em Communications on Pure and Applied Mathematics},
  74(10):2114--2131, 2021.

\bibitem{chang2019antisymmetricrnn}
Bo~Chang, Minmin Chen, Eldad Haber, and Ed~H Chi.
\newblock Antisymmetricrnn: A dynamical system view on recurrent neural
  networks.
\newblock In {\em International Conference on Machine Learning}, 2019.

\bibitem{choromanski2020rethinking}
Krzysztof Choromanski, Valerii Likhosherstov, David Dohan, Xingyou Song,
  Andreea Gane, Tamas Sarlos, Peter Hawkins, Jared Davis, Afroz Mohiuddin,
  Lukasz Kaiser, et~al.
\newblock Rethinking attention with performers.
\newblock In {\em International Conference on Machine Learning}, 2020.

\bibitem{cohn2003further}
P.~M. Cohn.
\newblock {\em Further algebra and applications}.
\newblock Springer-Verlag London, Ltd., London, 2003.

\bibitem{davies2009perturbations}
E~Brian Davies and Mildred Hager.
\newblock Perturbations of {J}ordan matrices.
\newblock {\em Journal of Approximation Theory}, 156(1):82--94, 2009.

\bibitem{davies2008approximate}
E.B. Davies.
\newblock Approximate diagonalization.
\newblock {\em SIAM journal on matrix analysis and applications},
  29(4):1051--1064, 2008.

\bibitem{demmel1992componentwise}
James Demmel.
\newblock The componentwise distance to the nearest singular matrix.
\newblock {\em SIAM Journal on Matrix Analysis and Applications}, 13(1):10--19,
  1992.

\bibitem{erichson2021lipschitz}
N~Benjamin Erichson, Omri Azencot, Alejandro Queiruga, Liam Hodgkinson, and
  Michael~W Mahoney.
\newblock Lipschitz recurrent neural networks.
\newblock In {\em International Conference on Learning Representations}, 2021.

\bibitem{erichson2022gated}
N~Benjamin Erichson, Soon~Hoe Lim, and Michael~W Mahoney.
\newblock Gated recurrent neural networks with weighted time-delay feedback.
\newblock {\em arXiv preprint arXiv:2212.00228}, 2022.

\bibitem{gu2020hippo}
Albert Gu, Tri Dao, Stefano Ermon, Atri Rudra, and Christopher R{\'e}.
\newblock Hippo: Recurrent memory with optimal polynomial projections.
\newblock {\em Advances in neural information processing systems},
  33:1474--1487, 2020.

\bibitem{gu2022parameterization}
Albert Gu, Karan Goel, Ankit Gupta, and Christopher R{\'e}.
\newblock On the parameterization and initialization of diagonal state space
  models.
\newblock {\em Advances in Neural Information Processing Systems},
  35:35971--35983, 2022.

\bibitem{gu2022efficiently}
Albert Gu, Karan Goel, and Christopher Re.
\newblock Efficiently modeling long sequences with structured state spaces.
\newblock In {\em International Conference on Learning Representations}, 2022.

\bibitem{gu2021combining}
Albert Gu, Isys Johnson, Karan Goel, Khaled Saab, Tri Dao, Atri Rudra, and
  Christopher R{\'e}.
\newblock Combining recurrent, convolutional, and continuous-time models with
  linear state space layers.
\newblock {\em Advances in neural information processing systems}, 34:572--585,
  2021.

\bibitem{gu2022train}
Albert Gu, Isys Johnson, Aman Timalsina, Atri Rudra, and Christopher R{\'e}.
\newblock How to train your hippo: State space models with generalized
  orthogonal basis projections.
\newblock {\em International Conference on Learning Representations}, 2023.

\bibitem{hasani2022liquid}
Ramin Hasani, Mathias Lechner, Tsun-Hsuan Wang, Makram Chahine, Alexander
  Amini, and Daniela Rus.
\newblock Liquid structural state-space models.
\newblock {\em International Conference on Learning Representations}, 2023.

\bibitem{katharopoulos2020transformers}
Angelos Katharopoulos, Apoorv Vyas, Nikolaos Pappas, and Fran{\c{c}}ois
  Fleuret.
\newblock Transformers are rnns: Fast autoregressive transformers with linear
  attention.
\newblock In {\em International conference on machine learning}, pages
  5156--5165. PMLR, 2020.

\bibitem{kerg2019non}
Giancarlo Kerg, Kyle Goyette, Maximilian Puelma~Touzel, Gauthier Gidel, Eugene
  Vorontsov, Yoshua Bengio, and Guillaume Lajoie.
\newblock Non-normal recurrent neural network (nnrnn): learning long time
  dependencies while improving expressivity with transient dynamics.
\newblock {\em Advances in neural information processing systems}, 32, 2019.

\bibitem{kitaev2020reformer}
Nikita Kitaev, {\L}ukasz Kaiser, and Anselm Levskaya.
\newblock Reformer: The efficient transformer.
\newblock In {\em International Conference on Machine Learning}, 2020.

\bibitem{krizhevsky2009learning}
Alex Krizhevsky, Geoffrey Hinton, et~al.
\newblock Learning multiple layers of features from tiny images.
\newblock 2009.

\bibitem{kumar2022non}
Ankit Kumar and Kristofer Bouchard.
\newblock Non-normality in neural networks.
\newblock In {\em AI and Optical Data Sciences III}, volume 12019, pages
  70--76. SPIE, 2022.

\bibitem{nie2023a}
Yuqi Nie, Nam~H Nguyen, Phanwadee Sinthong, and Jayant Kalagnanam.
\newblock A time series is worth 64 words: Long-term forecasting with
  transformers.
\newblock In {\em The Eleventh International Conference on Learning
  Representations}, 2023.

\bibitem{orhan2019improved}
A~Emin Orhan and Xaq Pitkow.
\newblock Improved memory in recurrent neural networks with sequential
  non-normal dynamics.
\newblock {\em Internation Conference on Learning Representations}, 2020.

\bibitem{orvieto2023resurrecting}
Antonio Orvieto, Samuel~L Smith, Albert Gu, Anushan Fernando, Caglar Gulcehre,
  Razvan Pascanu, and Soham De.
\newblock Resurrecting recurrent neural networks for long sequences.
\newblock {\em arXiv preprint arXiv:2303.06349}, 2023.

\bibitem{romero2021ckconv}
David~W Romero, Anna Kuzina, Erik~J Bekkers, Jakub~M Tomczak, and Mark
  Hoogendoorn.
\newblock Ckconv: Continuous kernel convolution for sequential data.
\newblock In {\em International Conference on Machine Learning}, 2022.

\bibitem{rusch2021unicornn}
T~Konstantin Rusch and Siddhartha Mishra.
\newblock Unicornn: A recurrent model for learning very long time dependencies.
\newblock In {\em International Conference on Machine Learning}, pages
  9168--9178. PMLR, 2021.

\bibitem{sengupta2018robust}
Biswa Sengupta and Karl~J Friston.
\newblock How robust are deep neural networks?
\newblock {\em arXiv preprint arXiv:1804.11313}, 2018.

\bibitem{smith2023simplified}
Jimmy~T.H. Smith, Andrew Warrington, and Scott Linderman.
\newblock Simplified state space layers for sequence modeling.
\newblock In {\em The Eleventh International Conference on Learning
  Representations}, 2023.

\bibitem{tay2020long}
Yi~Tay, Mostafa Dehghani, Samira Abnar, Yikang Shen, Dara Bahri, Philip Pham,
  Jinfeng Rao, Liu Yang, Sebastian Ruder, and Donald Metzler.
\newblock Long range arena: A benchmark for efficient transformers.
\newblock {\em International Conference in Learning Representations}, 2021.

\bibitem{trefethen2005spectra}
Lloyd~N Trefethen and Mark Embree.
\newblock {\em Spectra and Pseudospectra: The Behaviour of Non-normal Matrices
  and Operators}.
\newblock Springer, 2005.

\bibitem{voelker2019legendre}
Aaron Voelker, Ivana Kaji{\'c}, and Chris Eliasmith.
\newblock Legendre memory units: Continuous-time representation in recurrent
  neural networks.
\newblock {\em Advances in neural information processing systems}, 32, 2019.

\bibitem{woodbury1950inverting}
Max~A. Woodbury.
\newblock {\em Inverting modified matrices}.
\newblock Princeton University, Princeton, N. J., 1950.
\newblock Statistical Research Group, Memo. Rep. no. 42,.

\bibitem{yildiz2021continuous}
Cagatay Yildiz, Markus Heinonen, and Harri L{\"a}hdesm{\"a}ki.
\newblock Continuous-time model-based reinforcement learning.
\newblock In {\em International Conference on Machine Learning}, pages
  12009--12018. PMLR, 2021.

\bibitem{zhou1998essentials}
Kemin Zhou and John~Comstock Doyle.
\newblock {\em Essentials of robust control}, volume 104.
\newblock Prentice Hall, Upper Saddle River, NJ, 1998.

\bibitem{zhou2022fedformer}
Tian Zhou, Ziqing Ma, Qingsong Wen, Xue Wang, Liang Sun, and Rong Jin.
\newblock Fedformer: Frequency enhanced decomposed transformer for long-term
  series forecasting.
\newblock In {\em International Conference on Machine Learning}, pages
  27268--27286. PMLR, 2022.

\end{thebibliography}
